\RequirePackage{fix-cm}
\documentclass[smallcondensed]{svjour3}     

\usepackage{hyperref}
\usepackage{url}
\usepackage[utf8]{inputenc} 
\usepackage[T1]{fontenc}    
\usepackage{hyperref}       
\usepackage{url}            
\usepackage{booktabs}       
\usepackage{amsfonts}       
\usepackage{nicefrac}       
\usepackage{microtype}      
\usepackage{graphicx}
\usepackage{enumitem}       
\usepackage{xcolor}         
\usepackage{subcaption}     
\usepackage{csquotes}       
\usepackage{tabularx}
\usepackage{etoolbox}
\newcommand*{\affaddr}[1]{#1} 
\newcommand*{\affmark}[1][*]{\textsuperscript{#1}}
\smartqed  
\usepackage{amsmath}
\DeclareMathOperator*{\argmax}{\arg\!\max}
\DeclareUnicodeCharacter{202A}{ }
\DeclareUnicodeCharacter{202C}{ }
\usepackage{appendix}
\usepackage[numbers]{natbib}

\usepackage{changes}
\usepackage{amsmath}
\usepackage{algorithm}
\usepackage[noend]{algpseudocode}

\makeatletter
\def\BState{\State\hskip-\ALG@thistlm}
\makeatother

\begin{document}
\title{Quick and Robust Feature Selection: the Strength of Energy-efficient Sparse Training for Autoencoders\thanks{This paper has been accepted for publication in the Machine Learning Journal (ECML-PKDD 2022 Journal Track)}}
\titlerunning{Quick and Robust Feature Selection}     
\author{Zahra Atashgahi \and Ghada Sokar\and Tim van der Lee\and  Elena Mocanu \and Decebal Constantin Mocanu \and Raymond Veldhuis \and ‪Mykola Pechenizkiy‬}


\institute{  Z. Atashgahi \and E. Mocanu \and D.C. Mocanu \and 
            R.N.J. Veldhuis\at
            Faculty of Electrical Engineering, Mathematics and Computer Science, University of Twente, Enschede, 7500AE, the Netherland \\
              \email{z.atashgahi@utwente.nl}        
           \and
           G.A.Z.N. Sokar\affmark[1] \and T. Lee\affmark[2] \and D.C. Mocanu\affmark[1] \and M.  Pechenizkiy\affmark[1]\at
              Eindhoven University of Technology, 5600 MB Eindhoven, the Netherlands\\
              \affaddr{\affmark[1]Department of Mathematics and Computer Science } 
\affaddr{\affmark[2]Department of Electrical Engineering}
\and
 M.  Pechenizkiy \at
Faculty of Information Technology, University of Jyväskylä, 40014 Jyväskylä, Finland
}

\date{Received: date / Accepted: date}
\maketitle


\begin{abstract}
Major complications arise from the recent increase in the amount of high-dimensional data, including high computational costs and memory requirements. Feature selection, which identifies the most relevant and informative attributes of a dataset, has been introduced as a solution to this problem. Most of the existing feature selection methods are computationally inefficient; inefficient algorithms lead to high energy consumption, which is not desirable for devices with limited computational and energy resources. In this paper, a novel and flexible method for unsupervised feature selection is proposed. This method, named QuickSelection\footnote{The code is available at: \url{https://github.com/zahraatashgahi/QuickSelection}}, introduces the strength of the neuron in sparse neural networks as a criterion to measure the feature importance. This criterion, blended with sparsely connected denoising autoencoders trained with the sparse evolutionary training procedure, derives the importance of all input features simultaneously. We implement QuickSelection in a purely sparse manner as opposed to the typical approach of using a binary mask over connections to simulate sparsity. It results in a considerable speed increase and memory reduction. When tested on several benchmark datasets, including five low-dimensional and three high-dimensional datasets, the proposed method is able to achieve the best trade-off of classification and clustering accuracy, running time, and maximum memory usage, among widely used approaches for feature selection. Besides, our proposed method requires the least amount of energy among the state-of-the-art autoencoder-based feature selection methods. 

\keywords{Feature Selection \and Deep Learning\and Sparse Autoencoders \and Sparse Training}
\end{abstract}

\section{Introduction}
    \label{intoduction}
    In the last few years, considerable attention has been paid to the problem of dimensionality reduction and many approaches have been proposed \cite{van2009dimensionality}. There are two main techniques for reducing the number of features of a high-dimensional dataset: feature extraction and feature selection. Feature extraction focuses on transforming the data into a lower-dimensional space. This transformation is done through a mapping which results in a new set of features \cite{liu1998feature}. Feature selection reduces the feature space by selecting a subset of the original attributes without generating new features \cite{chandrashekar2014survey}. Based on the availability of the labels, feature selection methods are divided into three categories: supervised \cite{ang2015supervised,chandrashekar2014survey}, semi-supervised \cite{zhao2007semi,sheikhpour2017survey}, and unsupervised \cite{miao2016survey,dy2004feature}. Supervised feature selection algorithms try to maximize some function of predictive accuracy given the class labels. In unsupervised learning, the search for discriminative features is done blindly, without having the class labels. Therefore, unsupervised feature selection is considered as a much harder problem \cite{dy2004feature}.\looseness=-1
    
    Feature selection methods improve the scalability of machine learning algorithms since they reduce the dimensionality of data. Besides, they reduce the ever-increasing demands for computational and memory resources that are introduced by the emergence of big data. This can lead to a considerable decrease in energy consumption in data centers. This can ease not only the problem of high energy costs in data centers but also the critical challenges imposed on the environment \cite{yang2018ai}. As outlined by the High-Level Expert Group on Artificial Intelligence (AI) \cite{AIethic2019}, environmental well-being is one of the requirements of a trust-worthy AI system. The development, deployment, and process of an AI-system should be assessed to ensure that they would function in the most environmentally friendly way possible. For example, resource usage and energy consumption through training can be evaluated.\looseness=-1 
    
\begin{figure}[!t]
    
        \vskip 0.1in
        \begin{center}
        \centerline{\includegraphics[width=\textwidth]{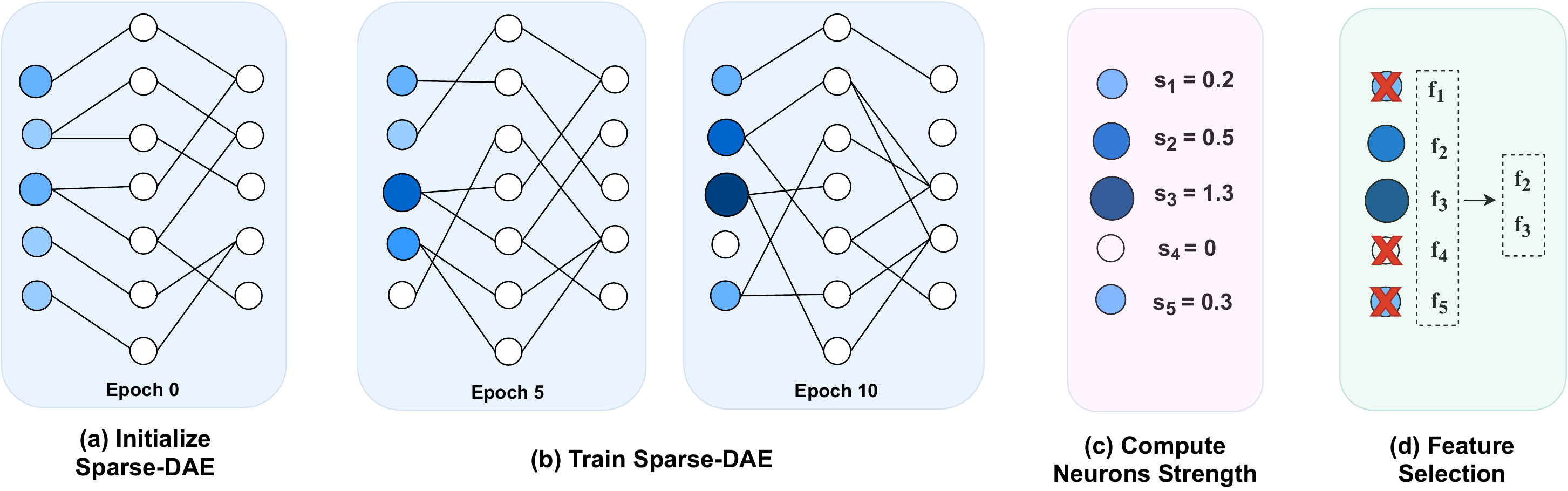}}
        \caption{A high-level overview of the proposed method, \enquote{QuickSelection}. (a) At epoch $0$, connections are randomly initialized. (b) After initializing the sparse structure, we start the training procedure. After $5$ epochs, some connections are changed during the training procedure, and as a result, the strength of some neurons has increased or decreased. At epoch 10, the network has converged, and we can observe which neurons are important (larger and darker blue circles) and which are not.  (c) When the network is converged, we compute the strength of all input neurons. (d) Finally, we select $K$ features corresponding to neurons with the highest strength values. }
        \label{fig:diagram}
        \end{center}
        \vskip -0.35in
        
    \end{figure}

    However, a challenging problem that arises in the feature selection domain is that selecting features from datasets that contain a huge number of features and samples, may require a massive amount of memory, computational, and energy resources. Since most of the existing feature selection techniques were designed to process small-scale data, their efficiency can be downgraded with high-dimensional data \cite{bolon2015feature}. Only a few studies have focused on designing feature selection algorithms that are efficient in terms of computation \cite{tan2014towards,aghazadeh2018mission}. The main contributions of this paper can be summarized as follows:
    
    \begin{itemize}

        \item[$\bullet$]We propose a new fast and robust unsupervised feature selection method, named QuickSelection. As briefly sketched in Figure~\ref{fig:diagram}, It has two key components: (1) Inspired by node strength in graph theory,  the method proposes the neuron strength of sparse neural networks as a criterion to measure the feature importance; and (2) The method introduces sparsely connected Denoising Autoencoders (sparse DAEs) trained from scratch with the sparse evolutionary training procedure to model the data distribution efficiently. The imposed sparsity before training also reduces the amount of required memory and the training running time.
        
        \item[$\bullet$]We implement QuickSelection in a completely sparse manner in Python using the SciPy library and Cython rather than using a binary mask over connections to simulate sparsity. This ensures minimum resource requirements, i.e., just Random-Access Memory (RAM) and Central Processing Unit (CPU), without demanding Graphic Processing Unit (GPU).
        \end{itemize}
 
    The experiments performed on 8 benchmark datasets suggest that QuickSelection has several advantages over the state-of-the-art, as follows:  
     \begin{itemize}
        \item[$\bullet$]It is the first or the second-best performer in terms of both classification and clustering accuracy in almost all scenarios considered.  
        \item[$\bullet$]It is the best performer in terms of the trade-off between classification and clustering accuracy, running time, and memory requirement. 
        \item[$\bullet$]The proposed sparse architecture for feature selection has at least one order of magnitude fewer parameters than its dense equivalent. This leads to the outstanding fact that the wall clock training time of QuickSelection running on CPU is smaller than the wall clock training time of its autoencoder-based competitors running on GPU in most of the cases.  
        \item[$\bullet$]Last but not least, QuickSelection computational efficiency makes it have the minimum energy consumption among the autoencoder-based feature selection methods considered. 
     \end{itemize}
    


\section{Related Work}\label{ssec:related_work_feature_selection}
    \subsection{Feature Selection}
    The literature on feature selection shows a variety of approaches that can be divided into three major categories, including filter, wrapper, and embedded methods. Filter methods use a ranking criterion to score the features and then remove the features with scores below a threshold. These criteria can be Laplacian score \cite{he2006laplacian}, Correlation, Mutual Information \cite{chandrashekar2014survey}, and many other scoring methods such as Bayesian scoring function, t-test scoring, and Information theory-based criteria \cite{lazar2012survey}. These methods are usually fast and computationally efficient. Wrapper methods evaluate different subsets of features to detect the best subset. Wrapper methods usually give better performance than filter methods; they use a predictive model to score each subset of features. However, this results in high computation complexity. Seminal contributions for this type of feature selection have been made by \citeauthor{kohavi1997wrappers} \cite{kohavi1997wrappers}. In \cite{kohavi1997wrappers}, the authors used a tree structure to evaluate the subsets of features. Embedded methods unify the learning process, and the feature selection \cite{lal2006embedded}. Multi-Cluster Feature Selection (MCFS) \cite{cai2010unsupervised} is an unsupervised method for embedded feature selection, which selects features using spectral regression with L1-norm regularization. A key limitation of this algorithm is that it is computationally intensive since it depends on computing the eigenvectors of the data similarity matrix and then solving an L1-regularized regression problem for each eigenvector \cite{farahat2013efficient}. Unsupervised Discriminative Feature Selection (UDFS) \cite{yang2011l2} is another unsupervised embedded feature selection algorithm that simultaneously utilizes both feature and discriminative information to select features \cite{li2018feature}. 
    
    \subsection{Autoencoders for Feature Selection}  
    In the last few years, many deep learning-based models have been developed to select features from the input data using the learning procedure of deep neural networks \cite{li2016deep}. In \cite{lu2018deeppink}, a Multi-Layer Perceptron (MLP) is augmented with a pairwise-coupling layer to feed each input feature along with its knockoff counterpart into the network. After the training, the authors use the filter weights of the pairwise-coupling layer to rank input features. Autoencoders which are generally known as a strong tool for feature extraction \cite{bengio2013representation}, are being explored to perform unsupervised feature selection. In \cite{han2018autoencoder}, authors combine autoencoder regression and group lasso task for unsupervised feature selection named AutoEncoder Feature Selector (AEFS). In \cite{doquet2019agnostic}, an autoencoder is combined with three variants of structural regularization to perform unsupervised feature selection. These regularizations are based on slack variables, weights, and gradients, respectively. Another recently proposed autoencoder-based embedded method is feature selection with Concrete Autoencoder (CAE) \cite{balin2019concrete}. This method selects features by learning a concrete distribution over input features. They proposed a concrete selector layer that selects a linear combination of input features that converges to a discrete set of $K$ features during training. In \cite{singh2020fsnet}, the authors showed that a large set of parameters in CAE might lead to over-fitting in case of having a limited number of samples. In addition, CAE may select features more than once since there is no interaction between the neurons of the selector layer. To mitigate these problems, they proposed a concrete neural network feature selection (FsNet) method, which includes a selector layer and a supervised deep neural network. The training procedure of FsNet considers reducing the reconstruction loss and maximizing the classification accuracy simultaneously. In our research, we focus mostly on unsupervised feature selection methods.
    
    Denoising Autoencoder (DAE) is introduced to solve the problem of learning the identity function in the autoencoders. This problem is most likely to happen when we have more hidden neurons than inputs \cite{baldi2012autoencoders}. As a result, the network output may be equal to the inputs, which makes the autoencoder useless. DAEs solve the aforementioned problem by introducing noise on the input data and trying to reconstruct the original input from its noisy version \cite{vincent2008extracting}. As a result, DAEs learn a representation of the input data that is robust to small irrelevant changes in the input. In this research, we use the ability of this type of neural network to encode the input data distribution and select the most important features. Moreover, we demonstrate the effect of noise addition on the feature selection results.
    
    \subsection{Sparse Training}  
    Deep neural networks usually have at least some fully-connected layers, which results in a large number of parameters. In a high-dimensional space, this is not desirable since it may cause a significant decrease in training speed and a rise in memory requirement. To tackle this problem, sparse neural networks have been proposed. Pruning the dense neural networks is one of the most well-known methods to achieve a sparse neural network \cite{lecun1990optimal, hassibi1993second}. In \cite{han2015learning}, \citeauthor{han2015learning} start from a pre-trained network, prune the unimportant weights, and retrain the network. Although this method can output a network with the desired sparsity level, the minimum computation cost is as much as the cost of training a dense network. To reduce this cost, \citeauthor{lee2018snip} \cite{lee2018snip} start with a dense neural network, and prune it prior to training based on connection sensitivity; then, the sparse network is trained in the standard way. However, starting from a dense neural network requires at least the memory size of the dense neural network and the computational resources for one training iteration of a dense network. Therefore, this method might not be suitable for low resource devices.      
    
    In 2016, \citeauthor{mocanu2016topological} \cite{mocanu2016topological} have introduced the idea of training sparse neural networks from scratch, a concept which recently has started to be known as sparse training. The sparse connectivity pattern was fixed before training using graph theory, network science, and data statistics. While it showed promising results, outperforming the dense counterpart, the static sparsity pattern did not always model the data optimally. In order to address these issues, in 2018, \citeauthor{mocanu2018scalable} \cite{mocanu2018scalable} have proposed the Sparse Evolutionary Training (SET) algorithm which makes use of dynamic sparsity during training. The idea is to start with a sparse neural network before training and dynamically change its connections during training in order to automatically model the data distribution. This results in a significant decrease in the number of parameters and increased performance. SET evolves the sparse connections at each training epoch by removing a fraction $\zeta$ connections with the smallest magnitude, and randomly adding new connections in each layer. \citeauthor{pmlr-v97-peterson19a} \cite{pmlr-v97-peterson19a} have shown that a sparse MLP trained with SET achieves state-of-the-art results on tabular data in predicting human decisions, outperforming fully-connected neural networks and Random Forest, among others. 
     
    In this work, we introduce for the first time sparse training in the world of denoising autoencoders, and we named the newly introduced model sparse denoising autoencoder (sparse DAE). We train the sparse DAE with the SET algorithm to keep the number of parameters low, during the training. Then, we then exploit the trained network to select the most important features. 

\section{Proposed Method}
    To address the problem of the high dimensionality of the data, we propose a novel method, named \enquote{QuickSelection}, to select the most informative attributes from the data, based on their strength (importance). In short, we train a sparse denoising autoencoder network from scratch in an unsupervised adaptive manner. Then, we use the trained network to derive the strength of each neuron in the input features.
    
    The basic idea of our proposed approach is to impose sparse connections on DAE, which proved its success in the related field of feature extraction, to efficiently handle the computational complexity of high-dimensional data in terms of memory resources. Sparse connections are evolved in an adaptive manner that helps in identifying informative features.
    
    A couple of methods have been proposed for training deep neural networks from scratch using sparse connections and sparse training \cite{dettmers2019sparse,mocanu2018scalable,bellec2017deep,pmlr-v97-mostafa19a, evci2019rigging, zhu2019multi}. All these methods are implemented using a binary mask over connections to simulate sparsity since all standard deep learning libraries and hardware (e.g. GPUs) are not optimized for sparse weight matrix operations. Unlike the aforementioned methods, we implement our proposed method in a purely sparse manner to meet our goal of actually using the advantages of a very small number of parameters during training. We decided to use SET in training our sparse DAE. 
    
   The choice of SET is due to its desirable characteristic. SET is a simple method yet achieves satisfactory performance. Unlike other methods that calculate and store information for all the network weights, including the non-existing ones, SET is memory efficient. It stores the weights for the existing sparse connections only. It does not need any high computational complexity as the evolution procedure depends on the magnitude of the existing connections only. This is a favourable advantage to our proposed method to select informative features quickly. In the following subsections, we first present the structure of our proposed sparse denoising autoencoder network and then explain the feature selection method. The pseudo-code of our proposed method can be found in Algorithm \ref{alg:QuickSelection}.

    \subsection{Sparse DAE}
    \textbf{Structure}
    As the goal of our proposed method is to do fast feature selection in a memory-efficient way, we consider here the model with the least possible number of hidden layers, one hidden layer, as more layers mean more computation. Initially, sparse connections between two consecutive layers of neurons are initialized with an Erdős–Rényi random graph, in which the probability of the connection between two neurons is given by
    \begin{equation}
        P(W^{l}_{ij})=  \frac{\epsilon(n^{l-1}+n^l)}{n^{l-1} \times n^{l}}, 
        \end{equation}
    where $\epsilon$ denotes the parameter that controls the sparsity level, $n^l$ denotes number of neurons at layer $l$, and ${{W}_{ij}^l}$ is the connection between neuron $i$ in layer $l-1$ and neuron $j$ in layer $l$, stored in the sparse weight matrix $\mathbf{W}^l$.
    
    \textbf{Input denoising} We use the additive noise model to corrupt the original data:  
    \begin{equation}
        \mathbf{\widetilde{x}} = \mathbf{x} + \mathit{nf} \mathcal{N}(\mu,\,\sigma^{2}),
        \end{equation}
    where $\mathbf{x}$ is the input data vector from dataset $X$, $\mathit{nf}$ (noise factor) is a hyperparameter of the model which determines the level of corruption, and $ \mathcal{N}(\mu,\,\sigma^{2})\ $is a Gaussian noise. After denoising the data, we derive the hidden representation $\mathbf{h}$ using this corrupted input. Then, the output $\mathbf{z}$ is reconstructed from the hidden representation. Formally, the hidden representation  $\mathbf{h}$ and the output  $\mathbf{z}$ are computed as follows:
    \begin{equation}
        \mathbf{h} = a(\mathbf{W}^1\mathbf{\widetilde{x}} + \mathbf{b}^1),
        \end{equation}
    \begin{equation}
        \mathbf{z} = a(\mathbf{W}^2\mathbf{h} + \mathbf{b}^2),
        \end{equation}
    where $\mathbf{W}^1$ and $\mathbf{W}^2$ are the sparse weight matrices of hidden and output layers respectively, $\mathbf{b}^1$ and $\mathbf{b}^2$ are the bias vectors of their corresponding layer, and $a$ is the activation function of each layer.
    The objective of our network is to reconstruct the original features in the output. For this reason, we use mean squared error (MSE) as the loss function to measure the difference between original features $\mathbf{x}$ and the reconstructed output $\mathbf{z}$:
    \begin{equation}
        L_{MSE}=  \left\Vert \mathbf{z}-\mathbf{x} \right\Vert_2^2.
        \end{equation}
    
    Finally, the weights can be optimized using the standard training algorithms (e.g., Stochastic Gradient Descent (SGD), AdaGrad, and Adam) with the above reconstruction error.
    
    \textbf{Training procedure} We adapt the SET training procedure \cite{mocanu2018scalable} in training our proposed network for feature selection. SET works as follows. After each training epoch, a fraction $\zeta$ of the smallest positive weights and a fraction $\zeta$ of the largest negative weights at each layer is removed. The selection is based on the magnitude of the weights. New connections in the same amount as the removed ones are randomly added in each layer. Therefore the total number of connections in each layer remains the same, while the number of connections per neuron varies, as represented in Figure \ref{fig:diagram}. The weights of these new connections are initialized from a standard normal distribution.The random addition of new connections do not have a high risk of not finding a good sparse connectivity at the end of the training process because it has been shown in \cite{liu2020topological} that sparse training can unveil a vast number of very different sparse connectivity local optima which achieve a very similar performance. 
    \begin{figure}[!t]
        \centering
        \includegraphics[width=\linewidth]{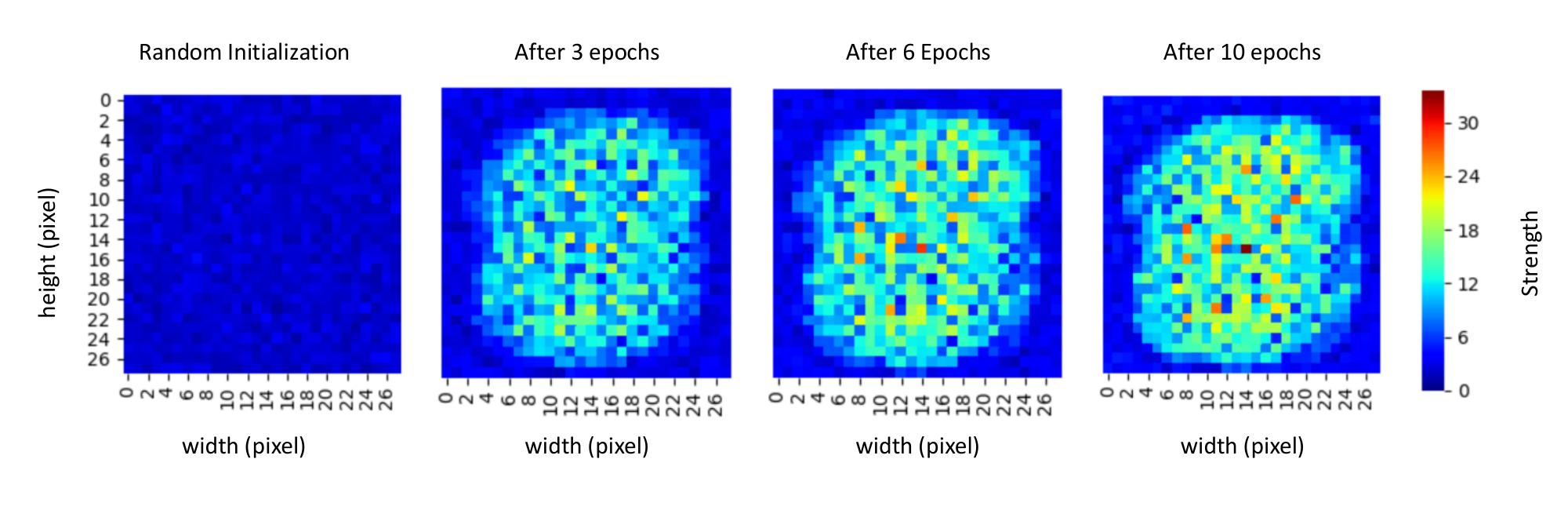}
        \caption{Neuron's strength on the MNIST dataset. The heat-maps above are a 2D representation of the input neuron's strength. It can be observed that the strength of neurons is random at the beginning of training. After a few epochs, the pattern changes, and neurons in the center become more important and similar to the MNIST data pattern. }
        \label{fig:weightdist}
        \end{figure} 
    

    \begin{algorithm}[!b]
    \caption{QuickSelection}\label{euclid}
        \begin{algorithmic}[1]
        
        \State \textbf{Input:} Dataset $X$, noise factor $\mathit{nf}$, sparsity hyperparameter $\epsilon$, number of hidden neurons $n^{h}$, number of selected features $K$
        \State Input denoising:  $\mathbf{\widetilde{x}} = \mathbf{x} + \mathit{nf} \mathcal{N}(\mu,\,\sigma^{2}) $
        \State \textbf{Structure initialization:} Initialize sparse-DAE with $n^h$ hidden neurons and sparsity level determined by $\epsilon$  
        \Procedure{Training sparse-DAE}{}
        \State Let the loss be $L_{MSE}=  \left\Vert \mathbf{z}-\mathbf{x} \right\Vert_2^2$ where $z$ is the output of sparse-DAE
        \For{\texttt{$i \in \{1,\dots,epochs\}$}}
            \State Perform standard forward propagation and backpropagation
            \State Perform weight removal and addition for topology optimization
        \EndFor
        \EndProcedure
        
        \Procedure{QuickSelection}{}
        \State \text{Compute neurons strength: }
        \For{\texttt{$i \in \{1,\dots,\#input features\}$}}
            \State \texttt{${{s}_{i}}=\sum\limits_{j=1}^{n^h}{|{{W}^{1}_{ij}}}|$
            }
        \EndFor
        \State Select $K$ features:  $\mathbb{F}^{*}_s = \argmax_{ \mathbb{F}_s \subset  \mathbb{F}, | \mathbb{F}_s|=K} \sum\limits_{f_i\in  \mathbb{F}_s}s_i,$

        \EndProcedure

        \end{algorithmic}
        \label{alg:QuickSelection}
    \end{algorithm}
    
    \subsection{Feature Selection}
    We select the most important features of the data based on the weights of their corresponding input neurons of the trained sparse DAE.
    Inspired by node strength in graph theory \cite{barrat2004architecture}, we determine the importance of each neuron based on its \textbf{strength}. We estimate the strength of each neuron (${{s}_{i}}$) by the summation of absolute weights of its outgoing connections.
    \begin{equation}
    {{s}_{i}}=\sum\limits_{j=1}^{n^1}{|{{W}^{1}_{ij}}}|,
    \end{equation}
    where $n^1$ is the number of neurons of the first hidden layer, and ${{W}^{1}_{ij}}$ denotes the weight of connection linking input neuron $i$ to hidden neuron $j$.

    As represented in Figure \ref{fig:diagram}, the strength of the input neurons changes during training; we have depicted the strength of the neurons according to their size and color. After convergence, we compute the strength for all of the input neurons; each input neuron corresponds to a feature. Then, we select the features corresponding to the neurons with $K$ largest strength values:
    \begin{equation}
        \mathbb{F}^{*}_s = \argmax_{ \mathbb{F}_s \subset  \mathbb{F}, | \mathbb{F}_s|=k} \sum\limits_{f_i\in  \mathbb{F}_s}s_i,
        \end{equation}
    where $ \mathbb{F}$ and $  \mathbb{F}^{*}_s$ are the original feature set and the final selected features respectively, $f_i$ is the $i^{th}$ feature of $ \mathbb{F}$, and $K$ is the number of features to be selected. In addition, by sorting all the features based on their strength, we will derive the importance of all features in the dataset. In short, we will be able to rank all input features by training just once a single sparse DAE model. 
    
    For a deeper understanding of the above process, we analyze the strength of each input neuron in a 2D map on the MNIST dataset. This is illustrated in Figure \ref{fig:weightdist}. At the beginning of training, all the neurons have small strength due to the random initialization of each weight to a small value. During the network evolution, stronger connections are linked to important features gradually. We can observe that after ten epochs, the neurons in the center of the map become stronger. This pattern is similar to the pattern of MNIST data in which most of the digits appear in the middle of the picture.
    
    We studied other metrics for estimating the neuron importance such as the strength of output neurons, degree of input and output neurons, and strength and degree of neurons simultaneously. However, in our experiments, all these methods have been outperformed by the strength of the input neurons in terms of accuracy and stability.
       

\section{Experiments}
     In order to verify the validity of our proposed method, we carry out several experiments. In this section, first, we state the settings of the experiments, including hyperparameters and datasets. Then, we perform feature selection with QuickSelection and compare the results with other methods, including MCFS, Laplacian Score, and three autoencoder-based feature selection methods. After that, we do different analyses on QuickSelection to understand its behavior. Finally, we discuss the scalability of QuickSelection and compare it with the other methods considered.
    
    \subsection{Settings}
    The experiment settings, including the values of hyperparameters, implementation details, the structure of the sparse DAE, datasets we use for evaluation, and the evaluation metric, are as follows.
    \subsubsection{Hyperparameters and Implementation}\label{ssec:settings}
    For feature selection, we consider the case of the simplest sparse DAE with one hidden layer consisting of 1000 neurons. This choice is made due to our main objective to decrease the model complexity and the number of parameters. The activation function used for the hidden and output layer neurons is \enquote{Sigmoid} and \enquote{Linear} respectively, except for the Madelon dataset where we use \enquote{Tanh} for the output activation function. We train the network with SGD and a learning rate of $0.01$. The hyperparameter $\zeta$, the fraction of weights to be removed in the SET procedure, is $0.2$. Also, $\epsilon$, which determines the sparsity level, is set to $13$. 
    We set the noise factor ($\mathit{nf}$) to 0.2 in the experiments. To improve the learning process of our network, we standardize the features of our dataset such that each attribute has zero mean and unit variance  . However, for SMK and PCMAC datasets, we use Min-Max scaling. The preprocessing method for each dataset is determined with a small experiment of the two preprocessing method. 
    
    We implement sparse DAE and QuickSelection\footnote{The implementation of QuickSelection is available at:      \url{https://github.com/zahraatashgahi/QuickSelection}} in a purely sparse manner in Python, using the Scipy library \cite{jones2001scipy} and Cython. We compare our proposed method to MCFS, Laplacian score (LS), AEFS, and CAE, which have been mentioned in Section \ref{ssec:related_work_feature_selection}. We also performed some experiments with UDFS; however, since we were not able to obtain many of the results in the considered time limit (24 hours), we do not include the results in the paper. We have used the scikit-feature repository for the implementation of MCFS, and Laplacian score \cite{li2018feature}. Also, we use the implementation of feature selection with CAE and AEFS from Github\footnote{The implementation of AEFS and CAE is available at: \url{https://github.com/mfbalin/Concrete-Autoencoders}}. In addition, to highlight the advantages of using sparse layers, we compare our results with a fully-connected autoencoder (FCAE) using the neuron strength as a measure of the importance of each feature. To have a fair comparison, the structure of this network is considered similar to our DAE, one hidden layer containing 1000 neurons implemented using TensorFlow. Furthermore, we have studied the effect of other components of QuickSelection, including input denoising and SET training algorithm, in Appendix \ref{app_ssec:appendix_select_noise_factor} and \ref{app_sec:appendix_sparse_training}, respectively. 
    
    For all the other methods (except FCAE for which all the hyperparameters and preprocessing are similar to QuickSelection), we scaled the data between zero and one, since it yields better performance than data standardization for these methods. The hyperparameters of the aforementioned methods have been set similar to the ones reported in the corresponding code or paper. For AEFS, we tuned the regularization hyperparameter between $0.0001$ and $1000$, since this method is sensitive to this value. We perform our experiments on a single CPU core, Intel Xeon Processor E5 v4, and for the methods that require GPU, we use NVIDIA TESLA P100.

    \subsubsection{Datasets} \label{sssec:datasets}
  
    We evaluate the performance of our proposed method on eight datasets, including five low-dimensional datasets and three high-dimensional ones. Table \ref{tab:datasets} illustrates the characteristics of these datasets. 
    \begin{itemize}
        \item \textbf{COIL-20} \cite{nene1996columbia} consists of $1440$ images taken from $20$ objects ($72$ poses for each object).
        \item \textbf{Madelon} \cite{guyon2008feature} is an artificial dataset with 5 informative features and 15 linear combinations of them. The rest of the features are distractor features since they have no predictive power. 
        \item \textbf{Human Activity Recognition} (HAR) \cite{anguita2013public} is created by collecting the observations of $30$ subjects performing $6$ activities such as walking, standing, and sitting. The data was recorded by a smart-phone connected to the subjects' body.
        \item \textbf{Isolet} \cite{fanty1991spoken} has been created with the spoken name of each letter of the English alphabet.
        \item \textbf{MNIST} \cite{lecun1998mnist} is a database of 28x28 images of handwritten digits.
        \item \textbf{SMK-CAN-187} \cite{spira2007airway} is a gene expression dataset with 19993 features. This dataset compares smokers with and without lung cancer.
        \item \textbf{GLA-BRA-180} \cite{sun2006neuronal} consists of the expression profile of Stem cell factor useful to determine tumor angiogenesis. 
        \item \textbf{PCMAC} \cite{lang1995newsweeder} is a subset of the 20 Newsgroups data. 
        \end{itemize}
       \begin{table}[!t]
        \centering
        \caption{Datasets characteristics.}
        \label{tab:datasets}
        \begin{scriptsize}
        \begin{tabularx}{0.92\textwidth}{cccccccl}
            \toprule
            \multicolumn{1}{c}{ \textbf{Dataset}} & \textbf{Dimensions} & \textbf{Type} & \textbf{Samples} & \textbf{Train} & \textbf{Test} & \textbf{Classes} &  \\\midrule
            Coil20 & 1024 & Image & 1440 & 1152 & 288 & 20 &  \\ 
            Isolet & 617 & Speech & 7737 & 6237 & 1560 & 26 &  \\ 
            HAR & 561 & Time Series & 10299 & 7352 & 2947 & 6 &  \\ 
            Madelon & 500 & Artificial & 2600 & 2000 & 600 & 2 &  \\ 
            MNIST & 784 & Image & 70000 & 60000 & 10000 & 10 &  \\ 
            SMK-CAN-187 & 19993 & Microarray & 187 & 149 & 38 & 2 &  \\ 
            GLA-BRA-180 & 49151 & Microarray & 180 & 144 & 36 & 4 &  \\ 
            PCMAC & 3289 & Text & 1943 & 1554 & 389 & 2 &  \\ 
            \bottomrule
            \end{tabularx}
            \end{scriptsize}
        \end{table}
    
    \subsubsection{Evaluation Metrics}
    To evaluate our model, we compute two metrics: clustering accuracy and classification accuracy. To derive clustering accuracy \cite{li2018feature}, first, we perform K-means using the subset of the dataset corresponding to the selected features and get the cluster labels. Then, we find the best match between the class labels and the cluster labels and report the clustering accuracy. We repeat the K-means algorithm 10 times and report the average clustering results since K-means may converge to a local optimal. 
    
    To compute classification accuracy, we use a supervised classification model named \enquote{Extremely randomized trees} (ExtraTrees), which is an ensemble learning method that fits several randomized decision trees on different parts of the data  \cite{geurts2006extremely}. The choice of the classification method is made due to the computational-efficiency of the ExtraTrees classifier. To compute classification accuracy, first, we derive the $K$ selected features using each feature selection method considered. Then, we train the ExtraTrees classifier with $50$ trees as estimators on the $K$ selected features of the training set. Finally, we compute the classification accuracy on the unseen test data. For the datasets that do not contain a test set, we split the data into training and testing sets, including $80\%$ of the total original samples for the training set and the remaining $20\%$ for the testing set. In addition, we have evaluated the classification accuracy of feature selection using the random forest classifier \cite{liaw2002classification} in Appendix \ref{app_sec:appendix_classification_RF}.

    \subsection{Feature Selection} \label{ssec:feature_selection}
    
    We select 50 features from each dataset except Madelon, for which we select just 20 features since most of its features are non-informative noise. Then, we compute the clustering and classification accuracy on the selected subset of features; the more informative features selected, the higher accuracy will be achieved. The clustering and classification accuracy results of our model and the other methods is summarized in Tables \ref{tab:clustering_acc} and \ref{tab:classification_acc}, respectively. These results are an average of 5 runs for each case. For the autoencoder-based feature selection methods, including CAE, AEFS, and FCAE, we consider 100 training epochs. However, we present the results of QuickSelection at epoch 10 and 100 named QuickSelection\textsubscript{10} and QuickSelection\textsubscript{100}, respectively. This is mainly due to the fact that our proposed method is able to achieve a reasonable accuracy after the first few epochs. Moreover, we perform hyperparameter tuning for $\epsilon$ and $\zeta$ using the grid search method over a limited number of values for all datasets; the best result is presented in Table \ref{tab:clustering_acc} and \ref{tab:classification_acc} as QuickSelection\textsubscript{best}. The results of hyperparameters selection can be found in Appendix \ref{app_ssec:appendix_select_set_parameter}. However, we do not perform hyperparameter optimization for the other methods (except AEFS). Therefore, in order to have a fair comparison between all methods, we do not compare the results of QuickSelection\textsubscript{best} with the other methods.   
    
    \begin{table}[!t]
        \centering
        \caption{Clustering accuracy (\%) using 50 selected features (except Madelon for which we select 20 features). On each dataset, the bold entry is the best-performer, and the italic one is the second-best performer.}
        \label{tab:clustering_acc}
        \begin{center}
        \begin{scriptsize}
        
        \begin{tabular}{c@{\hskip 0.07in}c@{\hskip 0.07in}c@{\hskip 0.07in}c@{\hskip 0.07in}c@{\hskip 0.07in}c@{\hskip 0.07in}c@{\hskip 0.07in}c@{\hskip 0.07in}c}
            \toprule
            \textbf{Method} &\textbf{COIL-20} &\textbf{Isolet}&\textbf{HAR} &\textbf{Madelon} &\textbf{MNIST} &\textbf{SMK}&\textbf{GLA}&\textbf{PCMAC}\\ 
            \midrule
            MCFS      & \textbf{67.0$\pm$0.7} & \textit{33.8$\pm$0.5} & \textbf{62.4$\pm$0.0} & 57.2$\pm$0.0 & 35.2$\pm$0   & 51.6$\pm$0.2 & \textbf{65.8$\pm$0.3} & 50.6$\pm$0.0 \\
             LS & 55.5$\pm$0.4 & 33.2$\pm$0.2 & \textit{61.2$\pm$0.0} & \textit{58.1$\pm$0.0} & 14.9$\pm$0.1 & 51.6$\pm$0.4 & 55.5$\pm$0.4 & 50.6$\pm$0.0 \\
             CAE  & 60.0$\pm$1.1 & 31.6$\pm$1.3 & 51.4$\pm$0.4 & 56.9$\pm$3.6 & \textbf{49.2$\pm$1.5} & \textbf{60.7$\pm$0.4} & 55.4$\pm$1.3 & \textit{52.0$\pm$1.2} \\
             AEFS & 51.2$\pm$1.7 & 31.0$\pm$2.7 & 55.0$\pm$2.2 & 50.8$\pm$0.2 & 40.0$\pm$1.9 & 52.4$\pm$1.8 & 56.1$\pm$5.2 & 50.9$\pm$0.5 \\
             FCAE  & \textit{60.2$\pm$1.7} & 28.7$\pm$2.5 & 49.5$\pm$8.7 & 50.9$\pm$0.4 & 28.2$\pm$8.5 & 51.5$\pm$0.8 & 53.5$\pm$3.0 & 50.9$\pm$0.1 \\\midrule
             QS$_{10}$     & 59.5$\pm$2.1 & 32.5$\pm$2.8 & 56.0$\pm$2.6 & 57.5$\pm$3.8 & 45.4$\pm$3.9 & \textit{54.0$\pm$3.1} & 53.6$\pm$4.7 & 50.9$\pm$0.5 \\
             QS$_{100}$    & \textit{60.2$\pm$2.0} & \textbf{35.1$\pm$2.7} & 54.6$\pm$4.5 & \textbf{58.2$\pm$1.5} & \textit{48.3$\pm$2.4} & 51.8$\pm$0.8 & \textit{59.5$\pm$1.8} & \textbf{52.5$\pm$1.1} \\
             QS$_{best}$ & 63.8$\pm$1.5 & 42.2$\pm$2.6  &59.5$\pm$4.3   & 58.6$\pm$0.9 &48.3 $\pm$2.4 &54.9$\pm$1.39 & 59.5$\pm$1.8 &53.1$\pm$0 \\
            \bottomrule
            \end{tabular}
        
        \end{scriptsize}
        \end{center}
        \end{table}
    
    \begin{table}[!t]
        \centering
        \caption{Classification accuracy (\%) using 50 selected features (except Madelon for which we select 20 features). On each dataset, the bold entry is the best-performer, and the italic one is the second-best performer.}
        \label{tab:classification_acc}
        \begin{center}  
        \begin{scriptsize}
           \begin{tabular}{c@{\hskip 0.07in}c@{\hskip 0.07in}c@{\hskip 0.07in}c@{\hskip 0.07in}c@{\hskip 0.07in}c@{\hskip 0.07in}c@{\hskip 0.07in}c@{\hskip 0.07in}c}
            \toprule
            \textbf{Method}&\textbf{COIL-20} &\textbf{Isolet}&\textbf{HAR} &\textbf{Madelon} &\textbf{MNIST} &\textbf{SMK}&\textbf{GLA}&\textbf{PCMAC}\\ 
            \midrule
             MCFS      & 99.2$\pm$0.3 & 79.5$\pm$0.4 & 88.9$\pm$0.3 & 81.7$\pm$0.8 & 88.7$\pm$0    & 75.8$\pm$1.5 & 70.6$\pm$3.8 & 55.5$\pm$0.0 \\
             LS & 89.8$\pm$0.4 & 83.0$\pm$0.2 & 86.4$\pm$0.4 & \textbf{91.4$\pm$0.9} & 20.7$\pm$0.1  & 71.6$\pm$5.6 & 71.7$\pm$1.1 & 50.4$\pm$0.0 \\
             CAE  & \textit{99.6$\pm$0.3} & \textbf{89.8$\pm$0.6} & \textbf{91.7$\pm$1.0} & 87.5$\pm$2.0 & \textbf{95.4$\pm$0.1}  & 71.6$\pm$3.1 & 70.0$\pm$4.1 & \textbf{59.9$\pm$1.5} \\
             AEFS& 93.0$\pm$2.7 & 85.1$\pm$2.4 & 87.7$\pm$1.4 & 52.1$\pm$2.8 & 86.1$\pm$2.0  & \textit{76.3$\pm$4.4} & 68.9$\pm$3.7 & 57.1$\pm$3.6 \\
             FCAE  & \textbf{99.7$\pm$0.2} & 81.6$\pm$5.9 & 87.4$\pm$2.4 & 53.5$\pm$8.1 & 68.8$\pm$28.7 & 71.6$\pm$3.5 & \textit{72.8$\pm$4.8} & 58.1$\pm$1.9 \\\midrule
             QS$_{10}$     & 98.8$\pm$0.6 & 86.9$\pm$1.1 & 88.8$\pm$0.7 & 86.6$\pm$3.6 & \textit{93.8$\pm$0.6}  & \textbf{76.9$\pm$4.6} & 69.4$\pm$3.0 & \textit{58.9$\pm$4.4} \\
             QS$_{100}$    & \textbf{99.7$\pm$0.3} & \textit{89.0$\pm$1.3} & \textit{90.2$\pm$1.2} & \textit{90.3$\pm$0.7} & 93.5$\pm$0.5  & 75.7$\pm$3.9 & \textbf{73.3$\pm$3.3} & 58.0$\pm$2.9 \\
             QS$_{best}$ & 99.7$\pm$0.3 & 89.0$\pm$1.3   &90.5$\pm$1.6   &90.9 $\pm$0.5 &94.2$\pm$0.5 &81.6 $\pm$2.9& 73.3$\pm$3.3 &61.3$\pm$6.1 \\
            \bottomrule
            \end{tabular}
 
        \end{scriptsize} 
        \end{center}
        \end{table}
        

    From Table \ref{tab:clustering_acc}, it can be observed that QuickSelection outperforms all the other methods on Isolet, Madelon, and PCMAC, in terms of clustering accuracy, while being the second-best performer on Coil20, MNIST, SMK, and GLA. Furthermore, On the HAR dataset, it is the best performer among all the autoencoder-based feature selection methods considered. 
    As shown in Table \ref{tab:classification_acc}, QuickSelection outperforms all the other methods on Coil20, SMK, and GLA, in terms of classification accuracy, while being the second-best performer on the other datasets. 
    From these Tables, it is clear that QuickSelection can outperform its equivalent dense network (FCAE) in terms of classification and clustering accuracy on all datasets. 
    
    It can be observed in Tables \ref{tab:clustering_acc} and \ref{tab:classification_acc}, that Lap\_score has a poor performance when the number of samples is large (e.g. MNIST). However, in the tasks with a low number of samples and features, even on noisy environments such as Madelon, Lap\_score has a relatively good performance. In contrast, CAE has a poor performance in noisy environments (e.g., Madelon), while it has a decent classification accuracy on the other datasets considered. It is the best or second-best performer on five datasets, in terms of classification accuracy, when $K=50$. AEFS and FCAE cannot achieve a good performance on Madelon, either. We believe that the dense layers are the main cause of this behaviour; the dense connections try to learn all input features, even the noisy features. Therefore, they fail to detect the most important attributes of the data. MCFS performs decently on most of the datasets in terms of clustering accuracy. This is due to the main objective of MCFS to preserve the multi-cluster structure of the data. However, this method also has a poor performance on the datasets with a large number of samples (e.g., MNIST) and noisy features (e.g., Madelon).
    
    However, since evaluating the methods using a single value of $K$ might not be enough for comparison, we performed another experiment using different values of $K$. In Appendix \ref{app_sec:appendix_running_time}, we test other values for $K$ on all datasets, and compare the methods in terms of classification accuracy, clustering accuracy, running time, and maximum memory usage. The summary of the results of this Appendix has been summarized in Section \ref{sec:discussion_speed_memory}.

    \begin{figure*}[!t]
        \centering
        \includegraphics[width=12cm]{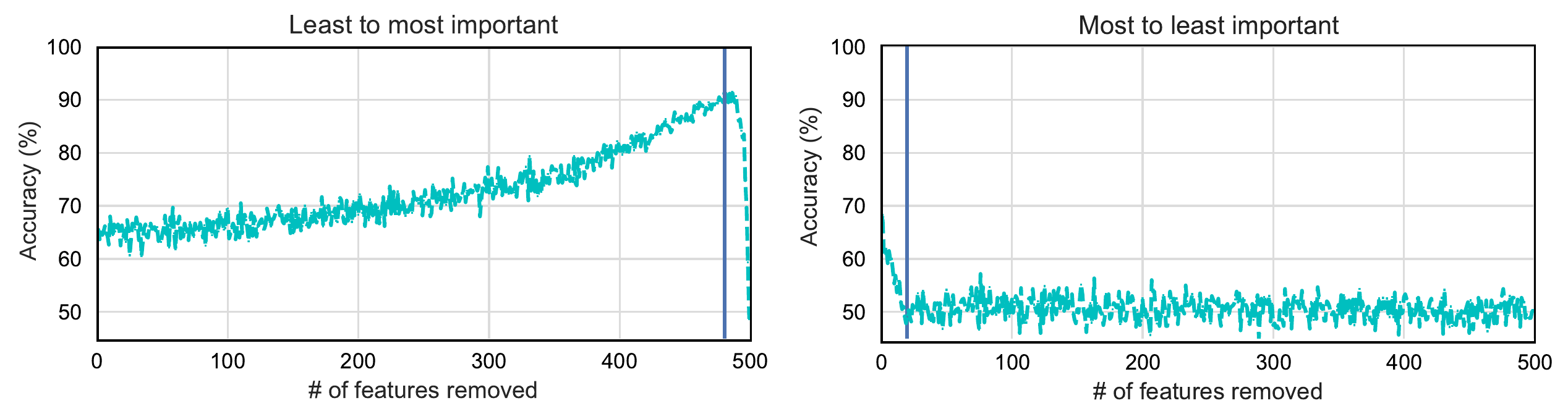}
        \caption{Influence of feature removal on Madelon dataset. After deriving the importance of the features with QuickSelection, we sort and then remove them based on the above two methods.}
        \label{fig:madelon}
    \end{figure*}
    
    \subsubsection{Relevancy of Selected Features}
    To illustrate the ability of QuickSelection in finding informative features, we analyze thoroughly the Madelon dataset results, which has the interesting property of containing many noisy features. We perform the following experiments; first, we sort the features based on their strength. Then, we remove the features one by one from the least important feature to the most important one. In each step, we train an ExtraTrees classifier with the remained features. We repeat this experiment by removing the feature from the most important ones to the least important ones. The result of classification accuracy for both experiments can be seen in Figure \ref{fig:madelon}. On the left side of Figure \ref{fig:madelon}, we can observe that removing the least important features, which are noise, increases the accuracy. The maximum accuracy occurs after we remove 480 noise features. This corresponds to the moment when all the noise features are supposed to be removed. In Figure \ref{fig:madelon} (right), it can be seen that removing the features in a reverse order results in a sudden decrease in the classification accuracy. After removing 20 features (indicated by the vertical blue line), the classifier performs like a random classifier. We conclude that QuickSelection is able to find the most informative features in good order.
     
\begin{figure}[!t]
        \centering
        \centering
        \includegraphics[width=0.9\columnwidth]{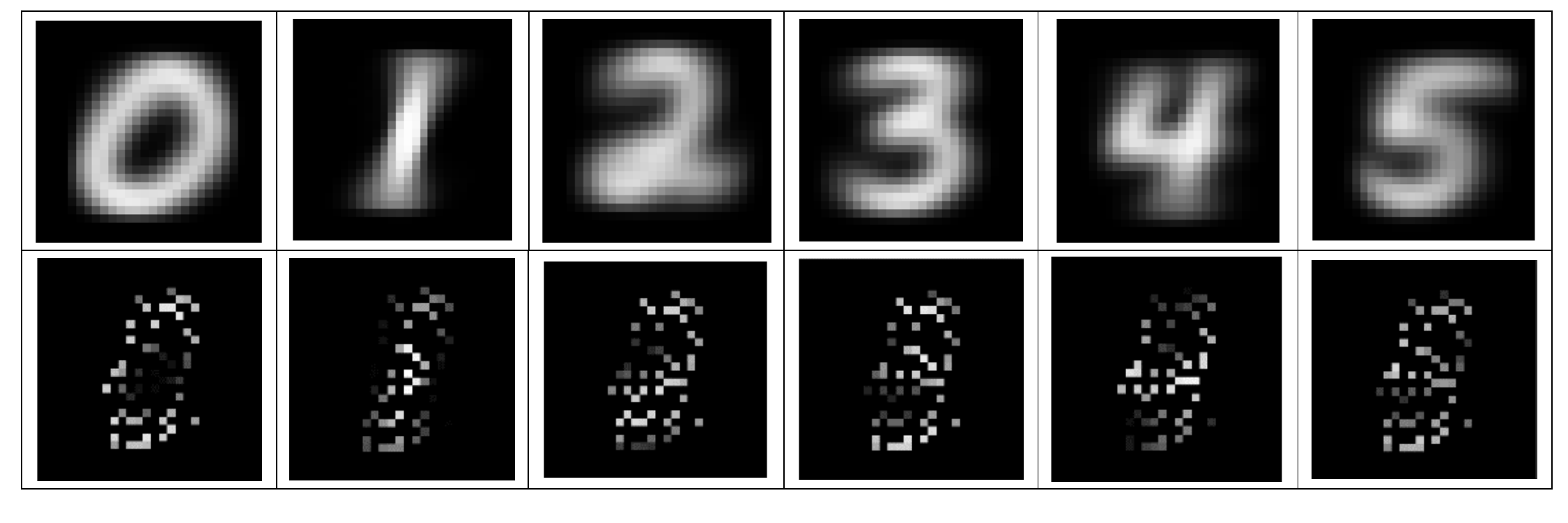}
        \captionof{figure}{Average values of all data samples of each class corresponding to the 50 selected features on MNIST after 100 training epochs (bottom), along with the average of the actual data samples of each class (top). }
        \label{fig:mnist059}
        \end{figure}

    To better show the relevancy of the features found by QuickSelection, we visualize the 50 features selected on the MNIST dataset per class, by averaging their corresponding values from all data samples belonging to one class. As can be observed in Figure \ref{fig:mnist059}, the resulting shape resembles the actual samples of the corresponding digit. We discuss the results of all classes at different training epochs in more detail in Appendix \ref{app_sec:mnist_features}.
    
\section{Discussion}  
    \subsection{Accuracy and Computational Efficiency Trade-off} \label{sec:discussion_speed_memory} 
    In this section, we perform a thorough comparison between the models in terms of running time, energy consumption, memory requirement, clustering accuracy, and classification accuracy. In short, we change the number of features to be selected ($K$) and measure the accuracy, running time, and maximum memory usage across all methods. Then, we compute two scores to summarize the results and compare methods.
    
    We analyse the effect of changing $K$ on QuickSelection performance and compare with other methods; the results are presented in Figure \ref{fig:runningt} in Appendix \ref{app_sec:appendix_running_time}. Figure \ref{fig:runningt_1} compares the performance of all methods when $K$ is changing between 5 and 100 on low-dimensional datasets, including Coil20, Isolet, HAR, and Madelon. Figure \ref{fig:runningt_2} illustrates performance comparison for $K$ between 5 and 300 on the MNIST dataset, which is also a low-dimensional dataset. We discuss this dataset separately since it has a large number of samples that makes it different from other low-dimensional datasets. Figure \ref{fig:runningt_3} represents a similar comparison on three high-dimensional datasets, including SMK, GLA, and PCMAC. It should be noted that to have a fair comparison, we use a single CPU core to run these methods; however, since the implementations of CAE and AEFS are optimized for parallel computation, we use a GPU to run these methods. We also measure the running time of feature selection with CAE on CPU.
    
    To compare the memory requirement of each method, we profile the maximum memory usage during feature selection for different values of $K$. The results are presented in Figure \ref{fig:memory} in Appendix \ref{app_sec:appendix_running_time}, derived using a Python library named resource\footnote{https://docs.python.org/2/library/resource.html}. Besides, to compare memory occupied by the autoencoder-based models, we count the number of parameters for each model. The results are shown in Figure \ref{fig:num_params} in the Appendix \ref{app_sec:appendix_running_time_num_params}.

    \begin{figure}
        \begin{subfigure}[t]{0.325\textwidth}
            \includegraphics[width=\textwidth]{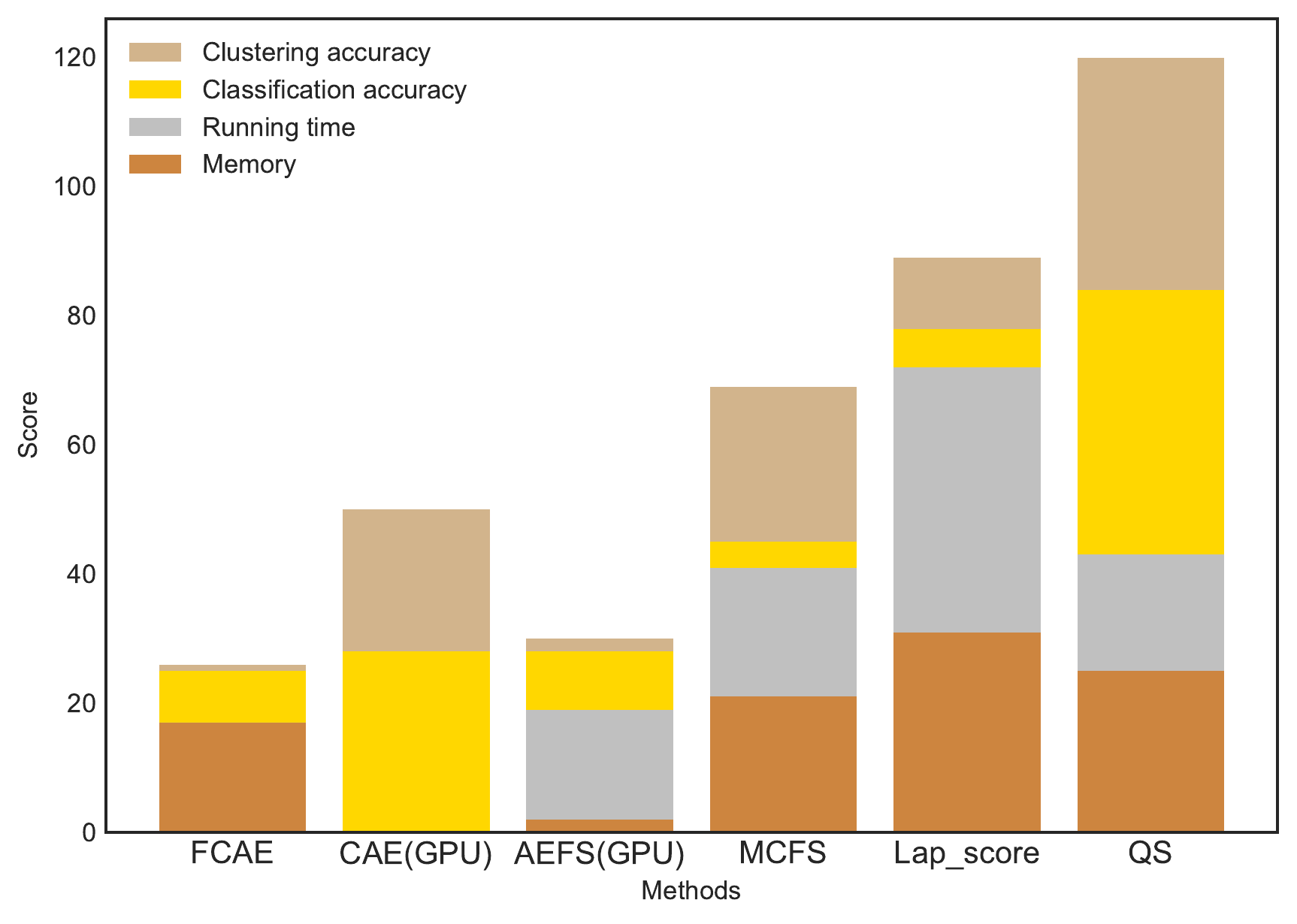}
            \caption{Score 1. Scores are given based on the ranking of the methods.}
            \label{fig:performance}
        \end{subfigure}%
        \hspace*{\fill}   
        \begin{subfigure}[t]{0.31\textwidth}
            \includegraphics[width=\textwidth]{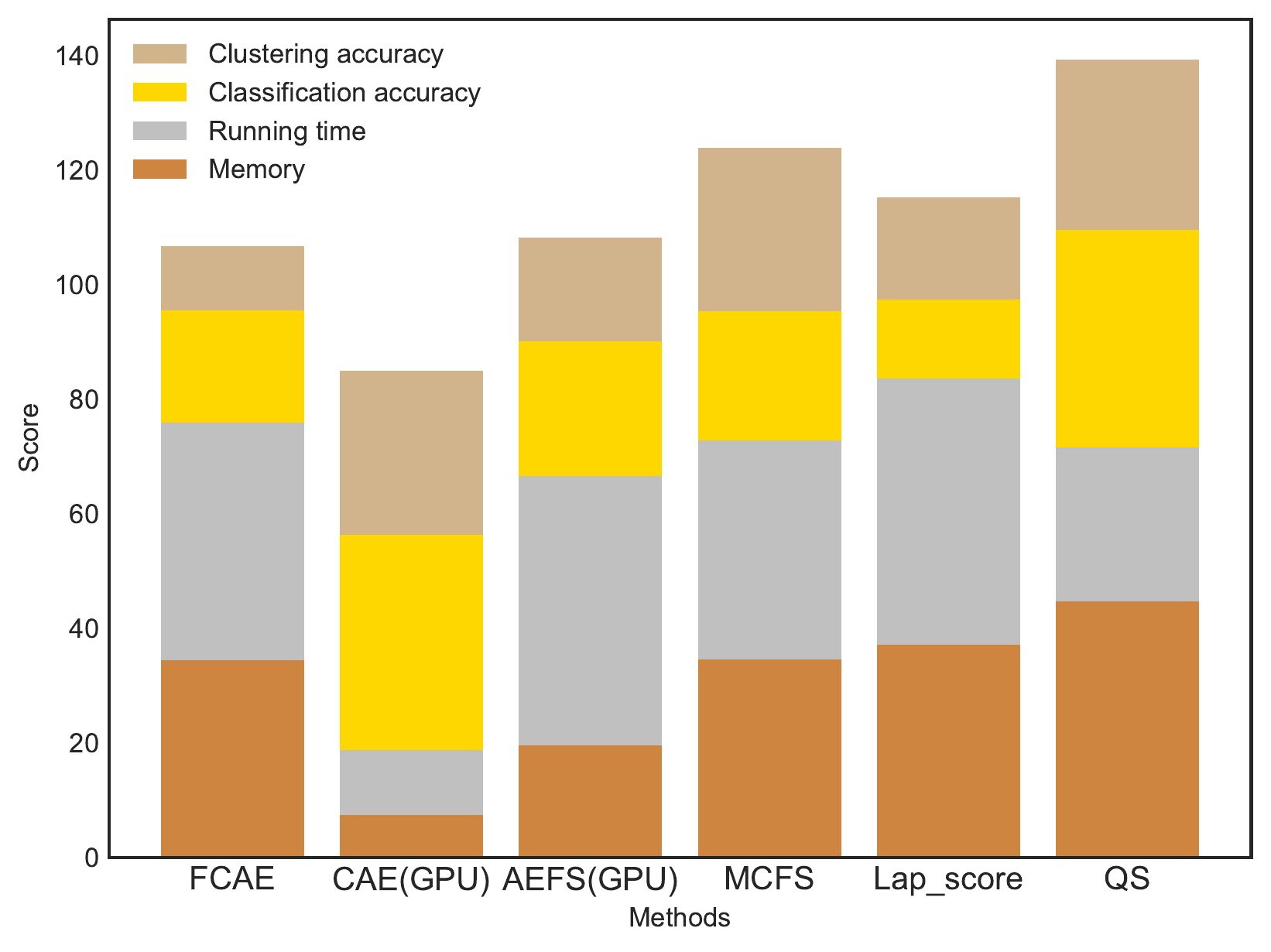}
            \caption{Score 2. Scores are given based on the normalized value of each objective.}
            \label{fig:performance_normalized}
        \end{subfigure}%
         \hspace*{\fill}   
        \begin{subfigure}[t]{0.33\textwidth}
            \includegraphics[width=\textwidth]{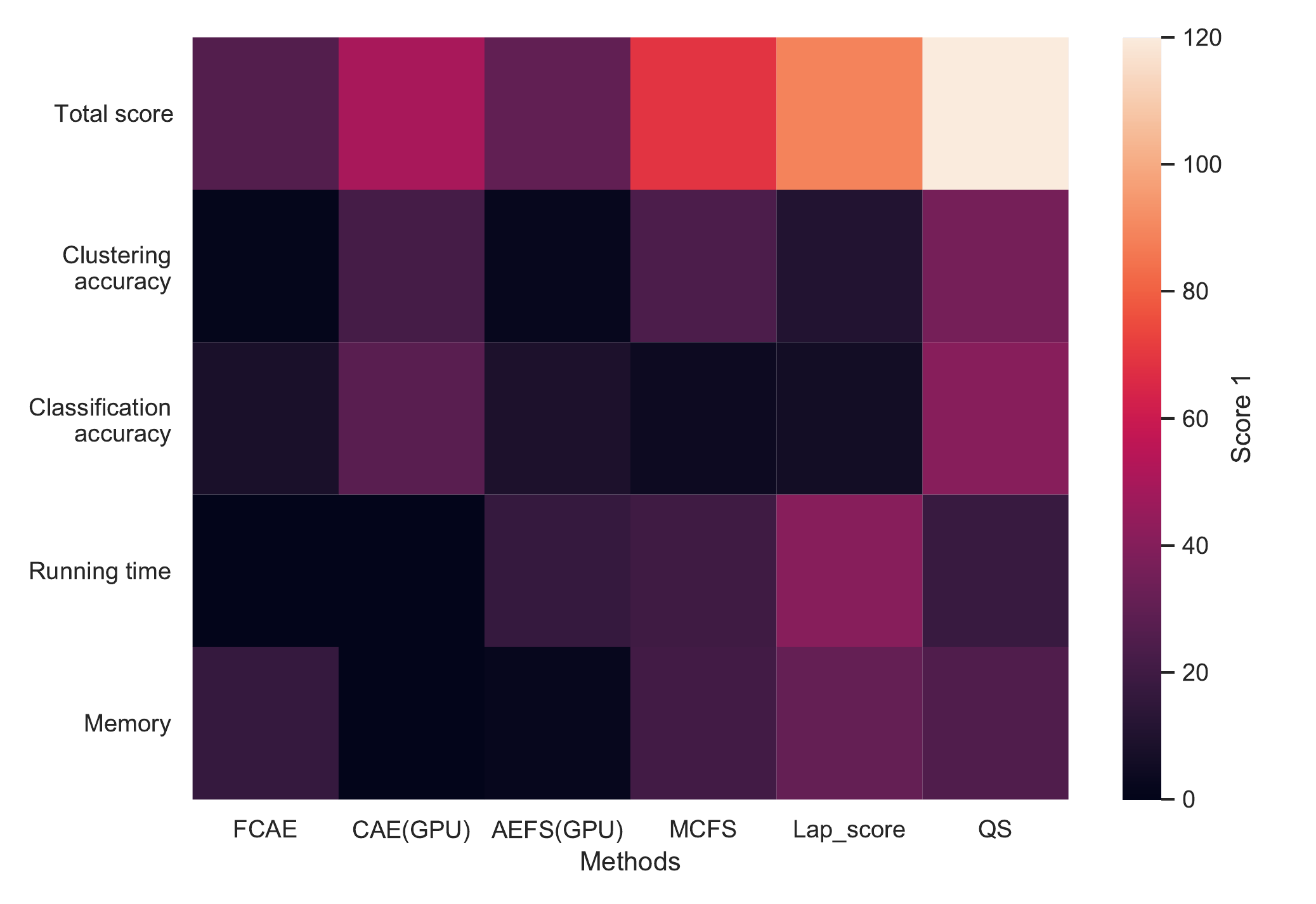}
            \caption{Heat-map visualization of Score 1.}
            \label{fig:performance_heatmap}
        \end{subfigure}%

    \caption{Feature selection comparison in terms of classification accuracy, clustering accuracy, speed, and memory requirement, on each dataset and for different values of $K$, using two scoring variants.} \label{fig:1}
    \end{figure}
    
    
    However, comparing all of these methods only by looking into the graphs in Figure \ref{fig:runningt} and Figure \ref{fig:memory} is not easily possible, and the trade-off between the factors is not clear. For this reason, we compute two scores to take all these metrics into account simultaneously. 
    
    \textbf{\textit{Score 1.}} To compute this score, on each dataset and for each value of $K$, we rank the methods based on the running time, memory requirement, clustering accuracy, and classification accuracy. Then, we give a score of $1$ to the best and second-best performer; this is mainly due to the fact that in most cases, the difference between these two is negligible. After that, we compute the summation of these scores for each method on all datasets. The results are presented in Figure \ref{fig:performance}; to ease the comparison of different components in the score, a heat-map visualization of the scores is presented in Figure \ref{fig:performance_heatmap}. The cumulative score for each method consists of four parts that correspond to each metric considered. As it is obvious in this Figure, QuickSelection (cumulative score of QuickSelection$_{10}$ and QuickSelection$_{100}$) outperforms all other methods by a significant gap. Our proposed method is able to achieve the best trade-off between accuracy, running time, and memory usage, among all these methods. Laplacian score, the second-best performer, has a decent performance in terms of running time and memory, while it cannot perform well in terms of accuracy. On the other hand, CAE has a satisfactory performance in terms of accuracy. However, it is not among the best two performers in terms of computational resources for any values of $K$. Finally, FCAE and AEFS cannot achieve a decent performance compared to the other methods.
    A more detailed version of Figure \ref{fig:performance} is available in Figure \ref{fig:performance_all} in Appendix \ref{app_sec:appendix_running_time}.
    
    \textbf{\textit{Score 2.}} In addition to the raking-based score, we calculate another score to consider all the methods, even the lower-ranking ones. With this aim, on each dataset and value of $K$, we normalize each performance metric between $0$ and $1$, using the values of the best performer and worst performer on each metric. The value of $1$ in the accuracy score means the highest accuracy. However, for the memory and running time, the value of $1$ means the least memory requirement and the least running time, respectively. After normalizing the metrics, we accumulate the normalized values for each method and on all datasets. The results are depicted in Figure \ref{fig:performance_normalized}. As can be seen in this diagram, QuickSelection (we consider the results of QuickSelection$_{100}$) outperforms the other methods by a large margin. CAE has a close performance to QuickSelection in terms of both accuracy metrics, while it has a poor performance in terms of memory and running time. In contrast, Lap\_score is computationally efficient while having the lowest accuracy score. In summary, it can be observed in Figure \ref{fig:performance_normalized}, that QuickSelection achieves the best trade-off of the four objectives among the considered methods.
    
    \textbf{Energy Consumption.} The next analysis we perform concerns the energy consumption of each method. We estimate the energy consumption of each method using the running time of the corresponding algorithm for each dataset and value of $K$. We assume that each method uses the maximum power of the corresponding computational resources during its running time. Therefore, we derive the power consumption of each method, using the running time and maximum power consumption of CPU and/or GPU, which can be found within the specification of the corresponding  CPU or GPU model. As shown in Figure \ref{fig:power} in Appendix \ref{app_sec:appendix_running_time_power}, the Laplacian score feature selection needs the least amount of energy among the methods on all datasets except the MNIST dataset. QuickSelection$_{10}$ is the best performer on MNIST in terms of energy consumption. Laplacian score and MCFS are sensitive to the number of samples. They cannot perform well on MNIST, either in terms of accuracy or efficiency. The maximum memory usage during feature selection for Laplacian score and MCFS on MNIST is 56 GB and 85 GB, respectively. Therefore, they are not a good choice in case of having a large number of samples. QuickSelection is the second-best performer in terms of energy consumption, and also the best performer among the autoencoder-based methods. QuickSelection is not sensitive to the number of samples or the number of dimensions.   
    
\begin{figure}[!t]    
        \centering
        \includegraphics[width=\textwidth]{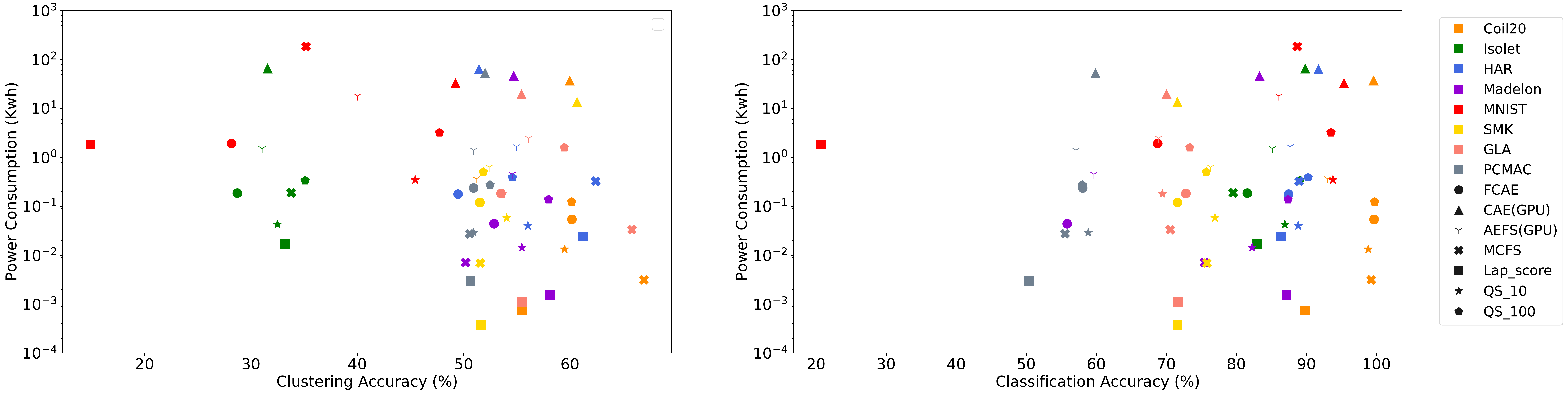} 
        \caption{Estimated power consumption (Kwh) vs. accuracy (\%) when selecting 50 features (except Madelon for which we select 20 features). Each point refers to the result of a single dataset (specified by colors) and method (specified by markers) where the x and y-axis show the accuracy and the estimated power consumption, respectively.}
        \label{fig:tradeoff1}
        \end{figure}

\begin{figure}[!t]    
        \centering
        \includegraphics[width=\textwidth]{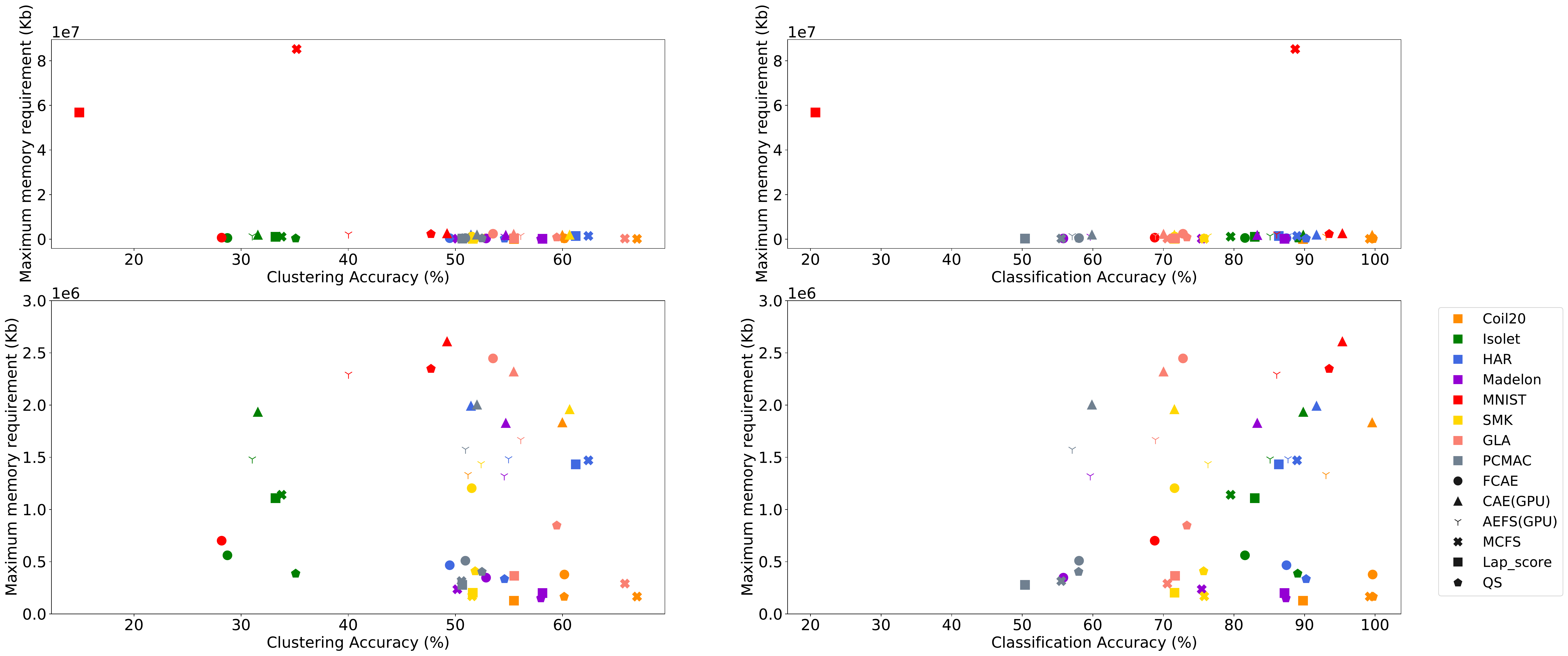} 
        \caption{Maximum memory requirement (Kb) vs. accuracy (\%) when selecting 50 features (except Madelon for which we select 20 features). Each point refers to the result of a single dataset (specified by colors) and method (specified by markers) where the x and y-axis show the accuracy and the maximum memory requirement, respectively. Due to the high memory requirement of MCFS and Lap\_score on the MNIST dataset which makes it difficult to compare the other results (upper plots), we zoom in this section in the bottom plots.}
        \label{fig:tradeoff2}
        \end{figure}

    \textbf{Efficiency vs Accuracy.}  In order to study the trade-off between accuracy and resource efficiency, we perform another in-depth analysis. In this analysis, we plot the trade-off between accuracy (including, classification and clustering accuracy) and resource requirement (including, memory and energy consumption). The results are shown in Figures \ref{fig:tradeoff1} and \ref{fig:tradeoff2} that correspond to the energy-accuracy and memory-accuracy trade-off, respectively. Each point in these plots refers to the results of a particular combination between a specific method and dataset when selecting 50 features (except Madelon, for which we select 20 features). As can be observed in these plots, QuickSelection, MCFS, and Lap\_score usually have a good trade-off between the considered metrics. A good trade-off between a pair of metrics is to maximize the accuracy (classification or clustering accuracy) while minimizing the computational cost (power consumption or memory requirement). However, when the number of samples increases (on the MNIST dataset), both MCFS and Lap\_score fail to maintain a low computational cost and high accuracy. Therefore, when the dataset size increases, these two methods are not an optimal choice. Among the autoencoder-based methods, in most cases QuickSelection\textsubscript{10} and QuickSelection\textsubscript{100} are among the Pareto optimal points. Another significant advantage of our proposed method is that it gives the ranking of the features as the output. Therefore, unlike the MCFS or CAE that need the value of $K$ as their input, QuickSelection is not dependent on $K$ and needs just a single training of the sparse DAE model for any values of $K$. Therefore, the computational cost of QuickSelection is the same for all values of $K$, and only a single run of this algorithm is required to get the hierarchical importance of features.
    
    \begin{figure}[t]
    \centering

    \begin{minipage}[t]{\textwidth}
      \centering
      \includegraphics[width=0.6\linewidth]{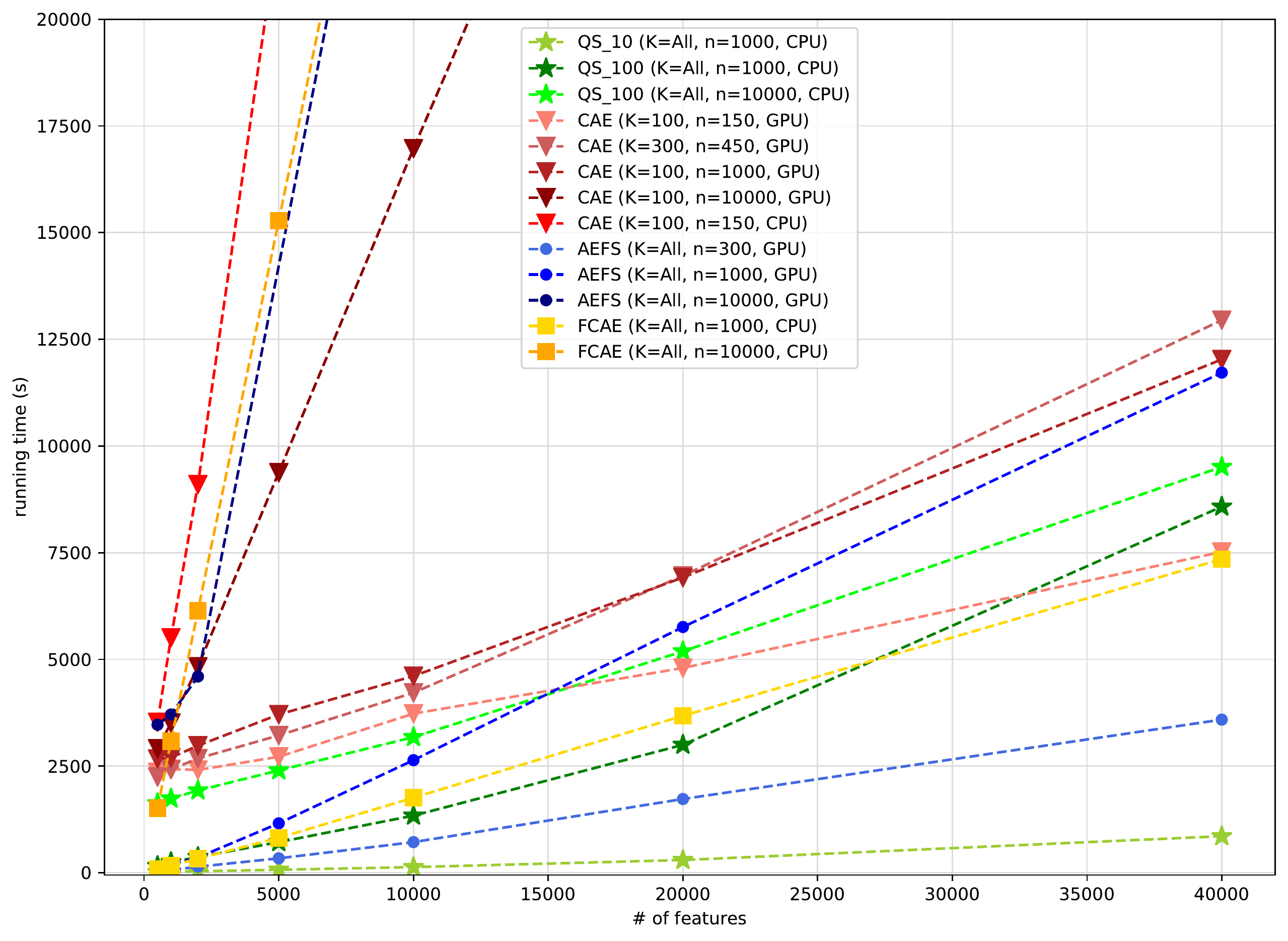}
      \captionof{figure}{Running time comparison on an artificially generated dataset. The features are generated using a standard normal distribution and the number of samples for each case is 5000.}
      \label{fig:running_t_random}
    \end{minipage}
    \end{figure}

    \subsection{Running Time Comparison on an Artificially Generated Dataset} \label{app_sec:random_running}
    
    In this section, we perform a comparison of the running time of the autoencoder-based feature selection methods on an artificially generated dataset. Since on the benchmark datasets both the number of features and samples are different, it is not easily possible to compare clearly the efficiency of the methods. This experiment aims at comparing the models real wall-clock training time in a controlled environment with respect to the number of input features and hidden neurons. In addition, in Appendix \ref{app_sec:appendix_feature_selection_large}, we have conducted another experiment regarding evaluation of the methods on a very large artificial dataset, in terms of both computational resources and accuracy.
            
    In this experiment, we aim to compare the speed of QuickSelection versus other autoencoder-based feature selection methods for different numbers of input features. 
    We run all of them on an artificially generated dataset with various numbers of features and 5000 samples, for 100 training epochs (10 epochs for QuickSelection$_{10}$). The features of this dataset are generated using a standard normal distribution. In addition, we aim to compare the running time of different structures for these algorithms. The specifications of the network structure for each method, the computational resources used for feature selection, and the corresponding results can be seen in Figure \ref{fig:running_t_random}. 
    
    For CAE, we consider two different values of $K$. The structure of CAE depends on this value. CAE has two hidden layers including a concrete selector and a decoder that have $K$ and $1.5K$ neurons, respectively. Therefore, by increasing the number of selected features, the running time of the model will also increase. In addition, we consider the cases of CAE with $1000$ and $10000$ hidden neurons in the decoder layer (manually changed in the code) to be able to compare it with the other models. We also measure the running time of performing feature selection with CAE using only a single CPU core. It can be seen from Figure \ref{fig:running_t_random} that its running time is considerably high. The general structures of AEFS, QuickSelection, and FCAE are similar in terms of the number of hidden layers. They are basic autoencoders with a single hidden layer. For AEFS, we considered three structures with different numbers of hidden neurons, including 300, 1000, and 10000. Finally, for QuickSelection and FCAE, we consider two different values for the number of hidden neurons, including 1000 and 10000.
    
    It can be observed that the running time of AEFS with $1000$ and $10000$ hidden neurons using a GPU, is much larger than the running time of QuickSelection$_{100}$ with similar numbers of hidden neurons using only a single CPU core, respectively. The same pattern is also visible in the case of CAE with $1000$ and $10000$ hidden neurons. This pattern also repeats in the case of FCAE with $10000$ hidden neurons. The running time of FCAE with 1000 hidden neurons is approximately similar to QuickSelection$_{100}$. However, the difference between these two methods is more significant when we increase the number of hidden neurons to $10000$. This is mainly due to the fact that the difference between the number of parameters of QuickSelection and the other methods become much higher for large values of $K$. Besides, these observations depict that the running time of QuickSelection does not change significantly by increasing the number of hidden neurons.
    
    As we have also mentioned before, QuickSelection gives the ranking of the features as the output. Therefore, unlike CAE which should be run separately for different values of $K$, QuickSelection is not affected by the choice of K because it computes the importance of all features at the same time and after finishing the training.
    In short, QuickSelection$_{10}$ has the least running time among other autoencoder-based methods while being independent of the value of $K$. In addition, unlike the other methods, the running time of QuickSelection is not sensitive to the number of hidden neurons since the number of parameters is low even for a very large hidden layer.

    \subsection{Neuron Strength Analysis}\label{ssec:neuron_strength}
    In this section, we discuss the validity of neurons strength as a measure of the feature importance. We observe the evolution of the network during training to analyze how the neuron strength of important and unimportant neurons changes during training. 
    
    We argue that the most important features that lead to the highest accuracy of feature selection are the features corresponding to neurons with the highest strength. In a neural network, weight magnitude is a metric that shows the importance of each connection \cite{kavzoglu1998assessing}. This stems from the fact that weights with a small magnitude have small effect on the performance of the model. At the beginning of training, we initialize all connections to a small random value. Therefore, all the neurons have almost the same strength/importance. As the training proceeds, some connections grow to a larger value while some others are pruned from the network during the dynamic connections removal and regrowth of the SET training procedure. The growth of the stable connection weights demonstrates their significance in the performance of the network. As a result, the neurons connected to these important weights contain important information. In contrast, the magnitude of the weights connected to unimportant neurons gradually decreases until they are removed from the network. In short, important neurons receive connections with a larger magnitude. As a result, neuron strength, which is the summation of the magnitude of weights connected to a neuron, can be a measure of the importance of an input neuron and its corresponding feature.
    
\begin{figure}[!t]    
        \centering
        \includegraphics[width=0.9\textwidth]{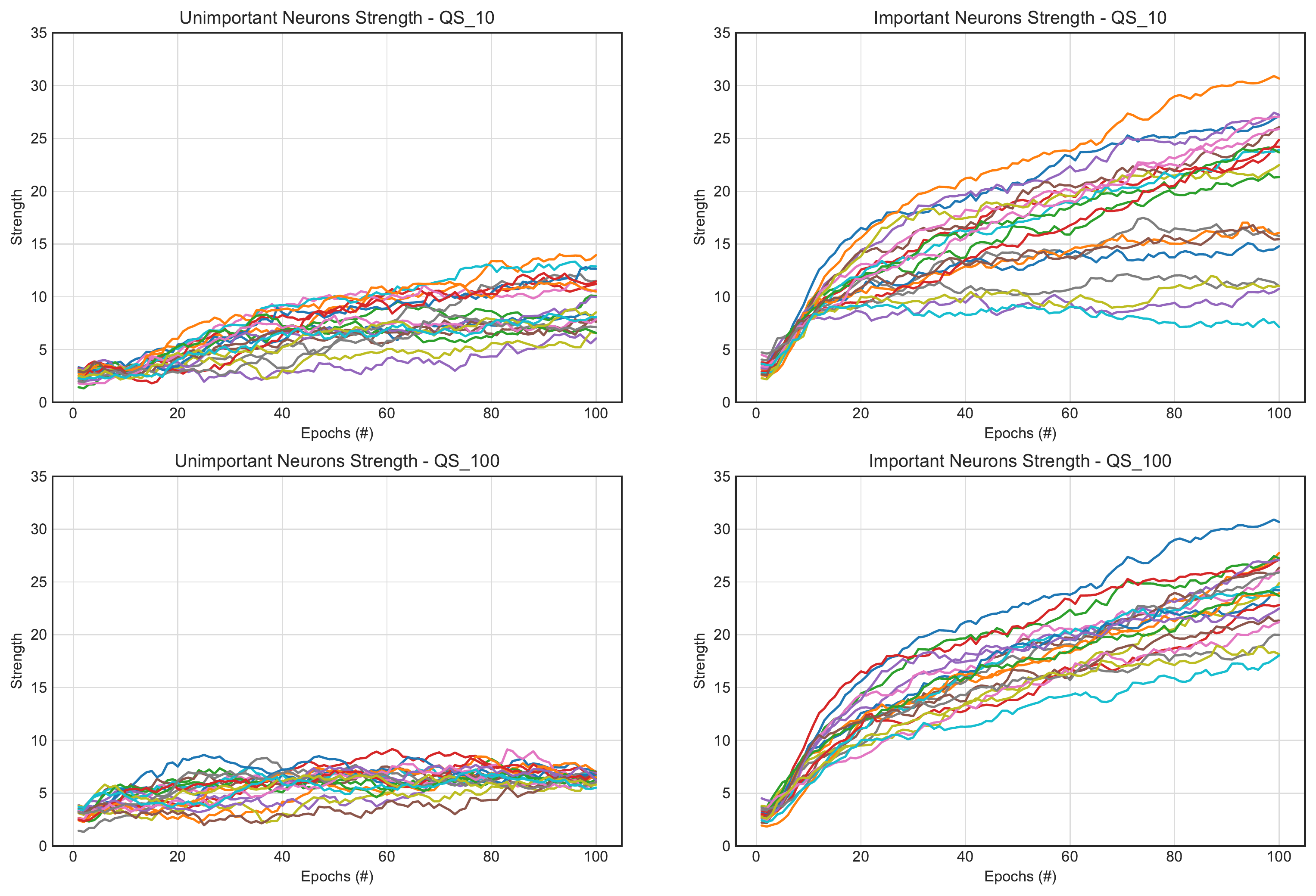} 
        \caption{Strength of the 20 most informative and non-informative features of Madelon dataset, selected by $QS_{10}$ and $QS_{100}$. Each line in the plots corresponds to the strength values of a selected feature by $QS_{10}$/$QS_{100}$ during training. The features selected by $QS_{10}$ have been observed until epoch $100$ to compare the quality of these features with $QS_{100}$.}
        \label{fig:strength_madelon}
        \end{figure}

    To support our claim, we observe the evolution of neurons' strength on the Madelon dataset. This choice is made due to the distinction of informative and non-informative features in the Madelon dataset. As described earlier, this dataset has 20 informative features, and the rest of the features are non-informative noise. We consider 20 most informative and non-informative features detected by $QS_{10}$ and $QS_{100}$, and monitor their strength during training (as observed in Figure \ref{fig:madelon}, the maximum accuracy is achieved using the 20 most informative features, while the least accuracy is achieved using the least important features). The features selected by $QS_{10}$ are also being monitored after the algorithm is finished (epoch $10$) until epoch $100$, in order to compare the quality of the selected features by $QS_{10}$ with $QS_{100}$. In other words, we extract the index of important features using $QS_{10}$, and continue the training without making any changes in the network and monitor how the strength of the neurons corresponding to the selected index would evolve after epoch 10. The results are presented in Figure \ref{fig:strength_madelon}. At the initialization (epoch 0), the strength of all these neurons is almost similar and below $5$. As the training starts, the strength of significant neurons increases, while the strength of unimportant neurons does not change significantly. As can be seen in Figure \ref{fig:strength_madelon}, some of the important features selected by $QS_{10}$ are not among those of $QS_{100}$; this can explain the difference in the performance of these two methods in Table \ref{tab:clustering_acc} and \ref{tab:classification_acc}. However, $QS_{10}$ is able to detect a large majority of the features found by $QS_{100}$; these features are among the most important ones among the final $20$ selected features. Therefore, we can conclude that most of the important features are detectable by QuickSelection, even at the first few epochs of the algorithm.    
    
\section{Conclusion}
    In this paper, a novel method (QuickSelection) for energy-efficient unsupervised feature selection has been proposed. It introduces neuron strength in sparse neural networks as a measure of feature importance. Besides, it proposes sparse DAE to accurately model the data distribution and to rank all features simultaneously based on their importance. By using sparse layers instead of dense ones from the beginning, the number of parameters drops significantly. As a result, QuickSelection requires much less memory and computational resources than its equivalent dense model and its competitors. For example, on low-dimensional datasets, including Coil20, Isolet, HAR, and Madelon, and for all values of $K$, QuickSelection\textsubscript{100} which runs on one CPU core is at least 4 times faster than its direct competitor, CAE, which runs on a GPU, while having a close performance in terms of classification and clustering accuracy.   
    We empirically demonstrate that QuickSelection achieves the best trade-off between clustering accuracy, classification accuracy, maximum memory requirement, and running time, among other methods considered. Besides, our proposed method requires the least amount of energy among autoencoder-based methods considered.
    
    The main drawback of the the proposed method is the lack of a parallel implementation. The running time of QuickSelection can be further decreased by an implementation that takes advantage of multi-core CPU or GPU. We believe that interesting future research would be to study the effects of sparse training and neuron strength in other types of autoencoders for feature selection, e.g. CAE. Nevertheless, this paper has just started to explore one of the most important characteristics of QuickSelection, i.e. scalability, and we intend to explore further its full potential on datasets with millions of features. Besides, this paper showed that we can perform feature selection using neural networks efficiently in terms of computational cost and memory requirement. This can pave the way for reducing the ever-increasing computational costs of deep learning models imposed on data centers. As a result, this will not only save the energy costs of processing high-dimensional data but also will ease the challenges of high energy consumption imposed on the environment.


\begin{acknowledgements}
    This research has been partly funded by the NWO EDIC project.
\end{acknowledgements}

\bibliographystyle{plainnat}
\bibliography{QuickSelection}

\newpage
\appendix
\section*{Appendix} 
\section{Performance Evaluation}
    In this appendix, we compare all methods from different aspects including accuracy, memory usage, running time, energy consumption, and the number of parameters. We perform different experiments to gain a deep insight into the performance of QuickSelection.
    
    \subsection{Discussion: Accuracy and Computational Efficiency Trade-off} \label{app_sec:appendix_running_time}
        In this section, we compare the performance of all methods in more detail. We run feature selection for different values of K on each dataset and then measure the performance.   
        
\begin{figure}[htbp]
             \centering
             \begin{subfigure}[!b]{\textwidth}
                 \centering
                 \includegraphics[width=\textwidth]{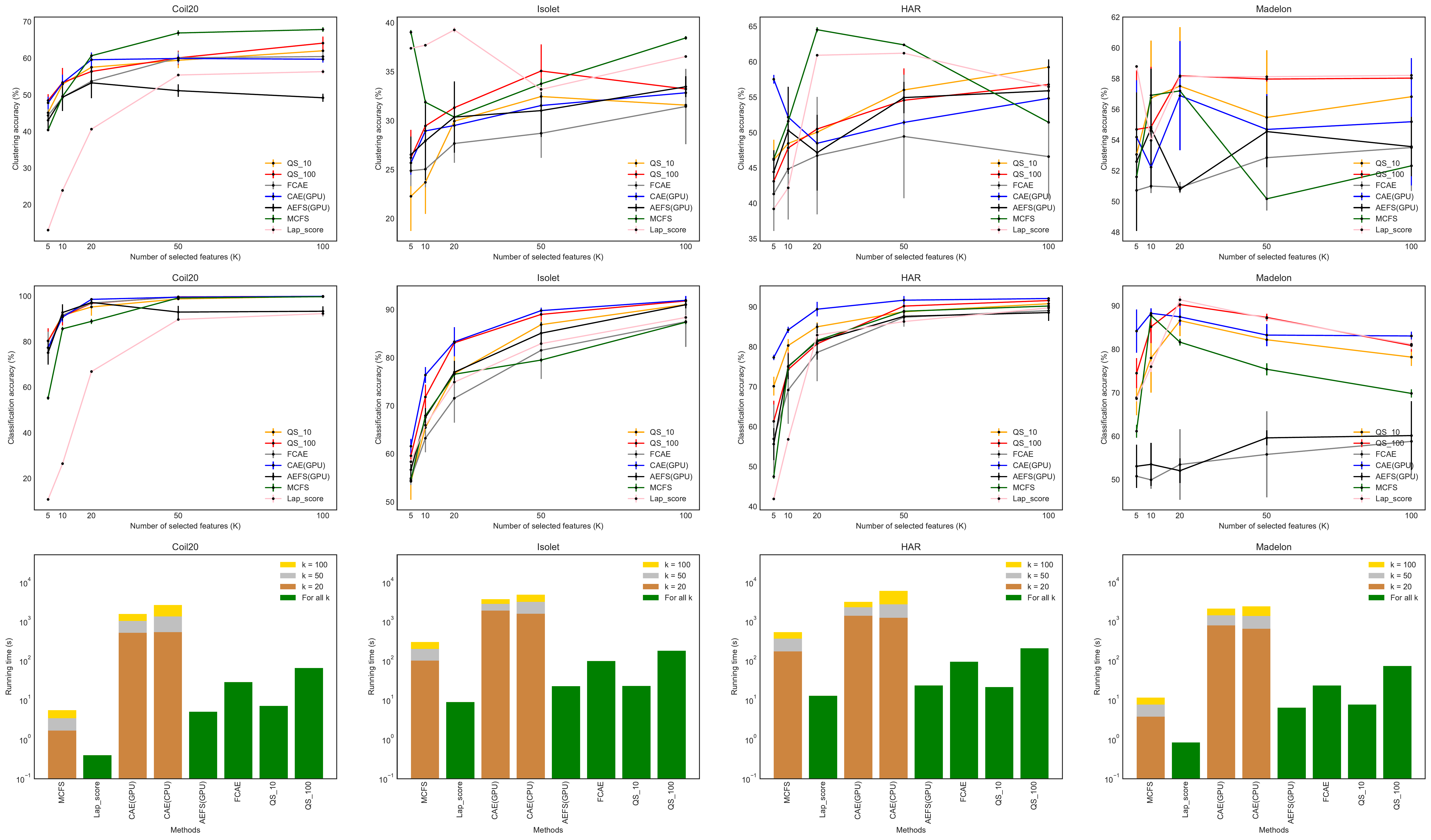}
                 \caption{Low-dimensional datasets}
                 \label{fig:runningt_1}
             \end{subfigure}
             \hfill
             \begin{subfigure}[b]{0.25\textwidth}
                 \centering
                 \includegraphics[width=\textwidth]{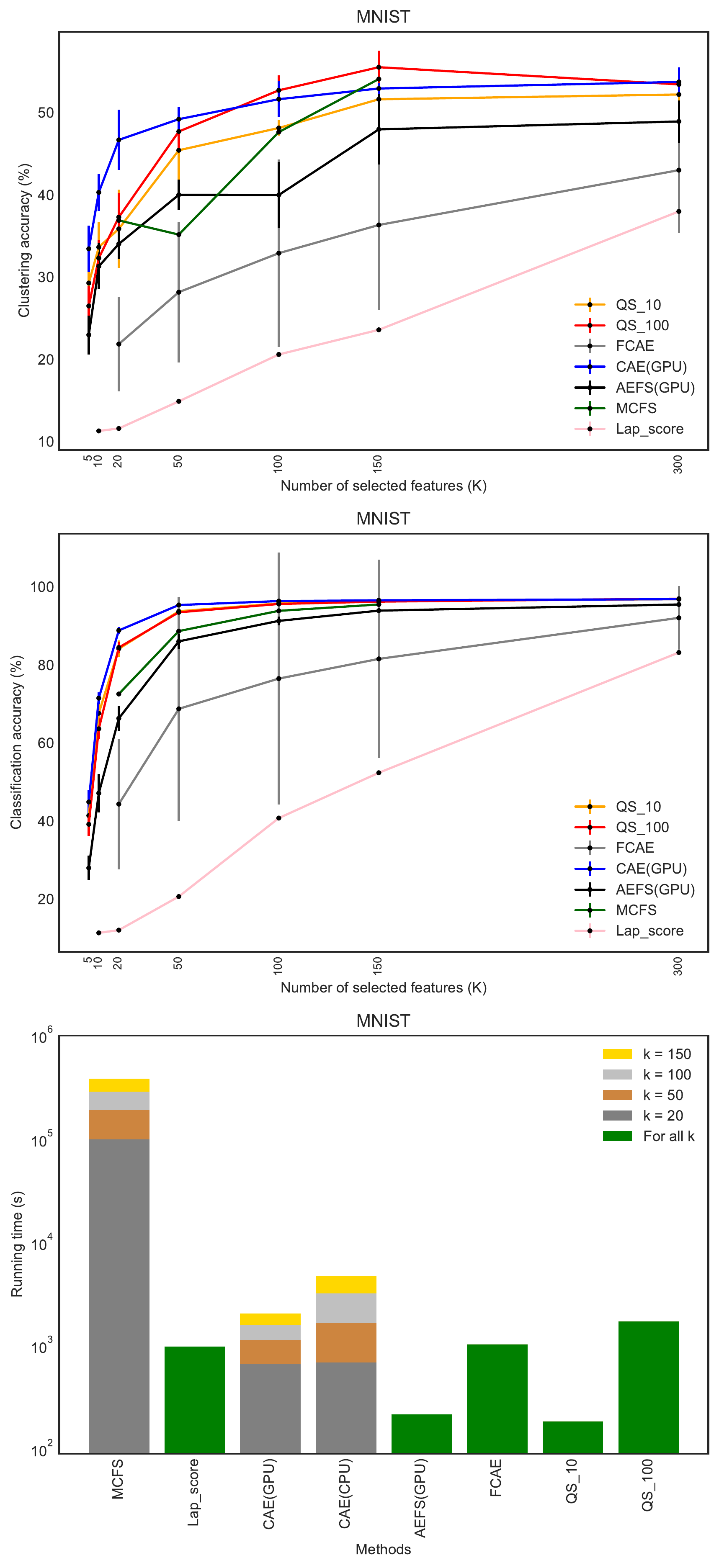}
                 \caption{MNIST}
                 \label{fig:runningt_2}
             \end{subfigure}
             \hfill
             \begin{subfigure}[b]{0.71\textwidth}
                 \centering
                 \includegraphics[width=\textwidth]{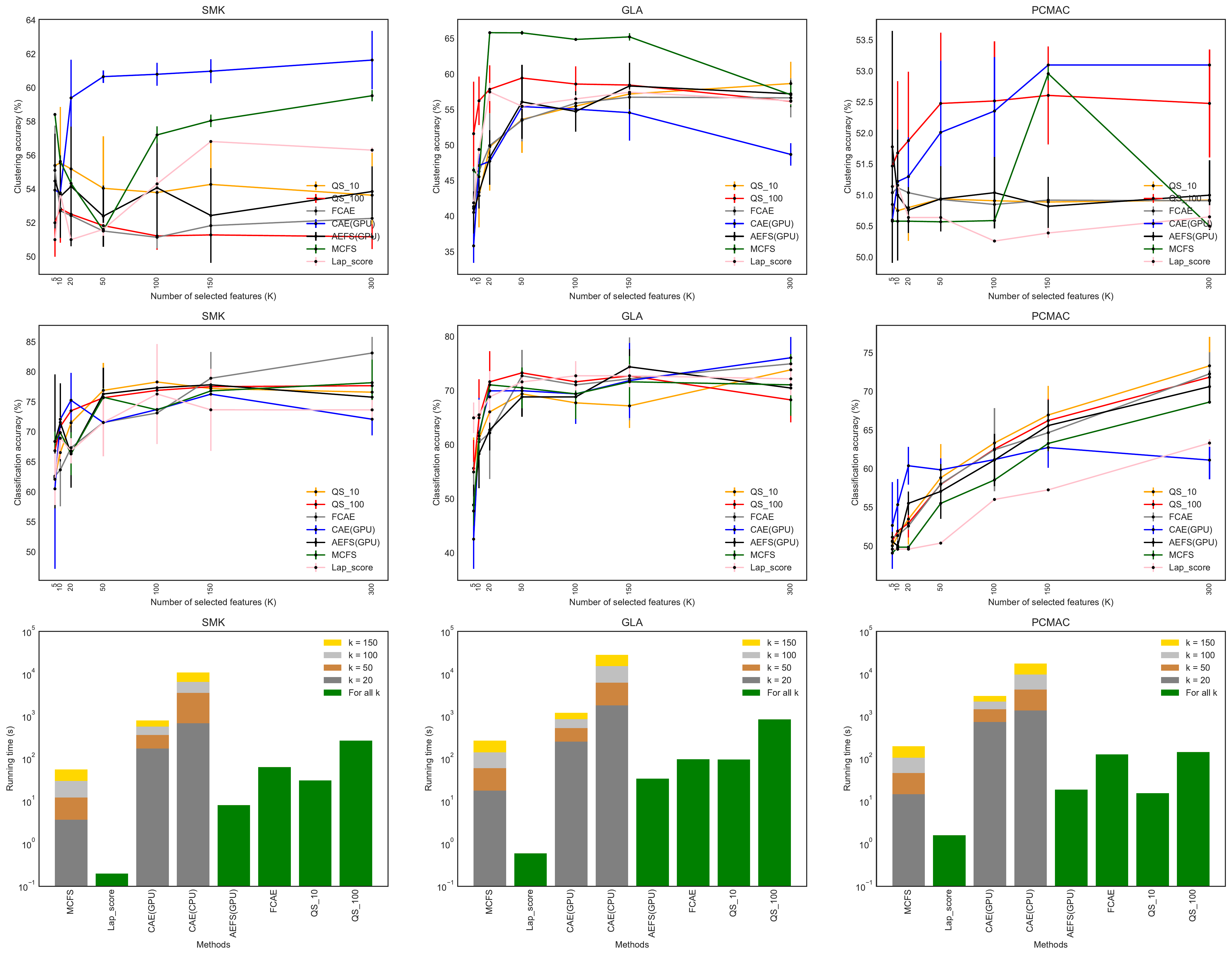}
                 \caption{High-dimensional datasets}
                 \label{fig:runningt_3}
             \end{subfigure}
                \caption{Comparison of clustering accuracy, classification accuracy, and running time for various values of $K$ among all the methods considered on eight datasets, including low-dimensional and high-dimensional datasets. The running time of CAE and AEFS is measured using a GPU, while all the other methods use only a single CPU core.}
                \label{fig:runningt}
            \end{figure}

\begin{figure}[htbp]    
            \centering
            \includegraphics[width=\textwidth]{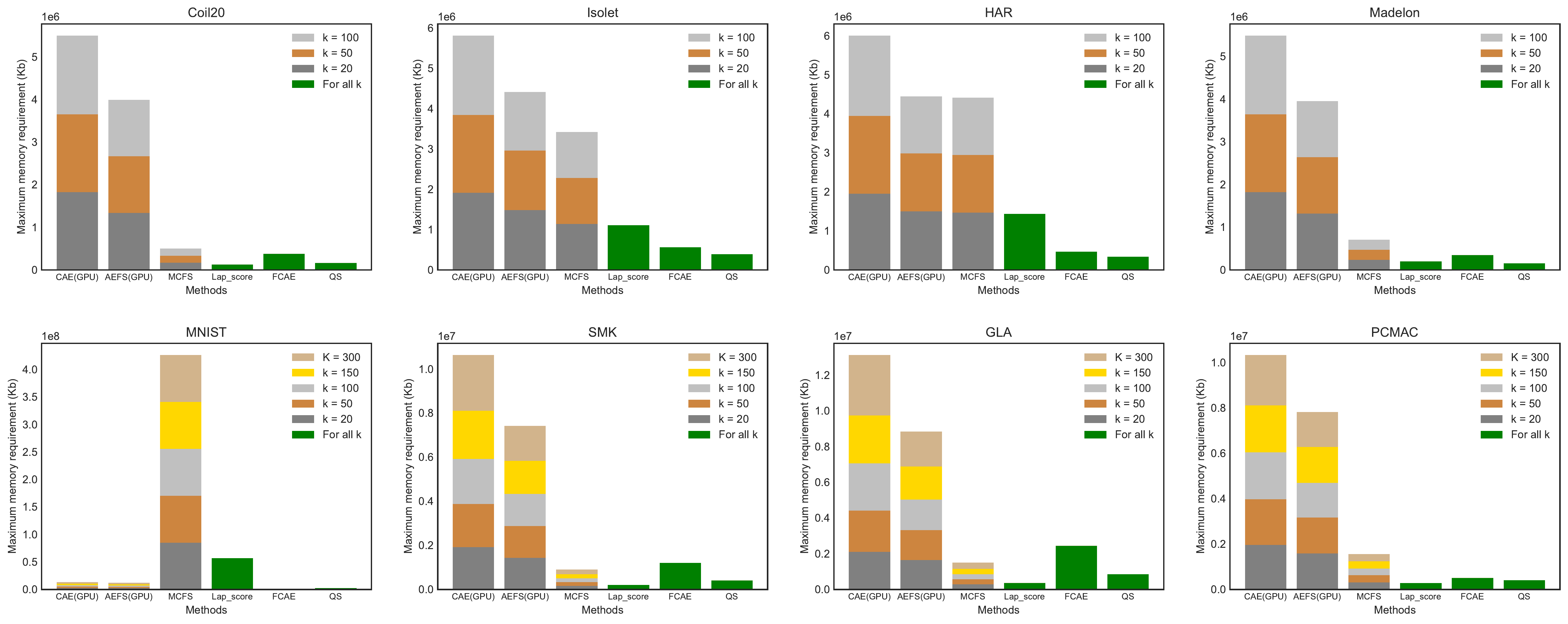} 
            \caption{Maximum memory usage during feature selection for different values of $K$.}
            \label{fig:memory}
            \end{figure}

        As shown in Figure \ref{fig:runningt}, we compare clustering accuracy, classification accuracy, and running time among the methods for different values of $K$. The comparison of maximum memory (RAM) requirement is also depicted in Figure \ref{fig:memory}. For all methods except CAE and AEFS, we run the experiments on a single CPU core. Since the implementations of CAE and AEFS are optimized for GPU, we measure the running time of these methods using a GPU. However, we also consider the running time of CAE using a single CPU core. It should be noticed that since Laplacian score, AEFS, FCAE, and QuickSelection give the ranking of the features as the output of the feature selection process, we need to run them just once for all values of $K$. However, MCFS and CAE need the $K$ value as an input of their algorithm. So, the running time depends on the value of $K$. In the implementation of AEFS, $K$ is used to set the number of hidden values. However, it is not the requirement of the algorithm.
        
\begin{figure}[htbp]    
            \centering
            \includegraphics[width=\textwidth]{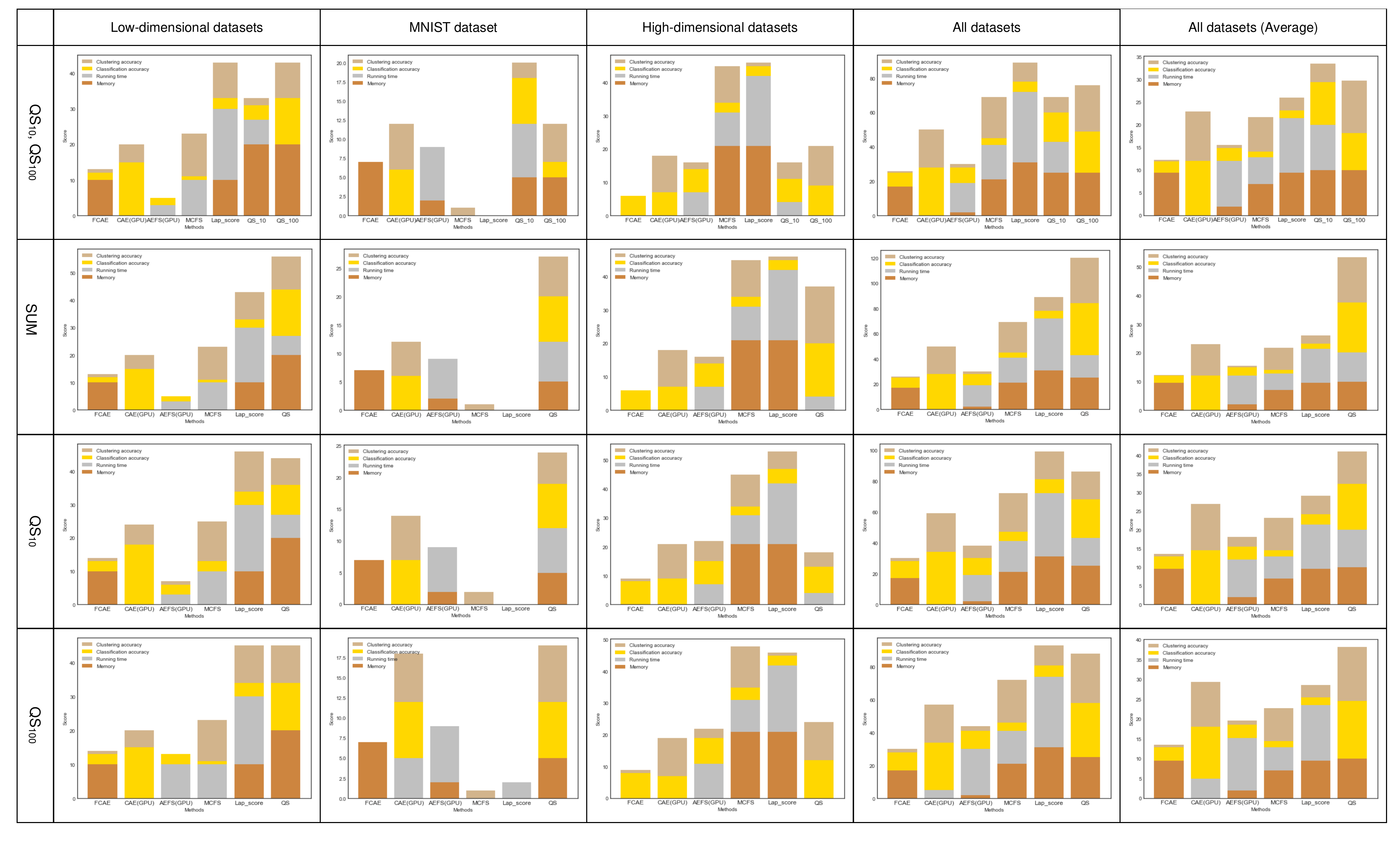} 
            \caption{Feature selection results comparison in terms of classification accuracy, clustering accuracy, speed, and memory. The Scores are based on the ranking of the methods on each dataset and for different values of $K$ (Score $1$). }
            \label{fig:performance_all}
            \end{figure}

        We summarize the results of the aforementioned plots in Figure \ref{fig:performance_all}; we compare the methods using the \textit{score 1}, which is introduced in Section \ref{sec:discussion_speed_memory}. This score is computed based on the methods' ranking in clustering accuracy, classification accuracy, running time, and memory. As explained in Section \ref{sec:discussion_speed_memory}, we give a score of one to each method that is the first or second-best performer in each of the considered metrics. Then, we compute a sum over all of these scores on all datasets and on all values of $K$; the final scores for each method can be seen in Figure \ref{fig:performance_all}. The first column depicts the results on low-dimensional datasets with a low number of samples, including Coil20, Isolet, HAR, and Madelon. The second column shows the results corresponding to MNIST. Similarly, the third column corresponds to high-dimensional datasets, including SMK, GLA, and PCMAC. The total score over all of these datasets is shown in the $4^{th}$ column. In Figure \ref{fig:performance_all}, there exist four rows; the first row corresponds to considering QuickSelection$_{10}$ and QuickSelection$_{100}$ simultaneously, and the sum of their scores are depicted in the second row. The last two rows correspond to considering each of these two methods separately. 
        
        However, since the performance of each method can be different in each of the three groups of datasets, we compute a normalized version of the score $1$, based on the number of datasets in each group. For example, the Laplacian score has a poor performance on MNIST, and this pattern would be similar on other datasets with a large number of samples. However, there is just one dataset with a large number of samples in this experiment. On the other hand, on high-dimensional datasets with a low number of samples, this method has a good performance in terms of running time, and we have three datasets with such characteristics. So, we normalize the values of score $1$, such that instead of giving a score of one to each method, we give a score of one divided by the number of datasets in the corresponding group. The results of the normalized score $1$ are shown in the last column of Figure \ref{fig:performance_all}.


    \subsection{Energy Consumption}\label{app_sec:appendix_running_time_power}
\begin{figure}[htbp]    
            \centering
            \includegraphics[width=\textwidth]{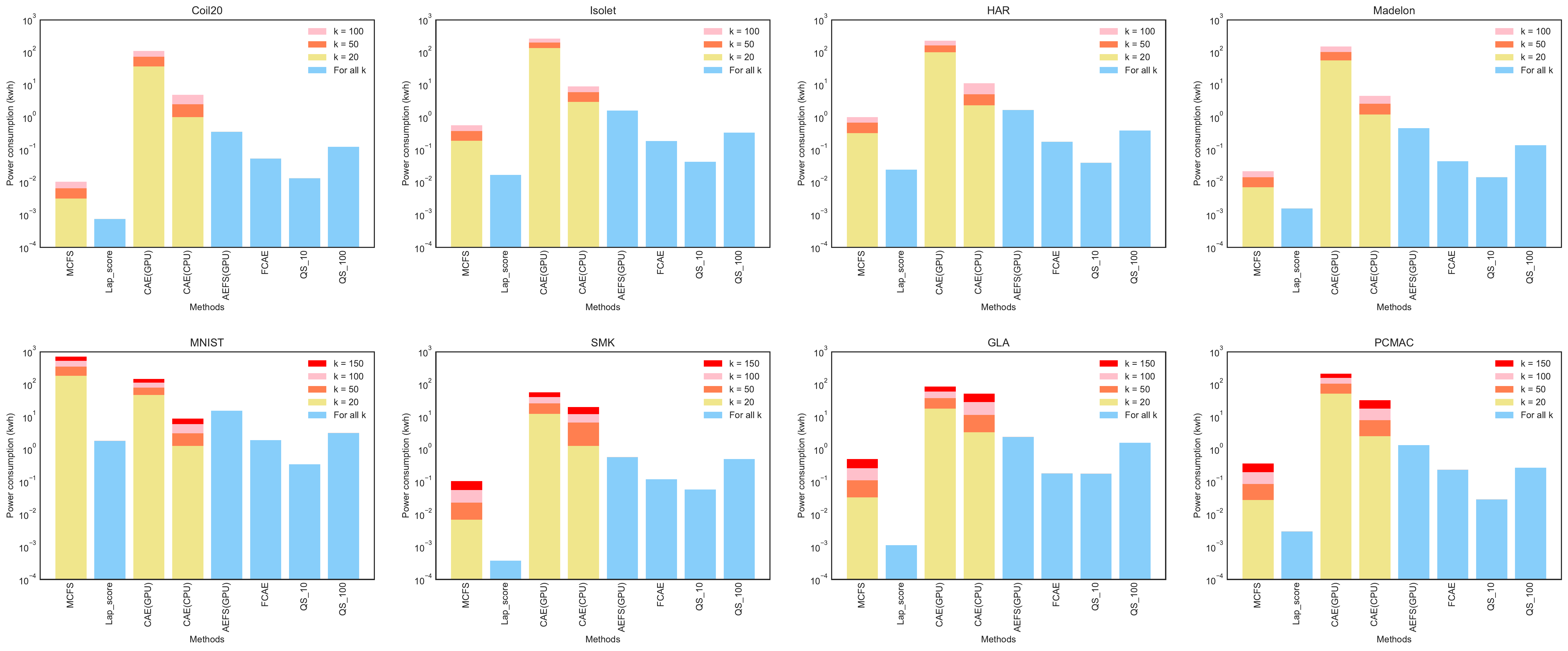} 
            \caption{Energy consumption of all methods for different values of $K$.}
            \label{fig:power}
            \end{figure}

        We perform another experiment regarding the comparison of energy consumption among all methods. The results are presented in Figure \ref{fig:power}. More details regarding this plot are given in the paper in Section \ref{sec:discussion_speed_memory}.
    
    \subsection{Number of Parameters}\label{app_sec:appendix_running_time_num_params}
\begin{figure}[htbp]
            \centering
            \includegraphics[width=\linewidth]{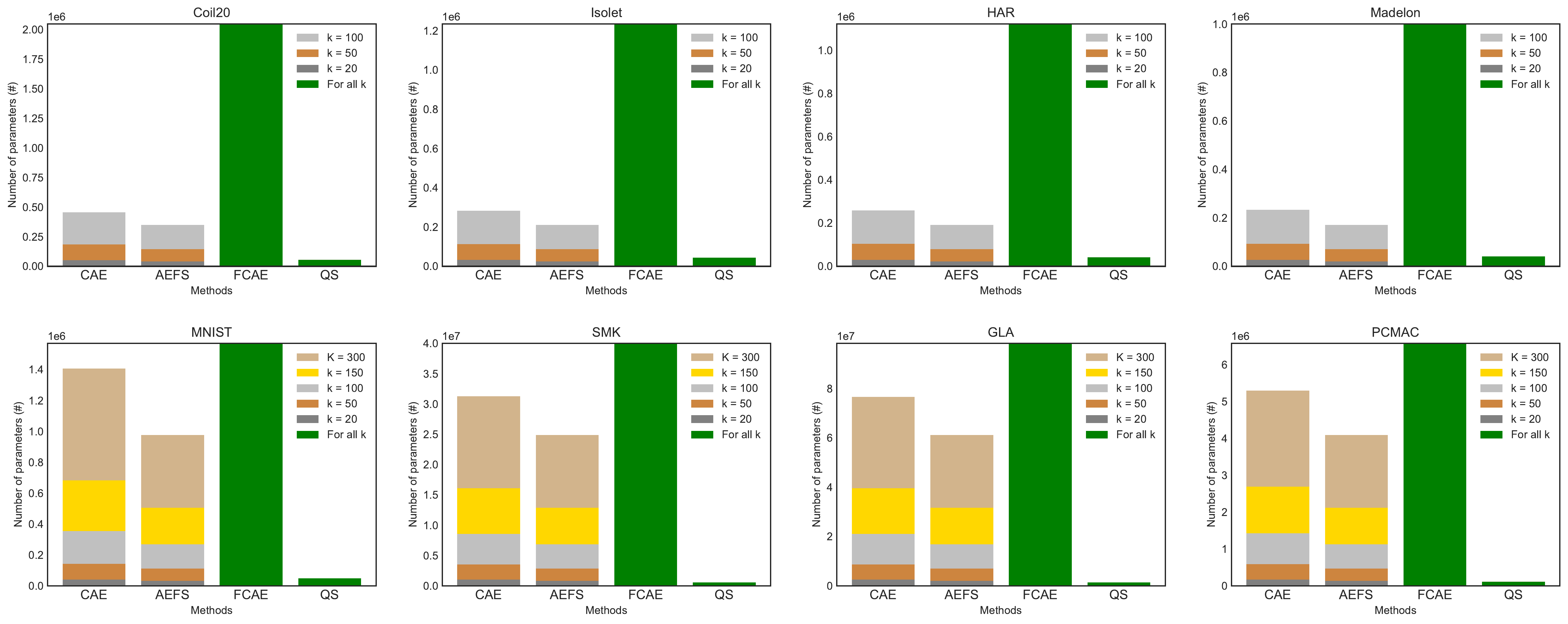}
            \caption{Number of parameters of autoencoder-based models for different values of $K$.}
            \label{fig:num_params}
            \end{figure}

        In Figure \ref{fig:num_params}, we compare the number of parameters of the autoencoder-based methods. FCAE, a fully connected-autoencoder with 1000 hidden neurons, has the highest number of parameters on all datasets. Our proposed network, sparse DAE, has the lowest number of parameters in most cases. It has 1000 hidden neurons that are sparsely connected to input and output neurons. The number of parameters of AEFS and CAE depends on the number of selected features. As also mentioned earlier, the structure of AEFS is similar to FCAE with a difference in the number of hidden neurons. The number of hidden neurons in the implementation of AEFS is set to $K$.
        
\section{Parameter Selection}  \label{app_sec:appendix_parameter_selection}
    In this appendix, we discuss the effect of three hyperparameters of QuickSelection on feature selection performance. 
    \subsection{Noise Factor}\label{app_ssec:appendix_select_noise_factor}
\begin{figure}[!t]
            \centering
            \includegraphics[width=0.9\columnwidth]{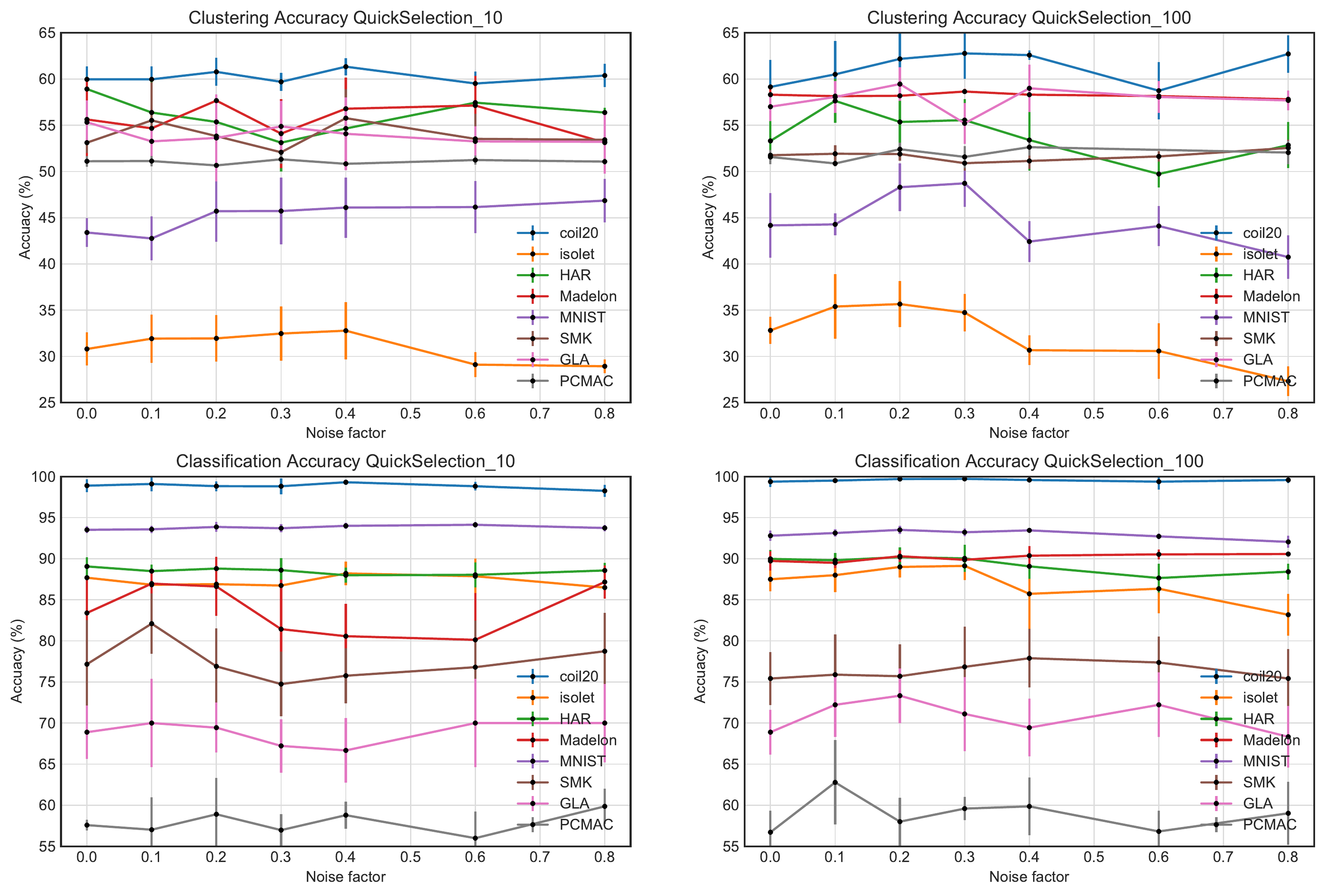}
            \caption{Clustering and classification accuracy for feature selection using QuickSelection\textsubscript{10} and QuickSelection\textsubscript{100} with different values of noise factor. We select 50 features from all datasets except Madelon for which we select 20 features.}
            \label{fig:noise_val}
        
            \end{figure}

        To analyze the effect of the noise level on QuickSelection behavior, we evaluate the sparse DAE model with different noise factors. To this end, we test different noise factors between $0$ and $0.8$. The results can be observed in Figure \ref{fig:noise_val}. These results are an average of 5 runs for each case. 
        
        We can observe that adding $20\%$ to $40\%$ noise on the data seems to be optimal; it improves the performance on most of the datasets for QuickSelection\textsubscript{10} and QuickSelection\textsubscript{100} compared to the model without any noise. We choose the noise factor of $0.2$ for all the experiments.
        
        It is clear in Figure \ref{fig:noise_val}, that setting the noise factor to a large value may corrupt the input data in such a way that the network would not be able to model the data distribution accurately. For example, on the Isolet dataset, the clustering accuracy degrades for $10\%$ when we add $80\%$ noise on the input data compared to the model with the noise factor of $0.2$. Also, the result is less stable when we add a large amount of noise. In this example, we can observe that adding $20\%$ noise to the original data improves both classification and clustering accuracy of QuickSelection\textsubscript{100} by approximately $3\%$. 
        
        From this figure, it can be observed that the improvement of adding noise, is more obvious in QuickSelection\textsubscript{100} than QuickSelection\textsubscript{10}. When we add noise to the data, it needs more time to learn the original structure of the data. So, we need to run it for more epochs to get a proper result. 
       
    \subsection{SET Hyperparameters} \label{app_ssec:appendix_select_set_parameter}
     
        \begin{table}[!b]
            \centering \tabcolsep=0.09cm
            \caption{$\epsilon$ values and their corresponding density level.}
            \label{tab:epsilon_density}
            \vskip 0.01in\begin{center}
            \resizebox{0.5\columnwidth}{!}{%
                \begin{tabular}{ccccccccc}
                \toprule
                &\multicolumn{6}{c}{\textbf{Density [\%]}}\\\cmidrule{2-9}
                \textbf{$\epsilon$} &COIL-20  &Isolet &HAR &Madelon &MNIST &SMK   &GLA &PCMAC\\
                \midrule
                2  & 0.39 & 0.52 & 0.39 &0.59 & 0.45 & 0.20 &0.2  &0.26\\
                5  & 0.98 & 1.30 & 0.98 &1.48 & 1.13 & 0.53 &0.51 &0.65\\
                10 & 1.95 & 2.58 & 1.95 &2.95 & 2.25 & 1.04 &1.02 &1.13\\
                13 & 2.53 & 3.35 & 2.53 &3.82 & 2.91 & 1.35 &1.32 &1.69\\
                20 & 3.87 & 5.10 & 3.87 &5.82 & 4.45 & 2.07 &2.04 &2.6 \\
                25 & 4.87 & 6.45 & 4.87 &7.37 & 5.63 & 2.65 &2.55 &3.26\\
                \bottomrule
            \end{tabular}}\end{center}\vskip -0.1in
        \end{table}
        
        As explained in the paper, $\zeta$ and $\epsilon$ are the hyperparameters of the SET algorithm which control the number of connections to remove/add for each topology change and the sparsity level, respectively. The corresponding density level of each $\epsilon$ value for each dataset can be observed in Table \ref{tab:epsilon_density}.   
        
        To illustrate the effect of the hyperparameters  $\zeta$ and $\epsilon$, we perform a grid search within a small set of values on all of the datasets. The obtained results can be found in Tables \ref{tab:hyperparameter_qs_10} and \ref{tab:hyperparameter_qs_100}. As we increase the $\epsilon$ value, the number of connections in our model increases, and therefore, the computation time will increase. So, we prefer using small values for this parameter. Additionally, for a large value of $\epsilon$, in some cases the model is not able to converge in 100 epochs; for example, on the MNIST dataset, we can observe that for an $\epsilon$ value of 25, the model has lower performance in terms of clustering and classification accuracy. 
        
        It can be observed that $\zeta=0.2$ and $\epsilon=13$ (as chosen for the experiments performed in the paper) lead to a decent performance on all datasets. For these values, QuickSelection is able to achieve high clustering and classification accuracy.
        
        Overall, although searching for the best pair of  $\zeta$ and $\epsilon$ will improve the performance, QuickSelection is not extremely sensitive to these values. As can be seen in Tables \ref{tab:hyperparameter_qs_10} and \ref{tab:hyperparameter_qs_100}, for all values of these hyperparameters QuickSelection has a reasonable performance. Even with $\epsilon = 2$ which leads to a very sparse model, QuickSelection has decent performance, and in some cases better than a denser network.

    \begin{table}[htbp]
        \centering
        \caption{Hyper-parameter selection for QuickSelection\textsubscript{10}. Each entry of each table contains clustering accuracy and classification accuracy in percentages (\%), respectively.}
        \label{tab:hyperparameter_qs_10}
        \begin{center}
        \begin{scriptsize}
        \begin{subtable}[t]{\textwidth}
            \caption{Coil20}
            \label{tab:QS10_parameter_coil20}
                \begin{tabular}{l@{\hskip 0.07in}l@{\hskip 0.07in}l@{\hskip 0.07in}l@{\hskip 0.07in}l@{\hskip 0.07in}l@{\hskip 0.07in}l}
                \hline
                &\multicolumn{6}{c}{\textbf{$\epsilon$}}\\\cmidrule{2-7}  
                 \textbf{$\zeta$}   & \textbf{2}                 & \textbf{5}                 & \textbf{10}                & \textbf{13}                & \textbf{20}                & \textbf{25}                \\
                \hline
                 \textbf{0.1}       & 61.4$\pm$2.3, 98.6$\pm$1.5 & 59.7$\pm$1.8, 96.9$\pm$1.6 & 61.7$\pm$2.8, 98.3$\pm$0.8 & 60.3$\pm$2.0, 98.6$\pm$1.7 & 61.9$\pm$0.9, 99.5$\pm$0.3 & 60.8$\pm$2.4, 99.7$\pm$0.2 \\
                 \textbf{0.2}       & 59.4$\pm$3.8, 97.9$\pm$1.4 & 59.3$\pm$0.9, 98.5$\pm$1.2 & 60.1$\pm$2.6, 98.9$\pm$0.6 & 59.5$\pm$2.1, 98.8$\pm$0.6 & 63.6$\pm$2.5, 99.7$\pm$0.3 & 61.0$\pm$3.0, 99.7$\pm$0.4 \\
                 \textbf{0.3}       & 61.3$\pm$2.1, 98.7$\pm$1.6 & 59.5$\pm$1.9, 96.7$\pm$0.6 & 60.3$\pm$0.9, 99.1$\pm$0.6 & 60.1$\pm$1.9, 98.2$\pm$1.0 & 59.9$\pm$2.6, 98.7$\pm$0.9 & 60.8$\pm$1.9, 99.4$\pm$0.4 \\
                 \textbf{0.4}       & 60.5$\pm$3.2, 97.5$\pm$1.3 & 59.0$\pm$2.4, 96.8$\pm$1.8 & 57.2$\pm$1.5, 97.7$\pm$1.7 & 62.0$\pm$2.5, 99.2$\pm$0.3 & 62.1$\pm$2.5, 99.7$\pm$0.3 & 63.8$\pm$1.5, 99.4$\pm$0.6 \\
                 \textbf{0.5}       & 61.2$\pm$2.3, 98.0$\pm$0.9 & 58.4$\pm$2.6, 97.8$\pm$1.2 & 58.9$\pm$2.6, 97.3$\pm$1.3 & 60.6$\pm$1.6, 98.2$\pm$1.6 & 59.1$\pm$2.9, 99.0$\pm$0.6 & 62.8$\pm$1.6, 99.0$\pm$1.0 \\
                \hline
                \end{tabular}
         
            \end{subtable}
        \begin{subtable}[t]{\textwidth}
            \caption{Isolet}
            \label{tab:QS10_parameter_isolet}
            \begin{tabular}{l@{\hskip 0.07in}l@{\hskip 0.07in}l@{\hskip 0.07in}l@{\hskip 0.07in}l@{\hskip 0.07in}l@{\hskip 0.07in}l}
            \hline
                &\multicolumn{6}{c}{\textbf{$\epsilon$}}\\\cmidrule{2-7}  
             \textbf{$\zeta$}   & \textbf{2}                 & \textbf{5}                 & \textbf{10}                & \textbf{13}                & \textbf{20}                & \textbf{25}                \\
            \hline
             \textbf{0.1}       & 28.6$\pm$2.1, 82.5$\pm$3.7 & 31.7$\pm$2.4, 84.9$\pm$2.4 & 31.2$\pm$2.2, 87.4$\pm$1.7 & 29.4$\pm$1.5, 88.4$\pm$1.3 & 32.4$\pm$1.8, 86.9$\pm$1.5 & 31.7$\pm$1.0, 86.9$\pm$0.7 \\
             \textbf{0.2}       & 26.1$\pm$2.2, 81.5$\pm$1.8 & 30.2$\pm$2.4, 86.2$\pm$2.8 & 30.9$\pm$2.3, 88.2$\pm$0.6 & 32.5$\pm$2.8, 86.9$\pm$1.1 & 32.4$\pm$1.6, 87.8$\pm$1.6 & 33.4$\pm$2.1, 87.5$\pm$1.0 \\
             \textbf{0.3}       & 27.9$\pm$2.0, 83.0$\pm$3.6 & 30.6$\pm$3.3, 87.2$\pm$2.1 & 31.9$\pm$2.5, 87.3$\pm$1.1 & 32.9$\pm$1.1, 87.7$\pm$1.1 & 32.2$\pm$1.8, 87.3$\pm$0.9 & 32.3$\pm$2.4, 88.1$\pm$1.7 \\
             \textbf{0.4}       & 26.9$\pm$2.0, 82.5$\pm$2.5 & 30.0$\pm$0.8, 86.4$\pm$3.3 & 31.5$\pm$2.3, 86.2$\pm$2.6 & 29.0$\pm$1.4, 85.4$\pm$1.6 & 34.5$\pm$2.9, 86.5$\pm$2.3 & 31.9$\pm$3.4, 87.1$\pm$1.1 \\
             \textbf{0.5}       & 26.9$\pm$1.9, 81.8$\pm$1.7 & 30.7$\pm$1.9, 84.9$\pm$3.1 & 30.7$\pm$2.3, 86.7$\pm$2.3 & 31.2$\pm$3.5, 86.9$\pm$1.7 & 33.1$\pm$1.5, 87.4$\pm$0.8 & 33.8$\pm$2.1, 87.1$\pm$0.6 \\
            \hline
            \end{tabular}
         
            \end{subtable} 
        \begin{subtable}[t]{\textwidth}
            \caption{HAR}
            \label{tab:QS10_parameter_har}
            \begin{tabular}{l@{\hskip 0.07in}l@{\hskip 0.07in}l@{\hskip 0.07in}l@{\hskip 0.07in}l@{\hskip 0.07in}l@{\hskip 0.07in}l}
                \hline
                &\multicolumn{6}{c}{\textbf{$\epsilon$}}\\\cmidrule{2-7} 
                 \textbf{$\zeta$}   & \textbf{2}                 & \textbf{5}                 & \textbf{10}                & \textbf{13}                & \textbf{20}                & \textbf{25}                \\
                \hline
                 \textbf{0.1}       & 43.0$\pm$2.7, 83.2$\pm$1.6 & 56.8$\pm$1.0, 88.6$\pm$1.5 & 55.8$\pm$3.8, 88.4$\pm$0.6 & 56.0$\pm$2.6, 87.7$\pm$0.8 & 56.6$\pm$4.9, 90.1$\pm$1.9 & 56.4$\pm$4.7, 88.4$\pm$1.3 \\
                 \textbf{0.2}       & 43.8$\pm$2.1, 82.7$\pm$2.6 & 56.5$\pm$0.9, 89.2$\pm$2.2 & 58.8$\pm$1.1, 88.4$\pm$0.5 & 56.0$\pm$2.6, 88.8$\pm$0.7 & 54.5$\pm$4.1, 88.9$\pm$1.3 & 55.0$\pm$2.8, 89.7$\pm$0.8 \\
                 \textbf{0.3}       & 43.0$\pm$2.1, 84.5$\pm$2.2 & 55.7$\pm$1.7, 89.6$\pm$1.4 & 59.2$\pm$1.0, 88.4$\pm$1.1 & 54.7$\pm$5.2, 88.4$\pm$1.5 & 56.3$\pm$2.4, 89.1$\pm$1.8 & 57.7$\pm$2.1, 89.3$\pm$1.7 \\
                 \textbf{0.4}       & 42.4$\pm$3.7, 82.9$\pm$1.4 & 55.5$\pm$1.4, 89.8$\pm$1.5 & 56.9$\pm$1.8, 88.7$\pm$0.6 & 59.1$\pm$0.7, 89.4$\pm$0.6 & 56.7$\pm$3.2, 89.7$\pm$1.2 & 56.9$\pm$2.2, 89.1$\pm$1.8 \\
                 \textbf{0.5}       & 45.9$\pm$7.0, 84.0$\pm$3.1 & 52.2$\pm$5.2, 89.9$\pm$0.7 & 57.9$\pm$1.9, 89.2$\pm$0.6 & 59.5$\pm$4.3, 89.7$\pm$1.1 & 55.3$\pm$2.7, 89.5$\pm$0.6 & 52.6$\pm$2.6, 89.9$\pm$0.8 \\
                \hline
                \end{tabular}
         
            \end{subtable}
        \begin{subtable}[t]{\textwidth}
            \caption{Madelon}
            \label{tab:QS10_parameter_madelon}
                \begin{tabular}{l@{\hskip 0.07in}l@{\hskip 0.07in}l@{\hskip 0.07in}l@{\hskip 0.07in}l@{\hskip 0.07in}l@{\hskip 0.07in}l}
                \hline
                &\multicolumn{6}{c}{\textbf{$\epsilon$}}\\\cmidrule{2-7} 
                 \textbf{$\zeta$}   & \textbf{2}                  & \textbf{5}                 & \textbf{10}                & \textbf{13}                & \textbf{20}                & \textbf{25}                \\
                \hline
                 \textbf{0.1}       & 53.5$\pm$1.3, 55.3$\pm$1.9  & 54.0$\pm$4.0, 62.1$\pm$7.9 & 56.7$\pm$3.8, 78.5$\pm$7.0 & 56.6$\pm$2.5, 86.2$\pm$2.1 & 58.1$\pm$2.5, 87.6$\pm$1.9 & 56.5$\pm$3.9, 89.4$\pm$1.1 \\
                 \textbf{0.2}       & 52.9$\pm$2.8, 61.4$\pm$5.6  & 57.0$\pm$3.2, 68.6$\pm$4.2 & 57.6$\pm$2.9, 83.6$\pm$3.5 & 57.5$\pm$3.8, 86.6$\pm$3.6 & 55.5$\pm$3.7, 88.3$\pm$0.7 & 59.1$\pm$0.9, 86.0$\pm$1.5 \\
                 \textbf{0.3}       & 53.4$\pm$2.5, 58.7$\pm$10.3 & 56.4$\pm$3.5, 67.1$\pm$9.1 & 53.9$\pm$3.1, 81.4$\pm$5.6 & 54.4$\pm$3.3, 86.4$\pm$2.0 & 58.2$\pm$2.7, 88.3$\pm$1.8 & 55.5$\pm$2.6, 88.5$\pm$0.8 \\
                 \textbf{0.4}       & 53.8$\pm$4.1, 55.3$\pm$6.3  & 55.9$\pm$4.3, 62.6$\pm$6.4 & 54.4$\pm$2.4, 80.1$\pm$3.6 & 53.6$\pm$2.4, 84.9$\pm$3.2 & 56.8$\pm$3.0, 89.7$\pm$0.8 & 58.1$\pm$1.8, 86.4$\pm$2.8 \\
                 \textbf{0.5}       & 55.2$\pm$3.8, 56.0$\pm$3.9  & 52.2$\pm$2.3, 61.3$\pm$9.0 & 56.0$\pm$2.4, 81.7$\pm$3.9 & 57.9$\pm$2.8, 84.2$\pm$4.6 & 57.8$\pm$2.3, 86.9$\pm$0.6 & 57.2$\pm$2.5, 89.0$\pm$2.1 \\
                \hline
                \end{tabular}
                         
            \end{subtable}
        \begin{subtable}[t]{\textwidth}
            \caption{MNIST}
            \label{tab:QS10_parameter_mnist}
                \begin{tabular}{l@{\hskip 0.07in}l@{\hskip 0.07in}l@{\hskip 0.07in}l@{\hskip 0.07in}l@{\hskip 0.07in}l@{\hskip 0.07in}l}
                \hline
                &\multicolumn{6}{c}{\textbf{$\epsilon$}}\\\cmidrule{2-7} 
                 \textbf{$\zeta$}   & \textbf{2}                 & \textbf{5}                 & \textbf{10}                & \textbf{13}                & \textbf{20}                & \textbf{25}                \\
                \hline
                 \textbf{0.1}       & 42.6$\pm$3.5, 91.4$\pm$0.5 & 41.2$\pm$1.8, 93.2$\pm$0.4 & 46.8$\pm$4.4, 94.3$\pm$0.2 & 46.9$\pm$1.3, 93.9$\pm$0.3 & 43.3$\pm$1.7, 93.5$\pm$0.4 & 39.5$\pm$1.7, 92.4$\pm$1.2 \\
                 \textbf{0.2}       & 43.8$\pm$2.3, 92.5$\pm$0.6 & 43.9$\pm$1.4, 93.4$\pm$0.6 & 43.0$\pm$3.2, 93.6$\pm$0.4 & 45.4$\pm$3.9, 93.8$\pm$0.6 & 38.5$\pm$2.9, 92.7$\pm$0.6 & 37.6$\pm$3.4, 91.2$\pm$1.1 \\
                 \textbf{0.3}       & 42.1$\pm$1.8, 91.9$\pm$0.5 & 45.4$\pm$2.4, 94.2$\pm$0.4 & 47.4$\pm$1.2, 94.0$\pm$0.1 & 45.5$\pm$3.3, 94.0$\pm$0.4 & 39.7$\pm$1.9, 92.8$\pm$0.7 & 33.9$\pm$3.3, 88.3$\pm$2.1 \\
                 \textbf{0.4}       & 41.9$\pm$2.0, 92.9$\pm$0.7 & 46.9$\pm$2.6, 93.7$\pm$0.4 & 46.2$\pm$0.9, 94.1$\pm$0.3 & 43.4$\pm$1.5, 93.7$\pm$0.3 & 37.0$\pm$3.6, 90.9$\pm$1.5 & 27.9$\pm$2.2, 81.7$\pm$3.1 \\
                 \textbf{0.5}       & 43.3$\pm$3.6, 92.3$\pm$0.5 & 45.9$\pm$4.8, 93.8$\pm$0.6 & 45.2$\pm$2.9, 93.8$\pm$0.4 & 42.8$\pm$2.7, 93.8$\pm$0.7 & 39.7$\pm$2.8, 91.3$\pm$0.6 & 28.0$\pm$2.5, 77.7$\pm$6.1 \\
                \hline
                \end{tabular}
         
            \end{subtable}
        \begin{subtable}[t]{\textwidth}
            \caption{SMK}
            \label{tab:QS10_parameter_smk}
            \begin{tabular}{l@{\hskip 0.07in}l@{\hskip 0.07in}l@{\hskip 0.07in}l@{\hskip 0.07in}l@{\hskip 0.07in}l@{\hskip 0.07in}l}
             \hline
                &\multicolumn{6}{c}{\textbf{$\epsilon$}}\\\cmidrule{2-7} 
                 \textbf{$\zeta$}   & \textbf{2}                 & \textbf{5}                 & \textbf{10}                & \textbf{13}                & \textbf{20}                & \textbf{25}                \\
                \hline
                 \textbf{0.1}       & 52.4$\pm$1.3, 72.1$\pm$7.0 & 53.8$\pm$3.0, 79.5$\pm$2.0 & 52.7$\pm$1.6, 76.8$\pm$4.5 & 56.0$\pm$1.7, 73.7$\pm$4.1 & 55.0$\pm$2.3, 74.2$\pm$7.3 & 53.7$\pm$2.4, 76.3$\pm$4.4 \\
                 \textbf{0.2}       & 54.1$\pm$1.7, 73.7$\pm$4.7 & 53.5$\pm$2.7, 74.2$\pm$8.4 & 55.3$\pm$2.6, 75.3$\pm$4.9 & 54.0$\pm$3.1, 76.9$\pm$4.6 & 52.9$\pm$1.2, 81.6$\pm$5.0 & 54.5$\pm$3.0, 76.8$\pm$6.3 \\
                 \textbf{0.3}       & 56.9$\pm$2.7, 76.8$\pm$6.1 & 54.7$\pm$1.3, 75.3$\pm$5.2 & 53.9$\pm$2.4, 74.7$\pm$4.9 & 53.9$\pm$2.3, 74.2$\pm$4.5 & 54.5$\pm$0.7, 76.3$\pm$3.7 & 54.8$\pm$2.9, 75.8$\pm$3.9 \\
                 \textbf{0.4}       & 55.4$\pm$3.7, 74.7$\pm$2.1 & 55.5$\pm$1.7, 74.2$\pm$2.6 & 53.1$\pm$1.8, 72.6$\pm$4.9 & 52.8$\pm$1.6, 74.7$\pm$4.6 & 53.4$\pm$2.9, 72.6$\pm$4.3 & 53.1$\pm$2.5, 72.6$\pm$4.3 \\
                 \textbf{0.5}       & 53.3$\pm$1.3, 77.4$\pm$5.4 & 55.2$\pm$3.0, 76.3$\pm$3.7 & 53.6$\pm$2.5, 76.3$\pm$6.0 & 52.5$\pm$1.5, 78.9$\pm$4.4 & 52.3$\pm$1.0, 77.4$\pm$7.7 & 51.9$\pm$1.5, 77.9$\pm$2.7 \\
                \hline
                \end{tabular}
         
            \end{subtable}
        \begin{subtable}[t]{\textwidth}
            \caption{GLA}
            \label{tab:QS10_parameter_gla}
                \begin{tabular}{l@{\hskip 0.07in}l@{\hskip 0.07in}l@{\hskip 0.07in}l@{\hskip 0.07in}l@{\hskip 0.07in}l@{\hskip 0.07in}l}
                \hline
                &\multicolumn{6}{c}{\textbf{$\epsilon$}}\\\cmidrule{2-7} 
                 \textbf{$\zeta$}   & \textbf{2}                 & \textbf{5}                 & \textbf{10}                & \textbf{13}                & \textbf{20}                & \textbf{25}                \\
                \hline
                 \textbf{0.1}       & 54.1$\pm$2.8, 66.7$\pm$5.0 & 54.7$\pm$3.5, 67.2$\pm$5.9 & 55.0$\pm$4.6, 66.7$\pm$1.8 & 54.5$\pm$1.5, 67.8$\pm$5.7 & 56.6$\pm$4.0, 75.0$\pm$6.3 & 55.9$\pm$3.2, 68.9$\pm$5.4 \\
                 \textbf{0.2}       & 50.2$\pm$3.5, 67.8$\pm$3.8 & 53.4$\pm$3.3, 67.2$\pm$6.2 & 56.6$\pm$2.5, 70.0$\pm$4.4 & 53.6$\pm$4.7, 69.4$\pm$3.0 & 56.7$\pm$2.2, 68.3$\pm$2.8 & 52.6$\pm$1.5, 68.9$\pm$1.1 \\
                 \textbf{0.3}       & 56.2$\pm$3.5, 68.9$\pm$4.8 & 53.3$\pm$4.8, 68.3$\pm$3.8 & 54.4$\pm$2.4, 67.8$\pm$2.8 & 57.8$\pm$4.3, 70.0$\pm$3.2 & 56.1$\pm$1.9, 70.6$\pm$3.8 & 56.0$\pm$3.0, 71.1$\pm$4.5 \\
                 \textbf{0.4}       & 55.6$\pm$3.5, 68.9$\pm$2.1 & 54.2$\pm$1.5, 68.3$\pm$4.5 & 57.5$\pm$3.1, 68.3$\pm$2.2 & 56.9$\pm$1.1, 70.6$\pm$2.8 & 55.7$\pm$3.6, 68.3$\pm$4.5 & 55.4$\pm$2.4, 68.9$\pm$6.9 \\
                 \textbf{0.5}       & 54.9$\pm$2.6, 68.9$\pm$4.1 & 54.0$\pm$2.5, 66.1$\pm$3.7 & 54.8$\pm$2.4, 71.1$\pm$4.5 & 54.5$\pm$5.1, 67.2$\pm$6.4 & 56.5$\pm$5.6, 71.1$\pm$1.4 & 55.8$\pm$2.0, 65.6$\pm$3.8 \\
                \hline
                \end{tabular}
         
            \end{subtable}
        \begin{subtable}[t]{\textwidth}
            \caption{PCMAC}
            \label{tab:QS10_parameter_mac}

            \begin{tabular}{l@{\hskip 0.07in}l@{\hskip 0.07in}l@{\hskip 0.07in}l@{\hskip 0.07in}l@{\hskip 0.07in}l@{\hskip 0.07in}l}
                \hline             
                &\multicolumn{6}{c}{\textbf{$\epsilon$}}\\\cmidrule{2-7} 
                 \textbf{$\zeta$}   & \textbf{2}                 & \textbf{5}                 & \textbf{10}                & \textbf{13}                & \textbf{20}                & \textbf{25}                \\
                \hline
                 \textbf{0.1}       & 51.0$\pm$0.5, 61.1$\pm$4.2 & 51.0$\pm$0.6, 57.0$\pm$2.0 & 51.1$\pm$1.1, 59.3$\pm$3.4 & 51.4$\pm$0.5, 56.6$\pm$3.0 & 50.9$\pm$0.2, 55.5$\pm$3.5 & 51.3$\pm$0.5, 59.4$\pm$2.2 \\
                 \textbf{0.2}       & 50.5$\pm$0.4, 61.3$\pm$6.1 & 50.8$\pm$0.5, 57.0$\pm$3.5 & 50.7$\pm$0.4, 55.8$\pm$2.1 & 50.9$\pm$0.5, 58.9$\pm$4.4 & 51.0$\pm$0.2, 59.2$\pm$4.0 & 51.0$\pm$0.6, 57.8$\pm$2.1 \\
                 \textbf{0.3}       & 51.3$\pm$1.0, 58.7$\pm$2.9 & 50.9$\pm$0.3, 57.4$\pm$1.1 & 51.0$\pm$0.4, 57.0$\pm$2.0 & 51.2$\pm$0.5, 59.2$\pm$3.2 & 50.7$\pm$0.3, 58.2$\pm$1.9 & 51.1$\pm$0.6, 58.3$\pm$1.9 \\
                 \textbf{0.4}       & 50.7$\pm$0.3, 58.1$\pm$2.4 & 51.3$\pm$0.4, 55.7$\pm$2.8 & 50.9$\pm$0.5, 55.2$\pm$1.0 & 51.1$\pm$0.3, 58.1$\pm$2.5 & 51.1$\pm$0.2, 57.9$\pm$3.7 & 51.6$\pm$0.9, 55.4$\pm$2.2 \\
                 \textbf{0.5}       & 51.1$\pm$0.5, 57.4$\pm$2.4 & 51.1$\pm$0.4, 57.0$\pm$1.6 & 51.2$\pm$0.9, 58.1$\pm$3.0 & 51.0$\pm$0.6, 56.4$\pm$1.4 & 50.9$\pm$0.3, 55.8$\pm$1.9 & 51.6$\pm$0.9, 58.0$\pm$2.4 \\
                \hline
                \end{tabular}

            \end{subtable}

        \end{scriptsize}
        \end{center}
        \end{table}
    \begin{table}[htbp]
        \centering
        \caption{Hyper-parameter selection for QuickSelection\textsubscript{100}. Each entry of each table contains clustering accuracy and classification accuracy in percentages (\%), respectively.}
        \label{tab:hyperparameter_qs_100}
        \begin{center}
        \begin{scriptsize}
        \begin{subtable}[t]{\textwidth}
            \caption{Coil20}
            \label{tab:QS100_parameter_coil20}
            
            \begin{tabular}{l@{\hskip 0.07in}l@{\hskip 0.07in}l@{\hskip 0.07in}l@{\hskip 0.07in}l@{\hskip 0.07in}l@{\hskip 0.07in}l}
            \hline
                &\multicolumn{6}{c}{\textbf{$\epsilon$}}\\\cmidrule{2-7} 
             \textbf{$\zeta$}   & \textbf{2}                 & \textbf{5}                 & \textbf{10}                & \textbf{13}                & \textbf{20}                & \textbf{25}                \\
            \hline
             \textbf{0.1}       & 63.2$\pm$0.7, 99.7$\pm$0.2 & 62.8$\pm$1.1, 99.4$\pm$0.7 & 60.2$\pm$3.5, 99.2$\pm$0.4 & 61.8$\pm$1.5, 99.7$\pm$0.5 & 56.0$\pm$2.3, 98.8$\pm$1.0 & 53.4$\pm$1.7, 98.8$\pm$0.5 \\
             \textbf{0.2}       & 61.3$\pm$0.9, 99.1$\pm$0.7 & 62.1$\pm$3.2, 99.7$\pm$0.1 & 61.7$\pm$2.3, 99.6$\pm$0.4 & 60.2$\pm$2.0, 99.7$\pm$0.3 & 56.9$\pm$1.8, 99.1$\pm$0.6 & 53.9$\pm$1.5, 98.9$\pm$0.7 \\
             \textbf{0.3}       & 62.1$\pm$1.5, 98.5$\pm$0.8 & 62.0$\pm$2.6, 99.4$\pm$0.7 & 60.0$\pm$1.8, 99.5$\pm$0.2 & 60.2$\pm$2.5, 99.3$\pm$0.2 & 55.0$\pm$1.6, 98.8$\pm$0.9 & 53.8$\pm$1.7, 98.3$\pm$0.8 \\
             \textbf{0.4}       & 58.9$\pm$1.3, 98.3$\pm$0.6 & 62.9$\pm$1.0, 99.7$\pm$0.3 & 62.0$\pm$3.0, 99.5$\pm$0.5 & 62.3$\pm$1.4, 99.7$\pm$0.4 & 57.8$\pm$2.5, 99.2$\pm$0.2 & 57.2$\pm$2.2, 99.0$\pm$0.7 \\
             \textbf{0.5}       & 58.1$\pm$1.9, 97.1$\pm$1.7 & 59.9$\pm$1.5, 99.4$\pm$0.4 & 63.2$\pm$2.6, 99.0$\pm$0.8 & 64.2$\pm$1.3, 99.6$\pm$0.3 & 59.2$\pm$2.9, 98.8$\pm$1.1 & 58.0$\pm$1.4, 99.1$\pm$1.0 \\
            \hline
            \end{tabular}
         
            \end{subtable}
        \begin{subtable}[t]{\textwidth}
            \caption{Isolet}
            \label{tab:QS100_parameter_isolet}
            \begin{tabular}{l@{\hskip 0.07in}l@{\hskip 0.07in}l@{\hskip 0.07in}l@{\hskip 0.07in}l@{\hskip 0.07in}l@{\hskip 0.07in}l}
            \hline
                &\multicolumn{6}{c}{\textbf{$\epsilon$}}\\\cmidrule{2-7}
             \textbf{$\zeta$}   & \textbf{2}                 & \textbf{5}                 & \textbf{10}                & \textbf{13}                & \textbf{20}                & \textbf{25}                \\
            \hline
             \textbf{0.1}       & 29.4$\pm$2.2, 87.1$\pm$1.1 & 29.7$\pm$1.5, 84.8$\pm$3.2 & 28.3$\pm$2.7, 83.4$\pm$4.2 & 33.2$\pm$3.0, 89.3$\pm$1.8 & 37.7$\pm$1.9, 87.5$\pm$1.8 & 36.2$\pm$2.4, 88.3$\pm$1.2 \\
             \textbf{0.2}       & 29.4$\pm$2.2, 85.9$\pm$2.1 & 29.6$\pm$2.7, 86.0$\pm$1.8 & 31.5$\pm$2.0, 85.5$\pm$3.7 & 35.1$\pm$2.7, 89.0$\pm$1.3 & 35.5$\pm$2.5, 87.5$\pm$2.2 & 38.9$\pm$1.7, 87.5$\pm$0.4 \\
             \textbf{0.3}       & 30.3$\pm$2.2, 85.7$\pm$3.1 & 30.2$\pm$1.8, 84.2$\pm$3.8 & 30.0$\pm$2.6, 84.5$\pm$1.8 & 33.5$\pm$2.3, 87.6$\pm$1.8 & 35.7$\pm$3.0, 87.1$\pm$2.5 & 38.1$\pm$1.7, 87.4$\pm$1.7 \\
             \textbf{0.4}       & 31.1$\pm$3.3, 85.9$\pm$3.8 & 29.5$\pm$2.5, 86.1$\pm$3.2 & 30.4$\pm$3.4, 83.7$\pm$3.5 & 29.6$\pm$1.3, 85.4$\pm$0.8 & 33.1$\pm$3.6, 87.6$\pm$2.7 & 35.4$\pm$1.5, 87.9$\pm$1.2 \\
             \textbf{0.5}       & 30.4$\pm$2.5, 88.0$\pm$2.1 & 29.5$\pm$1.8, 86.2$\pm$2.7 & 31.5$\pm$2.7, 86.4$\pm$1.9 & 31.4$\pm$2.1, 86.2$\pm$1.9 & 33.3$\pm$2.0, 86.4$\pm$2.1 & 35.7$\pm$2.0, 89.5$\pm$0.6 \\
            \hline
            \end{tabular}
         
            \end{subtable} 
        \begin{subtable}[t]{\textwidth}
            \caption{HAR}
            \label{tab:QS100_parameter_har}
                \begin{tabular}{l@{\hskip 0.07in}l@{\hskip 0.07in}l@{\hskip 0.07in}l@{\hskip 0.07in}l@{\hskip 0.07in}l@{\hskip 0.07in}l}
                \hline
                &\multicolumn{6}{c}{\textbf{$\epsilon$}}\\\cmidrule{2-7}
                 \textbf{$\zeta$}   & \textbf{2}                 & \textbf{5}                 & \textbf{10}                & \textbf{13}                & \textbf{20}                & \textbf{25}                \\
                \hline
                 \textbf{0.1}       & 52.7$\pm$5.2, 88.4$\pm$2.3 & 57.0$\pm$1.4, 89.5$\pm$1.2 & 56.1$\pm$1.9, 88.8$\pm$1.2 & 54.2$\pm$4.1, 88.5$\pm$2.1 & 56.0$\pm$2.0, 89.2$\pm$2.5 & 55.2$\pm$3.5, 87.5$\pm$2.1 \\
                 \textbf{0.2}       & 48.9$\pm$4.0, 85.8$\pm$1.7 & 57.4$\pm$0.4, 90.0$\pm$0.6 & 52.1$\pm$4.5, 88.6$\pm$2.7 & 54.6$\pm$4.5, 90.2$\pm$1.2 & 54.2$\pm$1.8, 89.4$\pm$0.7 & 53.9$\pm$3.0, 89.2$\pm$1.9 \\
                 \textbf{0.3}       & 50.9$\pm$6.7, 88.5$\pm$2.8 & 56.3$\pm$6.4, 89.5$\pm$1.2 & 54.2$\pm$3.6, 90.8$\pm$1.5 & 53.6$\pm$4.4, 90.5$\pm$3.2 & 52.0$\pm$6.3, 88.0$\pm$1.4 & 51.1$\pm$3.1, 89.4$\pm$1.1 \\
                 \textbf{0.4}       & 48.9$\pm$6.1, 88.7$\pm$2.6 & 55.7$\pm$6.6, 90.5$\pm$1.4 & 54.1$\pm$4.1, 91.1$\pm$0.3 & 50.8$\pm$4.1, 89.2$\pm$1.3 & 49.7$\pm$3.2, 90.2$\pm$1.1 & 54.1$\pm$4.4, 89.0$\pm$1.9 \\
                 \textbf{0.5}       & 46.0$\pm$5.7, 87.3$\pm$3.2 & 55.5$\pm$8.1, 90.5$\pm$1.6 & 52.8$\pm$5.3, 90.2$\pm$1.3 & 55.4$\pm$0.7, 90.4$\pm$1.0 & 50.4$\pm$4.3, 89.4$\pm$0.4 & 49.6$\pm$5.3, 88.3$\pm$1.1 \\
                \hline
                \end{tabular}
         
            \end{subtable}
        \begin{subtable}[t]{\textwidth}
            \caption{Madelon}
            \label{tab:QS100_parameter_madelon}
                \begin{tabular}{l@{\hskip 0.07in}l@{\hskip 0.07in}l@{\hskip 0.07in}l@{\hskip 0.07in}l@{\hskip 0.07in}l@{\hskip 0.07in}l}
                \hline
                &\multicolumn{6}{c}{\textbf{$\epsilon$}}\\\cmidrule{2-7}
                 \textbf{$\zeta$}   & \textbf{2}                 & \textbf{5}                 & \textbf{10}                & \textbf{13}                & \textbf{20}                & \textbf{25}                \\
                \hline
                 \textbf{0.1}       & 53.7$\pm$3.3, 75.1$\pm$6.8 & 58.5$\pm$2.8, 86.1$\pm$3.5 & 57.1$\pm$2.2, 90.3$\pm$1.0 & 56.7$\pm$2.2, 89.9$\pm$1.2 & 58.4$\pm$0.5, 90.3$\pm$0.8 & 58.3$\pm$0.5, 89.8$\pm$1.1 \\
                 \textbf{0.2}       & 54.3$\pm$2.4, 81.2$\pm$4.7 & 55.6$\pm$2.5, 88.2$\pm$2.1 & 57.1$\pm$2.6, 89.6$\pm$0.9 & 58.2$\pm$1.5, 90.3$\pm$0.7 & 58.1$\pm$0.1, 90.3$\pm$1.3 & 58.1$\pm$0.0, 90.8$\pm$0.5 \\
                 \textbf{0.3}       & 53.1$\pm$2.6, 82.0$\pm$4.5 & 60.1$\pm$0.8, 87.5$\pm$1.3 & 57.6$\pm$2.0, 89.6$\pm$1.2 & 57.6$\pm$1.5, 89.4$\pm$1.3 & 58.1$\pm$0.0, 90.9$\pm$0.4 & 58.3$\pm$0.5, 89.5$\pm$1.5 \\
                 \textbf{0.4}       & 55.0$\pm$2.8, 78.6$\pm$8.3 & 58.4$\pm$2.9, 87.0$\pm$4.6 & 55.8$\pm$2.2, 90.6$\pm$0.6 & 58.4$\pm$0.7, 90.1$\pm$0.9 & 57.4$\pm$1.6, 90.3$\pm$0.7 & 58.1$\pm$0.0, 90.9$\pm$1.2 \\
                 \textbf{0.5}       & 55.6$\pm$3.2, 74.3$\pm$3.3 & 57.1$\pm$3.2, 87.1$\pm$2.4 & 57.1$\pm$3.5, 90.0$\pm$0.6 & 58.9$\pm$0.4, 90.3$\pm$0.3 & 58.5$\pm$0.6, 89.3$\pm$0.7 & 58.5$\pm$0.4, 89.4$\pm$1.3 \\
                \hline
                \end{tabular}
         
            \end{subtable}
        \begin{subtable}[t]{\textwidth}
            \caption{MNIST}
            \label{tab:QS100_parameter_mnist}
                \begin{tabular}{l@{\hskip 0.07in}l@{\hskip 0.07in}l@{\hskip 0.07in}l@{\hskip 0.07in}l@{\hskip 0.07in}l@{\hskip 0.07in}l}
                \hline
                &\multicolumn{6}{c}{\textbf{$\epsilon$}}\\\cmidrule{2-7}
                 \textbf{$\zeta$}   & \textbf{2}                 & \textbf{5}                 & \textbf{10}                & \textbf{13}                & \textbf{20}                & \textbf{25}                \\
                \hline
                 \textbf{0.1}       & 44.1$\pm$2.2, 92.8$\pm$1.0 & 46.3$\pm$3.4, 94.0$\pm$0.4 & 48.3$\pm$1.7, 94.0$\pm$0.6 & 44.3$\pm$2.7, 93.8$\pm$0.5 & 43.5$\pm$3.4, 92.8$\pm$0.5 & 31.3$\pm$4.3, 85.2$\pm$4.9 \\
                 \textbf{0.2}       & 43.3$\pm$5.1, 93.5$\pm$1.2 & 44.3$\pm$1.4, 93.7$\pm$0.3 & 47.1$\pm$3.1, 93.7$\pm$0.6 & 48.3$\pm$2.4, 93.5$\pm$0.5 & 37.7$\pm$1.3, 91.3$\pm$0.4 & 33.7$\pm$3.5, 87.8$\pm$1.5 \\
                 \textbf{0.3}       & 45.1$\pm$2.8, 93.2$\pm$0.3 & 48.4$\pm$4.5, 93.7$\pm$0.8 & 46.0$\pm$4.6, 93.4$\pm$0.3 & 44.9$\pm$4.4, 93.8$\pm$0.4 & 35.8$\pm$3.0, 91.6$\pm$0.9 & 38.1$\pm$1.3, 90.8$\pm$0.6 \\
                 \textbf{0.4}       & 45.1$\pm$2.4, 93.4$\pm$0.3 & 45.4$\pm$2.3, 94.2$\pm$0.3 & 45.0$\pm$2.1, 93.4$\pm$0.5 & 40.1$\pm$4.0, 92.4$\pm$0.7 & 41.5$\pm$4.9, 91.8$\pm$1.3 & 32.5$\pm$2.7, 87.7$\pm$3.7 \\
                 \textbf{0.5}       & 45.7$\pm$2.6, 93.8$\pm$0.7 & 44.1$\pm$2.6, 93.7$\pm$0.6 & 43.8$\pm$2.6, 93.8$\pm$0.5 & 43.2$\pm$1.8, 92.6$\pm$0.7 & 36.6$\pm$4.5, 91.5$\pm$1.0 & 36.0$\pm$2.7, 88.5$\pm$1.4 \\
                \hline
                \end{tabular}
         
            \end{subtable}
        \begin{subtable}[t]{\textwidth}
            \caption{SMK}
            \label{tab:QS100_parameter_smk}
            \begin{tabular}{l@{\hskip 0.07in}l@{\hskip 0.07in}l@{\hskip 0.07in}l@{\hskip 0.07in}l@{\hskip 0.07in}l@{\hskip 0.07in}l}
                \hline
                &\multicolumn{6}{c}{\textbf{$\epsilon$}}\\\cmidrule{2-7}
                 \textbf{$\zeta$}   & \textbf{2}                 & \textbf{5}                 & \textbf{10}                & \textbf{13}                & \textbf{20}                & \textbf{25}                \\
                \hline
                 \textbf{0.1}       & 53.1$\pm$1.3, 72.6$\pm$5.9 & 52.6$\pm$1.6, 76.3$\pm$2.4 & 51.6$\pm$0.9, 76.8$\pm$4.8 & 53.5$\pm$1.6, 75.8$\pm$2.6 & 54.6$\pm$3.2, 72.6$\pm$2.1 & 51.4$\pm$0.9, 76.8$\pm$5.4 \\
                 \textbf{0.2}       & 53.3$\pm$2.3, 74.2$\pm$5.4 & 53.0$\pm$1.5, 76.8$\pm$3.1 & 51.1$\pm$0.6, 78.4$\pm$5.9 & 51.8$\pm$0.8, 75.7$\pm$3.9 & 51.3$\pm$0.6, 78.4$\pm$3.5 & 51.9$\pm$1.6, 78.4$\pm$6.5 \\
                 \textbf{0.3}       & 53.3$\pm$1.7, 74.2$\pm$4.5 & 50.7$\pm$0.3, 76.3$\pm$6.5 & 51.6$\pm$0.9, 76.8$\pm$6.1 & 50.9$\pm$0.5, 74.7$\pm$3.6 & 51.2$\pm$0.6, 77.4$\pm$2.7 & 51.4$\pm$0.7, 77.4$\pm$5.4 \\
                 \textbf{0.4}       & 52.4$\pm$2.6, 78.4$\pm$4.5 & 51.6$\pm$0.7, 78.4$\pm$2.6 & 50.9$\pm$0.4, 75.8$\pm$5.4 & 51.9$\pm$0.8, 75.3$\pm$7.2 & 51.1$\pm$0.5, 76.8$\pm$4.5 & 50.8$\pm$0.3, 76.3$\pm$2.4 \\
                 \textbf{0.5}       & 53.0$\pm$1.3, 76.8$\pm$6.1 & 52.1$\pm$1.1, 74.7$\pm$2.7 & 51.9$\pm$0.8, 74.2$\pm$3.5 & 50.7$\pm$0.4, 75.8$\pm$3.9 & 50.3$\pm$0.0, 78.9$\pm$4.1 & 51.1$\pm$0.5, 81.0$\pm$4.2 \\
                \hline
                \end{tabular}
         
            \end{subtable}
        \begin{subtable}[t]{\textwidth}
            \caption{GLA}
            \label{tab:QS100_parameter_gla}
            \begin{tabular}{l@{\hskip 0.07in}l@{\hskip 0.07in}l@{\hskip 0.07in}l@{\hskip 0.07in}l@{\hskip 0.07in}l@{\hskip 0.07in}l}
            \hline
                &\multicolumn{6}{c}{\textbf{$\epsilon$}}\\\cmidrule{2-7}
             \textbf{$\zeta$}   & \textbf{2}                 & \textbf{5}                 & \textbf{10}                & \textbf{13}                & \textbf{20}                & \textbf{25}                \\
            \hline
             \textbf{0.1}       & 57.6$\pm$2.6, 68.9$\pm$5.9 & 57.1$\pm$1.9, 67.2$\pm$1.1 & 57.4$\pm$2.8, 72.2$\pm$3.9 & 57.7$\pm$2.9, 68.9$\pm$3.2 & 57.4$\pm$2.9, 73.3$\pm$4.8 & 59.2$\pm$2.7, 71.7$\pm$4.1 \\
             \textbf{0.2}       & 57.0$\pm$3.4, 64.4$\pm$3.2 & 60.8$\pm$3.8, 71.1$\pm$3.3 & 58.7$\pm$3.5, 67.8$\pm$6.0 & 59.5$\pm$1.8, 73.3$\pm$3.3 & 58.6$\pm$2.0, 72.8$\pm$2.1 & 55.6$\pm$1.3, 70.6$\pm$7.2 \\
             \textbf{0.3}       & 57.7$\pm$3.5, 73.9$\pm$3.3 & 58.3$\pm$4.1, 67.2$\pm$3.2 & 54.8$\pm$0.9, 72.2$\pm$3.5 & 58.0$\pm$4.3, 67.8$\pm$3.8 & 56.4$\pm$3.5, 68.3$\pm$4.2 & 57.3$\pm$2.8, 66.7$\pm$2.5 \\
             \textbf{0.4}       & 56.1$\pm$2.6, 71.1$\pm$3.3 & 57.9$\pm$2.9, 67.2$\pm$4.8 & 54.4$\pm$2.5, 67.2$\pm$3.2 & 59.0$\pm$4.0, 69.4$\pm$4.6 & 56.9$\pm$2.3, 69.4$\pm$2.5 & 59.9$\pm$3.6, 69.4$\pm$4.6 \\
             \textbf{0.5}       & 55.2$\pm$2.2, 67.2$\pm$6.4 & 56.0$\pm$1.7, 63.9$\pm$1.8 & 58.0$\pm$2.2, 68.3$\pm$6.0 & 59.0$\pm$3.1, 70.0$\pm$5.4 & 59.5$\pm$3.2, 71.1$\pm$6.2 & 53.6$\pm$1.7, 68.3$\pm$4.2 \\
            \hline
            \end{tabular}
         
            \end{subtable}
        \begin{subtable}[t]{\textwidth}
            \caption{PCMAC}
            \label{tab:QS100_parameter_mac}

            \begin{tabular}{l@{\hskip 0.07in}l@{\hskip 0.07in}l@{\hskip 0.07in}l@{\hskip 0.07in}l@{\hskip 0.07in}l@{\hskip 0.07in}l}
            \hline
                &\multicolumn{6}{c}{\textbf{$\epsilon$}}\\\cmidrule{2-7}
             \textbf{$\zeta$}   & \textbf{2}                 & \textbf{5}                 & \textbf{10}                & \textbf{13}                & \textbf{20}                & \textbf{25}                \\
            \hline
             \textbf{0.1}       & 50.6$\pm$0.3, 58.1$\pm$3.8 & 50.8$\pm$0.4, 57.4$\pm$3.1 & 51.4$\pm$1.2, 58.5$\pm$2.3 & 51.0$\pm$0.4, 59.2$\pm$3.2 & 50.8$\pm$0.2, 59.2$\pm$3.1 & 52.6$\pm$1.0, 58.8$\pm$3.4 \\
             \textbf{0.2}       & 50.7$\pm$0.4, 59.4$\pm$2.9 & 50.7$\pm$0.5, 60.6$\pm$3.4 & 52.1$\pm$1.7, 57.2$\pm$3.4 & 52.5$\pm$1.1, 58.0$\pm$2.9 & 53.1$\pm$0.0, 58.6$\pm$2.6 & 53.1$\pm$0.0, 60.1$\pm$2.0 \\
             \textbf{0.3}       & 51.5$\pm$0.9, 57.2$\pm$2.9 & 51.4$\pm$0.9, 56.0$\pm$2.2 & 51.7$\pm$1.2, 58.1$\pm$0.9 & 52.2$\pm$1.1, 56.5$\pm$1.7 & 53.1$\pm$0.0, 59.5$\pm$2.4 & 53.1$\pm$0.0, 57.3$\pm$4.1 \\
             \textbf{0.4}       & 50.9$\pm$0.4, 59.8$\pm$6.7 & 51.3$\pm$0.9, 56.3$\pm$4.1 & 52.0$\pm$1.3, 57.3$\pm$3.0 & 53.1$\pm$0.0, 56.7$\pm$2.2 & 53.1$\pm$0.0, 56.6$\pm$2.0 & 53.1$\pm$0.0, 57.6$\pm$2.0 \\
             \textbf{0.5}       & 50.7$\pm$0.2, 56.9$\pm$0.5 & 51.3$\pm$0.9, 57.1$\pm$2.1 & 52.6$\pm$0.9, 59.6$\pm$1.9 & 53.1$\pm$0.0, 57.7$\pm$1.8 & 53.1$\pm$0.0, 56.8$\pm$3.4 & 53.1$\pm$0.0, 59.8$\pm$1.6 \\
            \hline
            \end{tabular}
            \end{subtable}

        \end{scriptsize}
        \end{center}
        \end{table}


\section{Visualization of Selected Features on MNIST} \label{app_sec:mnist_features}
\begin{figure}[!h]
            \centering
            \includegraphics[width=0.3\columnwidth]{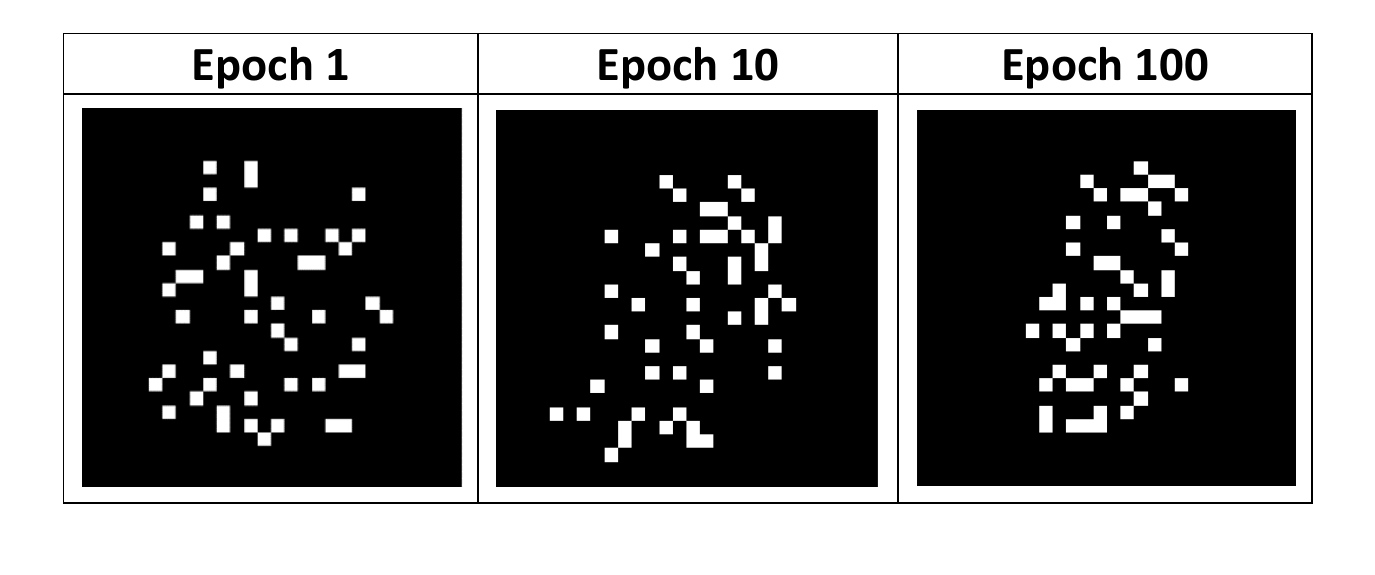}
            \caption{50 most informative features of MNIST dataset selected by QuickSelection after 1, 10, and 100 epochs of training.}
            \label{fig:mnist_50_features}
        \end{figure}

        In Figure \ref{fig:mnist_50_features}, we visualize the 50 best features found by QuickSelection on the MNIST dataset at different epochs. These features are mostly at the center of the image, similar to the pattern of MNIST digits.
        
\begin{figure}[!h]
                \centering
                \includegraphics[width=0.7\columnwidth]{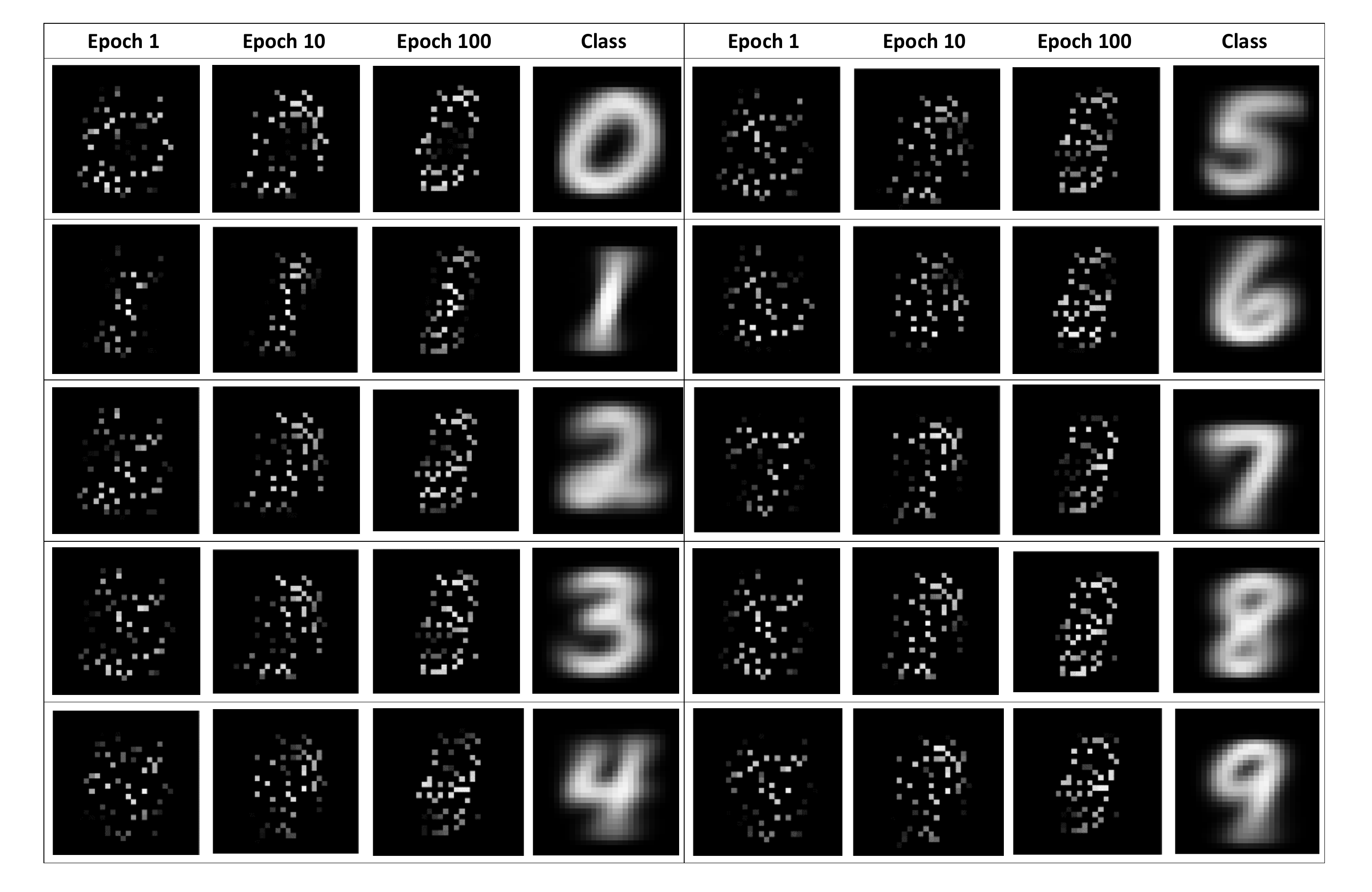}
                \caption{Average of the data samples of each MNIST class corresponding to the 50 selected features after 1, 10, and 100 epochs of training along with the average of the actual samples of each class. }
                \label{fig:mnist_features}
            \end{figure}

        Then, we visualize the features selected for each class separately. In Figure \ref{fig:mnist_features}, each picture at different epochs is the average of the 50 selected features of all the samples of each class along with the average of the actual samples of the corresponding class. As we can see, during training, these features become more similar to the pattern of digits of each class. Thus, QuickSelection is able to find the most relevant features for all classes.
        
     \newpage

\section{Feature Extraction} \label{app_sec:appendix_feature_extraction}
    
        Although it is not the main focus of the paper, we perform a small analysis on the MNIST dataset to study the performance of sparse DAE as a feature extractor. We train it to map the high-dimensional features into a lower-dimensional space.
        
\begin{figure}[!h]
            \centering
            \includegraphics[width=0.6\columnwidth]{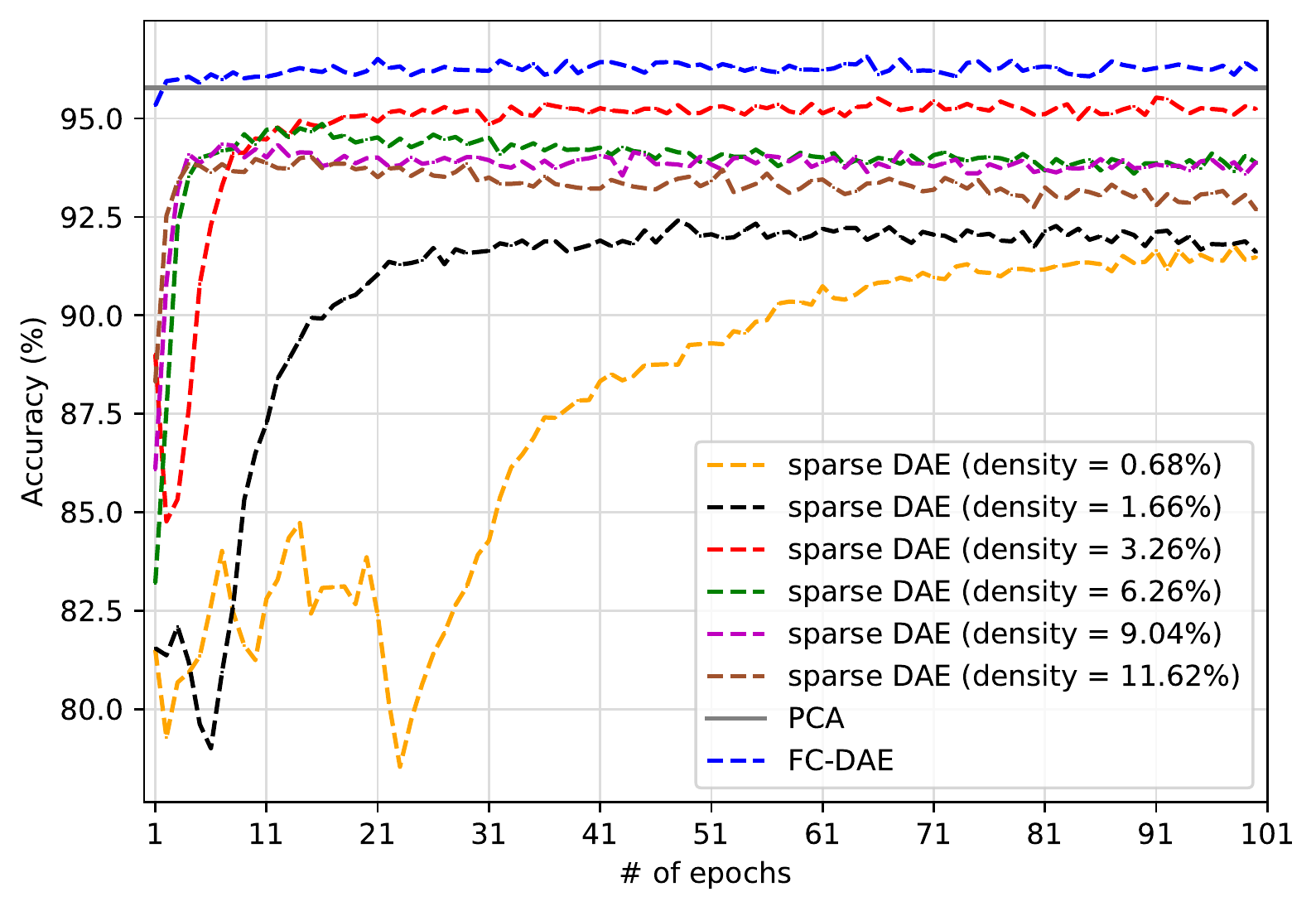}
            \caption{Classification accuracy for feature extraction using sparse DAE with different density level on the MNIST dataset (number of extracted features = 50) compared with FC-DAE and PCA.}
            \label{fig:fxacc}
        \end{figure}

        The structure we consider for feature extraction has three hidden layers with 1000, 50, and 1000 neurons, respectively; the middle layer (50 neurons) is the extracted low-dimensional representation. We compare the results with fully-connected DAE (FC-DAE - implemented in Keras \cite{chollet2015keras}). We also extract features using the Principal Component Analysis (PCA) \cite{wold1987principal} technique as a baseline method. Then, we train an ExtraTrees classifier on these extracted features and compute the classification accuracy. The results are presented in Figure \ref{fig:fxacc}.
    
        To achieve the best density level that suits our network, we test different $\epsilon$ values. As shown in Figure \ref{fig:fxacc}, sparse DAE (density = 3.26\%) has the best performance among them. Sparse DAE (density = 3.26\%), FC-DAE, and PCA achieve 95.2\%, 96.2\%, and 95.6\% accuracy, respectively. Although sparse DAE can not perform as well as the FC-DAE, it approximately has 54 k parameters compared to 1.67 m parameters of FC-DAE. Such a small number of parameters of this model results in a high rise in the running speed and a significant drop in the memory requirement. Furthermore, it is interesting to observe that a very sparse DAE (below 1\% density) can achieve more than 90.0\% accuracy on MNIST while having about 150 times fewer parameters than FC-DAE.


\section{Feature Selection on a Large Dataset} \label{app_sec:appendix_feature_selection_large}
In this appendix, we evaluate the performance of the methods on a very large dataset, in terms of both number of samples and dimensions. 

In this experiment, first, we generate two artificial datasets with high number of samples and features. The choice of an artificial dataset was made to easily control the number of relevant features of the dataset, as in most of the real-world datasets the number of informative features are not clear. These datasets are generated using \textit{sklearn}\footnote{https://scikit-learn.org/} library tools, \textit{make\_classification} function, which generates datasets with a desired number of features and samples. This function allows us to adjust the number of informative, redundant, and non-informative features. Table \ref{tab:fs_large_datasets} shows the characteristics of the two artificially generated datasets. We generated $2$ datasets with $40000$ samples and $8000$ features. However, the number of informative and redundant features are different in these datasets. Artificial$2$ dataset is much noisier than Artificial$1$; therefore, finding relevant features of Artificial$2$ is more difficult compared to finding them on the Artificial$1$ dataset. 

After generating the datasets, we evaluate feature selection performance of the methods considered in the manuscript, and compare the results with QuickSelection. The hyperparameters used in this experiment are similar to the ones used in Section \ref{ssec:settings}, except for hidden neurons and the sparsity level. The number of hidden neurons for autoencoder-based methods has been set to $2000$, and the hyperparameter of QuickSelection,  $\epsilon$, has been adjusted to $40$. The number of selected features ($K$) is $1000$. The number of training epochs for the autoencoder-based methods is $100$. However, since the QuickSelection did not converge in $100$ epochs on the Artificial$2$ dataset, we continued the training until epoch $200$. The results of this experiment are presented in Table \ref{tab:fs_large_results}. 

 As can be seen in Table \ref{tab:fs_large_results}, QuickSelection$_{100}$ outperforms all the other methods in terms of classification accuracy on both datasets. It can also outperform the other methods in terms of clustering accuracy on the Artificial$2$ dataset. As mentioned earlier, QuickSelection achieves a higher accuracy on the Artificial$2$ dataset when it is trained for more than $100$ epochs. However, since for all the other methods we use $100$ epochs, we only consider the results of QuickSelection$_{100}$ to have a fair comparison (it should be noted that increasing the number of training epochs did not improve the results of the other methods). On noisy and very large datasets, CAE, AEFS, and FCAE have a poor performance in feature selection. In addition, they have around $30$ times more parameters than QuickSelection. CAE has the lowest accuracy among these methods; this method is very sensitive to noise. Lap\_score and MCFS have a poor performance on the Artificial$2$ dataset that is noisier than Artificial$1$. On the Artificial$1$ dataset, MCFS achieves the highest clustering accuracy. However, the memory requirement of MCFS and Lap\_score is noticeably large.  On this dataset, they need about $26 GB$ of RAM. However, QuickSelection needs only about $8 GB$ memory. In summary, QuickSelection$_{100}$ has a decent performance on these large datasets, while having the lowest number of parameters.

\begin{table}[!t]
    \caption{Characteristics of the two artificially generated datasets. The classification and clustering accuracy have been obtained using all the features.}
    \label{tab:fs_large_datasets}
    \begin{center}
    \begin{scriptsize}
    \begin{tabular}{@{}c@{\hskip 0.07in}c@{\hskip 0.07in}c@{\hskip 0.07in}c@{\hskip 0.07in}c@{\hskip 0.07in}c@{\hskip 0.07in}c@{\hskip 0.07in}c@{\hskip 0.07in}@{}}
        \toprule
        \textbf{Dataset}& \textbf{Samples}     & \textbf{Features}    &\shortstack{\textbf{Informative}\\\textbf{Features}}  &\shortstack{\textbf{Redundant}\\\textbf{Features}}& \textbf{Classes}  &\shortstack{\textbf{Classification}\\\textbf{Accuracy ($\%$)}} &\shortstack{\textbf{Clustering}\\\textbf{Accuracy ($\%$)}}  \\ \midrule
        \textbf{Artificial1} & 40000  & 8000  & 500& 3000 & 5 & 59.8 & 30.6 \\
        \textbf{Artificial2} & 40000& 8000 & 1000& 0& 5& 26.6   & 22.7  \\
        \bottomrule
    \end{tabular}
    
    \end{scriptsize}
    \end{center}
    
\end{table}

\renewcommand{\arraystretch}{1.3}

\begin{table}[!t]
    \caption{ Feature selection results on two artificially generated datasets ($K = 1000$)}
    \label{tab:fs_large_results}
    \begin{center}
    \begin{scriptsize}
        \begin{tabular}{@{}c@{\hskip 0.07in}c@{\hskip 0.07in}c@{\hskip 0.07in}c@{\hskip 0.07in}c@{\hskip 0.07in}c@{\hskip 0.07in}@{}}
        \toprule
        \multicolumn{1}{l}{} & \multicolumn{2}{c}{\textbf{Artificial1 Dataset}}                                  & \multicolumn{2}{c}{\textbf{Artificial2 Dataset}}                                 & \multicolumn{1}{l}{{\textbf{Number of Parameters}}} \\
        \textbf{Method}      & \shortstack{\textbf{Classification}\\\textbf{Accuracy ($\%$)}} & \shortstack{\textbf{Clustering}\\\textbf{Accuracy ($\%$)}} & \shortstack{\textbf{Classification}\\\textbf{Accuracy ($\%$)}}& \shortstack{\textbf{Clustering}\\\textbf{Accuracy ($\%$)}} \\ \midrule  
        \textbf{Lap\_score}  & 49.4     & 24.4& 22.0    & 21.3& -      \\
        \textbf{MCFS}        & 68.2     & \textbf{29.3}      & 24.6    & 21.7& -      \\
        \textbf{CAE}         & 23.4     & 20.6& 21.1    & 20.4& $\sim$26$\times10^6$ \\
        \textbf{AEFS}        & 34.7 & 23.3& 22.8& 21.4& $\sim$$32\times10^6$        \\
        \textbf{FCAE}        & 43.8 & 24.3& 22.9& 21.5& $\sim$$32\times10^6$         \\
        \textbf{QS$_{10}$}        & 68.1 & 25.7& 24.8& 21.21   & $\sim$0.8$\times10^6$ \\
        \textbf{QS$_{100}$}       & \textbf{68.4}& 24.8& \textbf{34.5}& \textbf{24.6}  & $\sim$0.8$\times10^6$ \\
        \textbf{QS$_{200}$}       & \textbf{-}  & -   & \textbf{39.7} & \textbf{29.6} & $\sim$0.8$\times10^6$ \\ \bottomrule
        \end{tabular}
    \end{scriptsize}
    \end{center}
\end{table}

    
\newpage
\section{Sparse Training Algorithm Analysis} \label{app_sec:appendix_sparse_training}

In this appendix, we aim to analyze the effect of the SET training procedure on the performance of QuickSelection.

We perform QuickSelection using another algorithm to obtain and train the sparse network, and then, compare the result with the original QuickSelection. We derive the sparse denoising autoencoder using the lottery ticket hypothesis algorithm \cite{frankle2018lottery}, as follows. The lottery ticket hypothesis (LTH), first, starts with training a dense network. After that, it derives the topology of the sparse network by pruning the unimportant weights of the trained dense network. Then, using both the sparse topology and the initial weight values of the connections in the dense training phase, the network is retrained. On the final obtained sparse model, we apply QuickSelection principles to select the most informative features. 

In this experiment, the structure, sparsity level, and other hyperparameters are similar to the settings described in Section \ref{ssec:settings}; we use a simple autoencoder with one hidden layer containing $1000$ hidden neurons, trained for $100$ epochs. The results of feature selection ($K=50$) are available in Tables \ref{tab:fs_cacc_sparse_training} and \ref{tab:fs_dtacc_sparse_training}. We refer to the feature selection performed using QuickSelection principles and the Sparse DAE obtained with LTH as QS$_{100}^{LTH}$. We use QS$_{100}$ for the QuickSelection that is done using the Sparse DAE obtained with SET.

As can be observed in Tables \ref{tab:fs_cacc_sparse_training} and \ref{tab:fs_dtacc_sparse_training}, in most of the cases QS$_{100}$ outperforms QS$_{100}^{LTH}$. We believe that optimizing the sparse topology and the weights, simultaneously, results in feature strength that are more meaningful for the feature selection. We discussed neuron strength in more detail in Section \ref{ssec:neuron_strength}. In addition, due to having an extra phase of dense training, the computational resource requirements of LTH are much higher than the ones of SET. To clarify this aspect, we present a comparison for the number of parameters between these two methods. The results can be found in Table \ref{tab:fs_numparam_sparse_training}. The much higher number of parameters in QS$_{100}^{LTH}$ in comparison with the number of parameters in QS$_{100}$ is given by the dense training phase of LTH.

\begin{table}[!b]
    \caption{ Clustering accuracy (\%) using $50$ selected features (except Madelon for which we select $20$ features).}
    \label{tab:fs_cacc_sparse_training}
    \begin{center}
    \begin{scriptsize}
    \begin{tabular}{@{}c@{\hskip 0.07in}c@{\hskip 0.07in}c@{\hskip 0.07in}c@{\hskip 0.07in}c@{\hskip 0.07in}c@{\hskip 0.07in}c@{\hskip 0.07in}c@{\hskip 0.07in}c@{\hskip 0.07in}@{}}
        \toprule
        \textbf{Method}    & \textbf{Coil20}   & \textbf{Isolet}   & \textbf{HAR}      & \textbf{Madelon}  & \textbf{MNIST}    & \textbf{SMK}      & \textbf{GLA}      & \textbf{PCMAC}    \\ \midrule
        \textbf{QS$_{100}$} & \textbf{60.2±2.0} & \textbf{35.1±2.7} & \textbf{54.6±4.5} & \textbf{58.2±1.5} & \textbf{48.3±2.4} & 51.8±0.8          & \textbf{59.5±1.8} & \textbf{52.5±1.1} \\
        \textbf{QS$_{100}^{LTH}$} & 58.8±3.3          & 31.2±2.4          & 50.2±6.3          & 50.8±0.5          & 37.5±4.0          & \textbf{54.6±2.7} & 54.6±3.7          & 50.8±0.6          \\ \bottomrule
    \end{tabular}

    \end{scriptsize}
    \end{center}
\end{table}

\begin{table}[!b]
    \caption{ Classification accuracy (\%) using $50$ selected features (except Madelon for which we select $20$ features).}
    \label{tab:fs_dtacc_sparse_training}
    \begin{center}
    \begin{scriptsize}
        \begin{tabular}{@{}c@{\hskip 0.07in}c@{\hskip 0.07in}c@{\hskip 0.07in}c@{\hskip 0.07in}c@{\hskip 0.07in}c@{\hskip 0.07in}c@{\hskip 0.07in}c@{\hskip 0.07in}c@{\hskip 0.07in}@{}}
        \toprule
        \textbf{Method}    & \textbf{Coil20}   & \textbf{Isolet}   & \textbf{HAR}      & \textbf{Madelon}  & \textbf{MNIST}    & \textbf{SMK}      & \textbf{GLA}      & \textbf{PCMAC}    \\ \midrule
        \textbf{QS$_{100}$} & \textbf{99.7±0.3} & \textbf{89.0±1.3} & \textbf{90.2±1.2} & \textbf{90.3±0.7} & \textbf{93.5±0.5} & \textbf{75.7±3.9} & \textbf{73.3±3.3} & 58.0±2.9 \\
        \textbf{QS$_{100}^{LTH}$} & 99.6±0.6          & 84.5±3.9          & 86.3±6.3          & 53.0±7.2          & 82.6±2.4          & 74.2±2.7 & 71.3±4.2          & \textbf{59.5±5.9} \\ \bottomrule
    \end{tabular}
    \end{scriptsize}
    \end{center}
\end{table}

\begin{table}[!b]
    \caption{ Number of parameters of QS$_{100} $ and QS$_{100}^{LTH}$  (divided by $10^6$).}
    \label{tab:fs_numparam_sparse_training}
    \begin{center}
    \begin{scriptsize}
    \begin{tabular}{@{}c@{\hskip 0.07in}c@{\hskip 0.07in}c@{\hskip 0.07in}c@{\hskip 0.07in}c@{\hskip 0.07in}c@{\hskip 0.07in}c@{\hskip 0.07in}c@{\hskip 0.07in}c@{\hskip 0.07in}@{}}
        \toprule
        \textbf{Method}    & \textbf{Coil20}    & \textbf{Isolet}    & \textbf{HAR}       & \textbf{Madelon}   & \textbf{MNIST}     & \textbf{SMK}      & \textbf{GLA}  & \textbf{PCMAC}     \\ \midrule
        \textbf{QS$_{100}$} & \textbf{0.054} & \textbf{0.043} & \textbf{0.042} & \textbf{0.040} & \textbf{0.048} & \textbf{0.566}     & \textbf{1.3} & \textbf{0.115}      \\
        \textbf{QS$_{100}^{LTH}$} & 2.054       & 1.243     & 1.142     & 1.040       & 1.548     & 40.566 &99.3      & 6.715 \\ \bottomrule
    \end{tabular}
    \end{scriptsize}
    \end{center}
\end{table}

\newpage
\section{Performance Evaluation using Random Forest Classifier} \label{app_sec:appendix_classification_RF}
In this appendix, we validate the classification accuracy results using another classifier. We repeat the experiment from Section \ref{ssec:feature_selection} in the manuscript; however, we measure the accuracy of selecting 50 features (for Madelon, we select 20 features) using the RandomForest classifier \cite{liaw2002classification} instead of the ExtraTrees classifier. The results are presented in Table \ref{tab:classification_acc2}.

As can be seen in Table \ref{tab:classification_acc2}, QuickSelection\textsubscript{100} is the best performer in 5 out of 8 cases. By comparing the results with Table \ref{tab:classification_acc} which demonstrates the classification accuracy measured by the ExtraTrees classifier, it is clear that there have been subtle changes in the accuracy values. This has resulted in some changes in the ranking of the methods in terms of the performance, as in several cases, the performance of the methods are very close. The reason behind choosing ExtraTrees classifier in the experiment was due to the low computational cost. However, as discussed in the paper, to perform an extensive evaluation, we have also measured the performance using clustering accuracy. Overall, by looking into the results of the three approaches to compute accuracy, it is clear that QuickSelection is a performant feature selection method in terms of the quality of the selected features. 

    
    \begin{table}[!h]
        
        \centering
        \caption{Classification accuracy (\%) using 50 selected features (except Madelon for which we select 20 features). On each dataset, the bold entry is the best-performer, and the italic one is the second-best performer. The classifier used for evaluation is the random forest classifier.}
        \label{tab:classification_acc2}
        \begin{center}  
        \begin{scriptsize}
           \begin{tabular}{c@{\hskip 0.07in}c@{\hskip 0.07in}c@{\hskip 0.07in}c@{\hskip 0.07in}c@{\hskip 0.07in}c@{\hskip 0.07in}c@{\hskip 0.07in}c@{\hskip 0.07in}c}
            \toprule
            \textbf{Method}&\textbf{COIL-20} &\textbf{Isolet}&\textbf{HAR} &\textbf{Madelon} &\textbf{MNIST} &\textbf{SMK}&\textbf{GLA}&\textbf{PCMAC}\\ 
            \midrule
             MCFS       &\textbf{99.5$\pm$0.3} &79.9$\pm$0.4 &88.5$\pm$0.4 &81.9$\pm$0.7                  &89.2$\pm$0.0 &76.3$\pm$3.7 &69.4$\pm$3.9 &56.5$\pm$0.16 \\
             LS         &88.9$\pm$0.8 &83.4$\pm$0.2 &86.4$\pm$0.3 &\textbf{88.9$\pm$0.6}                  &20.7$\pm$0.1 &67.9$\pm$3.1 &71.1$\pm$2.8 &50.13$\pm$0 \\
             CAE        &\textit{99.3$\pm$0.6} &\textit{89.0$\pm$0.7} &\textbf{89.8$\pm$1.0} &\textit{84.2$\pm$0.9}                  &\textbf{95.2$\pm$0.2} &\textit{76.7$\pm$4.7} &\textbf{76.6$\pm$3.8} &\textit{61.6$\pm$2.3} \\
             AEFS       &92.4$\pm$2.3 &84.9$\pm$1.7 &87.8$\pm$1.1 &59.6$\pm$4.0                  &87.6$\pm$0.8 &71.1$\pm$6.2 &67.2$\pm$4.8 &57.7$\pm$2.2 \\
             FCAE       &99.0$\pm$0.6 &85.8$\pm$5.2 &83.6$\pm$2.6 &62.7$\pm$13.1                 &69.6$\pm$2.9 &74.2$\pm$2.6 &68.9$\pm$4.0 &58.8$\pm$2.5  \\\midrule
             QS$_{10}$  &98.5$\pm$0.9 &87.0$\pm$0.7 &87.6$\pm$0.5 &81.5$\pm$3.8                  &\textit{93.6$\pm$0.6} &75.1$\pm$2.3 &68.1$\pm$4.6 &60.0$\pm$3.7 \\
             QS$_{100}$ &\textbf{99.5$\pm$0.3} &\textbf{89.1$\pm$1.3} &\textit{89.0$\pm$1.2} &\textbf{88.9$\pm$0.7}                 &93.2$\pm$0.5 &\textbf{78.5$\pm$3.5} &\textit{73.3$\pm$3.1} &\textbf{67.9$\pm$3.8} \\
            \bottomrule
            \end{tabular}
 
        \end{scriptsize} 
        \end{center}
        \end{table}

\end{document}